\documentclass[10pt,journal,compsoc]{IEEEtran}
\ifCLASSOPTIONcompsoc
  \usepackage[nocompress]{cite}
\else
  \usepackage{cite}
\fi

\ifCLASSINFOpdf
\else
\fi
\usepackage{amsmath,amssymb,amsfonts}
\usepackage{algorithmic}
\usepackage{graphicx}
\usepackage{textcomp}

\usepackage{xcolor}
\usepackage{hyperref}
\usepackage[edges]{forest}
\usepackage{tikz}
\usepackage{amsmath}
\usepackage{ulem} 
\usepackage{caption}

\usepackage{booktabs} 
\usepackage{multirow} 
\usepackage{enumitem}

\usepackage{tabularx} 
\usepackage{multicol} 
\usepackage{array}

\usepackage{float}
\usepackage{cuted} 
\usepackage{capt-of} 

\usepackage{etoolbox}
\usepackage{pdfpages}

\newcommand{\red}[1]{\textcolor{black}{#1}}

\tikzset{
basic/.style = {draw, font=\footnotesize, rectangle},
root/.style = {basic, font=\bfseries\footnotesize,  text width=3.1cm, rounded corners=2pt, semithick, align=left, fill=yellow!10},
parent/.style = {basic, text width=3.3cm, rounded corners=1.5pt, thin, align=left, fill=blue!3},
child/.style = {basic, text width=9.5cm, rounded corners=1.5pt, thin, align=left, fill=gray!5},
}

\begin{document}

\title{Structured Pruning for Deep Convolutional Neural Networks: A survey}

\author{Yang He, Lingao Xiao
\IEEEcompsocitemizethanks{
\IEEEcompsocthanksitem Manuscript received 1 March 2023; revised 11 October 2023; accepted 16 November 2023. Recommended for acceptance by V. Morariu. (Corresponding author: Yang He.)
\IEEEcompsocthanksitem Y. He and L. Xiao are with the Centre for Frontier AI Research (CFAR), Agency for Science Technology and Research (A*STAR), Singapore, and also with the Institute of High Performance Computing (IHPC), Agency for Science, Technology and Research (A*STAR), Singapore.\protect\\
E-mail: \href{mailto:hyhy1992@gmail.com}{hyhy1992@gmail.com}; \href{mailto:xiao\_lingao@outlook.com}{xiao\_lingao@outlook.com}
\IEEEcompsocthanksitem  This work is done during L. Xiao's internship under the supervision of Y. He at CFAR, A*STAR.\protect\\
}
}

%
%

\markboth{IEEE TRANSACTIONS ON PATTERN ANALYSIS AND MACHINE INTELLIGENCE, NOVEMBER 2023}%
{HE AND XIAO: STRUCTURED PRUNING FOR DEEP CONVOLUTIONAL NEURAL NETWORKS: A SURVEY}

\IEEEtitleabstractindextext{%
\begin{abstract}

The remarkable performance of deep Convolutional neural networks (CNNs) is generally attributed to their deeper and wider architectures, which can come with significant computational costs. Pruning neural networks has thus gained interest since it effectively lowers storage and computational costs. 
In contrast to weight pruning, which results in unstructured models, structured pruning provides the benefit of realistic acceleration by producing models  that are friendly to hardware implementation.
The special requirements of structured pruning have led to the discovery of numerous new challenges and the development of innovative solutions.
This article surveys the recent progress towards structured pruning of deep CNNs.
We summarize and compare the state-of-the-art structured pruning techniques with respect to filter ranking methods, regularization methods, dynamic execution, neural architecture search, the lottery ticket hypothesis, and the applications of pruning.
While discussing structured pruning algorithms, we briefly introduce the unstructured pruning counterpart to emphasize their differences.
Furthermore, we provide insights into potential research opportunities in the field of structured pruning.
A curated list of neural network pruning papers can be found at: \url{https://github.com/he-y/Awesome-Pruning}.
\red{A dedicated website offering a more interactive comparison of structured pruning methods can be found at: \url{https://huggingface.co/spaces/he-yang/Structured-Pruning-Survey}.
}

\end{abstract}

}

\maketitle

\IEEEdisplaynontitleabstractindextext
\IEEEpeerreviewmaketitle

\IEEEraisesectionheading{\section{Introduction}\label{sec:introduction}}

\IEEEPARstart{D}{eep} convolutional neural networks (CNNs) have shown exceptional performance in a wide variety of applications, including image classification~\cite{hanDeepCompressionCompressing2016}, object detection~\cite{redmon2016you}, and image segmentation~\cite{minaee2021image}, amongst others~\cite{yang2021multiple}.
Numerous CNN structures including AlexNet~\cite{krizhevsky2017imagenet}, VGGNet~\cite{simonyan2014very}, Inceptions~\cite{inception}, ResNet~\cite{he2016deep} and DenseNet~\cite{huang2017densely}
have been proposed. These architectures contain millions of parameters and require large computing power, making deployments on resource-limited hardware challenging. Model compression is a solution for this problem, aiming to reduce the number of parameters, computational cost, and memory consumption. As such, its study has gained importance.

To generate more efficient models, model compression techniques including pruning~\cite{hanLearningBothWeights2015}, quantization~\cite{rastegari2016xnor}, decomposition~\cite{denton2014exploiting}, and knowledge distillation~\cite{hinton2015distilling} have been proposed. 
The term ``pruning" refers to removing components of a network to produce sparse models for acceleration and compression. The objective of pruning is to minimize the number of parameters without significantly affecting the performance of the models. Most research on pruning has been conducted on CNNs for the image classification task, which is the foundation for other computer vision tasks.

Pruning can be categorized into unstructured~\cite{hanLearningBothWeights2015} and structured pruning~\cite{liPruningFiltersEfficient2017}. 
\textbf{Unstructured pruning} removes connections (weights) of neural networks, resulting in unstructured sparsity. Unstructured pruning often leads to a high compression rate, but requires specific hardware or library support for realistic acceleration. \textbf{Structured pruning} removes entire filters of neural networks, and can achieve realistic acceleration and compression with standard hardware by taking advantage of a highly efficient library such as the Basic Linear Algebra Subprograms (BLAS) library.

Revisiting the properties of CNNs from the perspective of structured pruning is meaningful in the era of Transformers~\cite{liu2022convnet}.
Recently, there has been an increasing trend of incorporating the architectural design of CNNs into Transformer-based models~\cite{zhang2022bootstrapping, d2021convit, chen2022dearkd, zhang2023vitaev2, ren2022co}.
Although the self-attention~\cite{vaswani2017attention}
in Transformers is effective in computing a representation of the sequence, an enormous amount of training data is still needed since Transformers often lack induction biases~\cite{chen2022dearkd,goyal2022inductive, gordon1995evaluation}.
In contrast, the structure of CNNs enforces two key inductive biases on the weights: locality and weight sharing, to influence the generalization of learning algorithms and independent of data~\cite{chen2022dearkd}.
This survey provides a better understanding of CNNs and offers insights for efficiently designing architecture for the future.

In this survey, we focus on structured pruning.
Existing surveys on related compression studies are shown in Table~\ref{tab: survey}.
Some surveys cover orthogonal fields including quantization~\cite{gholami2021survey}, knowledge distillation~\cite{gou2021knowledge}, and neural architecture search~\cite{elskenNeuralArchitectureSearch}. Some surveys~\cite{menghani2021efficient} provide a broader overview. 
Although some surveys focus on pruning, they pay more attention to unstructured pruning and cover a small number of studies on structured pruning. The number of structured pruning papers referenced in~\cite{tangSurveySparseRegularization2022, xu2020convolutional, wimmerDimensionalityReducedTraining2022, vaderaMethodsPruningDeep2022, hoeflerSparsityDeepLearning, kulkarniSurveyFilterPruning2022, blalock2020state} are 1, 11, 15, 55, 38, 10, and 20, respectively. 
We provide a more comprehensive survey with more than 200 structured pruning papers. For example, \cite{vaderaMethodsPruningDeep2022} can be covered by Section~\ref{WEIGH-DEPENDENT}, \ref{ACTIVATION-BASED}, \ref{REGULARIZATION}, \ref{TAYLOR}, \ref{LTH}, \ref{PRUNING-TOPICS}.

\begin{figure*}[ht!]
    \hspace{-0.85cm} 
    \centering
\begin{forest}
    for tree={
        forked edges,
        text centered,
        grow'=east,
        reversed=true,
        font=\footnotesize,
        rectangle, /tikz/align=left, anchor=base west, tier/.pgfmath=level(),
        rounded corners,
    }
    [, main'
    [\ref{WEIGH-DEPENDENT} Weight-Dependent, root'
        [\ref{FILTER-NORM} Filter Norm, parent'
                [
                    PFEC~\cite{liPruningFiltersEfficient2017}
                    , child
                ]
        ]
        [\ref{FILTER-CORRELATION} Filter Correlation, parent'
                [
                    FPGM~\cite{heFilterPruningGeometric2019}\text{,}
                    RED~\cite{yvinecREDLookingRedundancies2021}\text{,}
                    RED++~\cite{yvinecREDDataFreePruning2022}\text{,}
                    COP~\cite{wangCOPCustomizedDeep2019}\text{,}
                    SRR~\cite{wangConvolutionalNeuralNetwork2021}\text{,}
                    CLR-RNF~\cite{linPruningNetworksCrossLayer2022}\text{,}
                    EPruner~\cite{linNetworkPruningUsing2022}
                    , child
                ]
        ]
    ]
    [\ref{ACTIVATION-BASED} Activation-Based, root', calign=child, calign child=2 
        [\ref{CURRENT-LAYER} Current Layer, parent'
            [
                CP~\cite{heChannelPruningAccelerating2017}\text{,}
                HRank~\cite{linHRankFilterPruning2020}\text{,}
                CBC~\cite{dubeyCoresetbasedNeuralNetwork2018}\text{,}
                CHIP~\cite{suiCHIPCHannelIndependencebased2021}\text{,}
                APoZ~\cite{huNetworkTrimmingDataDriven2016}\text{,}
                DropNet~\cite{tanDropNetReducingNeural2020}\text{,}
                LRMF~\cite{zhangFilterPruningLearned2021}\text{,}
                GCNP~\cite{jiangChannelPruningUsing2022}
                , child
            ]
        ]
        [\ref{ADJACENT-LAYER} Adjacent Layer, parent', after packing node={s/.average={s}{siblings}}
            [
                ThiNet~\cite{luoThiNetFilterLevel2017}\text{,}
                AOFP~\cite{dingApproximatedOracleFilter2019}\text{,}
                GFS~\cite{yeGoodSubnetworksProvably2020}
                , child
            ]
        ]
        [\ref{ALL-LAYER} All Layer, parent'
            [
                NISP~\cite{yuNISPPruningNetworks2018a}\text{,}
                DCP~\cite{zhuangDiscriminationawareChannelPruning2018}\text{,}
                PFP~\cite{liebenweinProvableFilterPruning2020}\text{,}
                DLRFC~\cite{heFilterPruningFeature2022}
                , child
            ]
        ]
    ]
    [\ref{REGULARIZATION} Regularization, root', calign=child, calign child=2
        [\ref{REG-ON-BN} on BN Parameters, parent'
                [
                    NS~\cite{liuLearningEfficientConvolutional2017}\text{,}
                    GBN~\cite{youGateDecoratorGlobal2019}\text{,}
                    PR~\cite{zhuangNeuronlevelStructuredPruning2020}\text{,}
                    RSNLI~\cite{yeRethinkingSmallernormlessinformativeAssumption2018}\text{,}
                    SCP~\cite{kangOperationAwareSoftChannel2020}\text{,}
                    EagleEye~\cite{liEagleEyeFastSubnet2020}
                    , child
                ]
        ]
        [\ref{REG-ON-EXTRA} on Extra Parameters, parent'
                [
                    SSS~\cite{huangDataDrivenSparseStructure2018}\text{,}
                    GAL~\cite{linOptimalStructuredCNN2019}\text{,}
                    DMC~\cite{gaoDiscreteModelCompression2020}\text{,}
                    GDP-Guo~\cite{guoGDPStabilizedNeural2021}\text{,}
                    ResRep~\cite{dingResRepLosslessCNN2021}\text{,}
                    SCOP~\cite{tangSCOPScientificControl2020}\text{,}\\
                    BAR~\cite{lemaireStructuredPruningNeural2019}\text{,}
                    ABP~\cite{tianAddingPruningSparse2021}\text{,}
                    WhiteBox~\cite{zhangCarryingOutCNN2022}\text{,}
                    LeGR~\cite{chinEfficientModelCompression2020}\text{,}
                    ML$_1$R~\cite{xieLearningOptimizedStructure2020}
                    , child
                ]
        ]
        [\ref{REG-ON-FILTER} on Filters, parent'
                [
                    SSL~\cite{wenLearningStructuredSparsity2016}\text{,}
                    OICSR~\cite{liOICSROutInChannelSparsity2019}\text{,}
                    OTO~\cite{chenOnlyTrainOnce2021}\text{,}
                    GREG~\cite{wangNeuralPruningGrowing2022}
                    , child
                ]
        ]
    ]
    [\ref{OPTIMIZATION-TOOLS} Optimization Tools, root', calign=child, calign child=2
        [\ref{TAYLOR} Taylor Expansion, parent'
            [
                First-Order~\cite{molchanovPruningConvolutionalNeural2017, molchanovImportanceEstimationNeural2019}\text{,}
                Second-Order~\cite{pengCollaborativeChannelPruning2019, wangEigenDamageStructuredPruning2019, liuGroupFisherPruning2021, yuHessianAwarePruningOptimal2022, nonnenmacherSOSPEfficientlyCapturing2022}
                , child
            ]
        ]
        [\ref{BAYESIAN} Variational Bayesian, parent'
            [
                VP~\cite{zhaoVariationalConvolutionalNeural2019}\text{,}
                RBP~\cite{zhouAccelerateCNNRecursive2019}\text{,}
                VIBNet~\cite{dai2018compressing}\text{,}
                Horseshoe~\cite{louizos2017bayesian}\text{,}
                Log-normal~\cite{neklyudov2017structured}
                , child
            ]
        ]
        [\ref{OPTIMIZATION-OTHER} Others, parent'
            [
                SGD\cite{dingCentripetalSGDPruning2019, huangTrainingStructuredNeural2022, mehtaImplicitFilterLevel2019, leeEnsembleKnowledgeGuided2022}\text{,}
                ADMM~\cite{zhangStructADMMAchievingUltrahigh2022, maNonStructuredDNNWeight2022}\text{,}
                BO~\cite{fanBayesianOptimizationClustering2022}\text{,}
                ST~\cite{kusupatiSoftThresholdWeight2020}
                , child
            ]
        ]
    ]
    [\ref{DYNAMIC} Dynamic Pruning, root'
        [\ref{DYNAMIC-TRAINING} Dynamic during Training, parent'
                [
                    SFP~\cite{heSoftFilterPruning2018}\text{,}
                    GDP-Lin~\cite{linAcceleratingConvolutionalNetworks2018}\text{,}
                    DPF~\cite{linDynamicModelPruning2022}\text{,}
                    CHEX~\cite{houCHEXCHannelEXploration2022}\text{,}
                    DSG~\cite{liuDynamicSparseGraph2019}\text{,}
                    SEP~\cite{dingWherePruneUsing2021}\text{,}\\
                    DCP-CAC~\cite{chenDynamicalChannelPruning2021}\text{,}
                    SMCP~\cite{humbleSoftMaskingCostConstrained2022}
                    , child
                ]
        ]
        [\ref{DYNAMIC-INFERENCE} Dynamic during Inference, parent'
                [
                    RNP~\cite{linRuntimeNeuralPruning2017}\text{,}
                    FBS~\cite{gaoDynamicChannelPruning2019}\text{,}
                    ManiDP~\cite{tangManifoldRegularizedDynamic2021}\text{,}
                    DRLP~\cite{chenStorageEfficientDynamic2020}\text{,}
                    DDG~\cite{liDynamicDualGating2021}\text{,}
                    FTWT~\cite{elkerdawyFireTogetherWire2022}\text{,}
                    CDG~\cite{mengContrastiveDualGating2022}
                    , child
                ]
        ]
    ]
    [\ref{NEURAL-ARCHITECTURE-SEARCH} \red{NAS-Based Pruning}, root', calign=child, calign child=2
        [\ref{REINFORCEMENT-LEARNING-BASED} Reinforcement Learning-Based, parent'
                [
                    AMC~\cite{heAMCAutoMLModel2018}\text{,}
                    AGMC~\cite{yuAutoGraphEncoderDecoder2021}\text{,}
                    DECORE~\cite{alwaniDECOREDeepCompression2022}\text{,}
                    GNN-RL~\cite{yuTopologyAwareNetworkPruning2022}\text{,}\\
                    AutoCompress~\cite{liuAutoCompressAutomaticDNN2020}\text{,}
                    RL-MCTS~\cite{wangChannelPruningLookahead2022}
                    , child
                ]
        ]
        [\ref{GRADIENT-BASED} Gradient-Based, parent'
                [
                    DMCP~\cite{guoDMCPDifferentiableMarkov2020}\text{,}
                    DSA~\cite{ningDSAMoreEfficient2020}\text{,}
                    DHP~\cite{liDHPDifferentiableMeta2020}\text{,}
                    PaS~\cite{liPruningasSearchEfficientNeural2022}\text{,}
                    LFPC~\cite{heLearningFilterPruning2020}\text{,}
                    TAS~\cite{dongNetworkPruningTransformable2019}\text{,}\\
                    EE~\cite{zhangExplorationEstimationModel2021}\text{,}
                    DDNP~\cite{gaoDisentangledDifferentiableNetwork2022}\text{,}
                    MFP~\cite{heFilterPruningSwitching2022}\text{,}
                    DNCP~\cite{zhengModelCompressionBased2022}\text{,}
                    DAIS~\cite{guanDAISAutomaticChannel2022}\text{,}
                    ReCNAS~\cite{pengReCNASResourceConstrainedNeural2022}
                    , child
                ]
        ]
        [\ref{EVOLUTIONARY-BASED} Evolutionary-Based, parent'
                [   
                    MetaPruning~\cite{liuMetaPruningMetaLearning2019}\text{,}
                    ABCPruner~\cite{linChannelPruningAutomatic2020}\text{,}
                    CCEP~\cite{shangNeuralNetworkPruning2022}\text{,}
                    EDropout~\cite{salehinejadEDropoutEnergyBasedDropout2022}
                    , child
                ]
        ]
    ]
    [\ref{COMPRESSION-EXTENSIONS} Extensions, root', calign=child, calign child=2
        [\ref{LTH} Lottery Ticket Hypothesis, parent'
            [
                RVNP~\cite{liuRethinkingValueNetwork2019}\text{,}
                EB~\cite{youDrawingEarlyBirdTickets2020}\text{,}
                ProsPr~\cite{alizadehProspectPruningFinding2022}\text{,}
                EarlyCroP~\cite{rachwanWinningLotteryAhead2022}\text{,}
                PaT~\cite{shenWhenPrunePolicy2022}\text{,}\\
                PnS~\cite{fischerPlantSeekCan2022}\text{,}
                SuperTickets~\cite{youSuperTicketsDrawingTaskAgnostic2022}\text{,}
                Cunha22~\cite{cunhaProvingLotteryTicket2022}\text{,}
                RRCP~\cite{liRevisitingRandomChannel2022}
                , child
            ]
        ]
        [\ref{JOINT-COMPRESSION} Joint Compression, parent'
            [
                NPAS~\cite{liNPASCompilerAwareFramework2021}\text{,}
                DJPQ~\cite{wangDifferentiableJointPruning2020}\text{,}
                BB~\cite{vanbaalenBayesianBitsUnifying2020}\text{,}
                IODF~\cite{wangFastLosslessNeural2022}\text{,}
                APQ~\cite{wangAPQJointSearch2020}\text{,}
                Hinge~\cite{liGroupSparsityHinge2020}\text{,}
                CC~\cite{liCompactCNNsCollaborative2021}\text{,}
                NM~\cite{kimNeuronMergingCompensating2020}\text{,}
                EDP~\cite{ruanEDPEfficientDecomposition2021}
                , child
            ]
        ]
        [\ref{PRUNING-FOR-SPECIAL-GRANULARITY} Special Granularity, parent'
            [
                GBD~\cite{lebedevFastConvNetsUsing2016}\text{,}
                SWP~\cite{mengPruningFilterFilter2020}\text{,}
                PCONV~\cite{maPCONVMissingDesirable2020}\text{,}
                GKP-TMI~\cite{zhongRevisitKernelPruning2022}\text{,}
                1xN~\cite{lin1xNPatternPruning2022}\text{,}
                SDN~\cite{chenShallowingDeepNetworks2019}\text{,}
                JMDP~\cite{liuJointMultiDimensionPruning2021}\text{,}
                SOKS~\cite{liuSOKSAutomaticSearching2022}\text{,}
                DPP~\cite{gonzalez-carabarinDynamicProbabilisticPruning2022}\text{,}
                JCW~\cite{zhaoMultigranularityPruningModel2022}
                , child
            ]
        ]
    ]
    [\ref{FUTURE-DIRECTIONS} Future Directions, root'
        [\ref{PRUNING-TOPICS} Pruning Topics, parent'
            [
                Theory~\cite{tanakaPruningNeuralNetworks2020, lee2019signal, linCompactConvNetsStructureSparsity2020, ganjdaneshInterpretationsSteeredNetwork2022, leeEnsembleKnowledgeGuided2022, bartoldson2020generalization, luo2017entropy, orseau2020logarithmic, yangProxSGDTrainingStructured2020, dingGlobalSparseMomentum2019, barsbeyHeavyTailsSGD2021}\text{,}\\
                Mechanism~\cite{miaoLearningPruningFriendlyNetworks2022, zhangOneshotPruningRecurrent2019, peste2021ac}\text{,}
                Rate~\cite{leeLayeradaptiveSparsityMagnitudebased2021, liebenwein2021compressing}\text{,}
                Domain~\cite{liu2018frequency, zhangFilterPruningLearned2021}
                , child
            ]
        ]
        [\ref{PRUNING-FOR-SPECIFIC-TASK} Pruning for Specific Tasks, parent'
            [
                FL~\cite{zhangFedDUAPFederatedLearning2022, jiangModelPruningEnables2022}\text{,}
                CL~\cite{pengOvercomingLongTermCatastrophic2022, golkar2019continual, yan2022learning, mallyaPackNetAddingMultiple2018}\text{,}
                Limited Dataset~\cite{tangRebornFiltersPruning2020}\text{,}\\
                Others\cite{zhanAchievingOnMobileRealTime2021, wangSoftPersonReidentification2022, fernandesAutomaticSearchingPruning2021, linFairGRAPEFairnessAwareGRAdient2022, bianSubarchitectureEnsemblePruning2022, whitaker2022prune}
                , child
            ]
        ]
        [\ref{PRUNING-SPECIFIC-NETWORKS} Pruning Specific Networks, parent'
            [
                GAN~\cite{shuCoEvolutionaryCompressionUnpaired2019, liuContentawareGANCompression2021, liRevisitingDiscriminatorGAN2021, songSPGANSelfGrowingPruning2021}\text{,}
                Transformers~\cite{liu2022convnet, chavan2022vision}\text{,}\\
                AGI~\cite{bommasani2021opportunities, brown2020language, reed2022a}\text{,}
                Others~\cite{wangWeightNoiseInjectionBased2019, serra2021scaling, kim2022exploring, chowdhury2022towards}
                , child
            ]
        ]
        [\ref{PRUNING-TARGET} Pruning Targets, parent'
            [
                Hardware~\cite{kimCPruneCompilerInformedModel2022}\text{,}
                Energy~\cite{yangDesigningEnergyefficientConvolutional2017}\text{,}
                Robustness~\cite{guiModelCompressionAdversarial2019, chenLinearityGraftingRelaxed2022, madaanAdversarialNeuralPruning2020}
                , child
            ]
        ]
    ]
    ]
\end{forest}
\caption{Taxonomy for structured pruning.}
\label{fig: taxonomy}
\end{figure*}
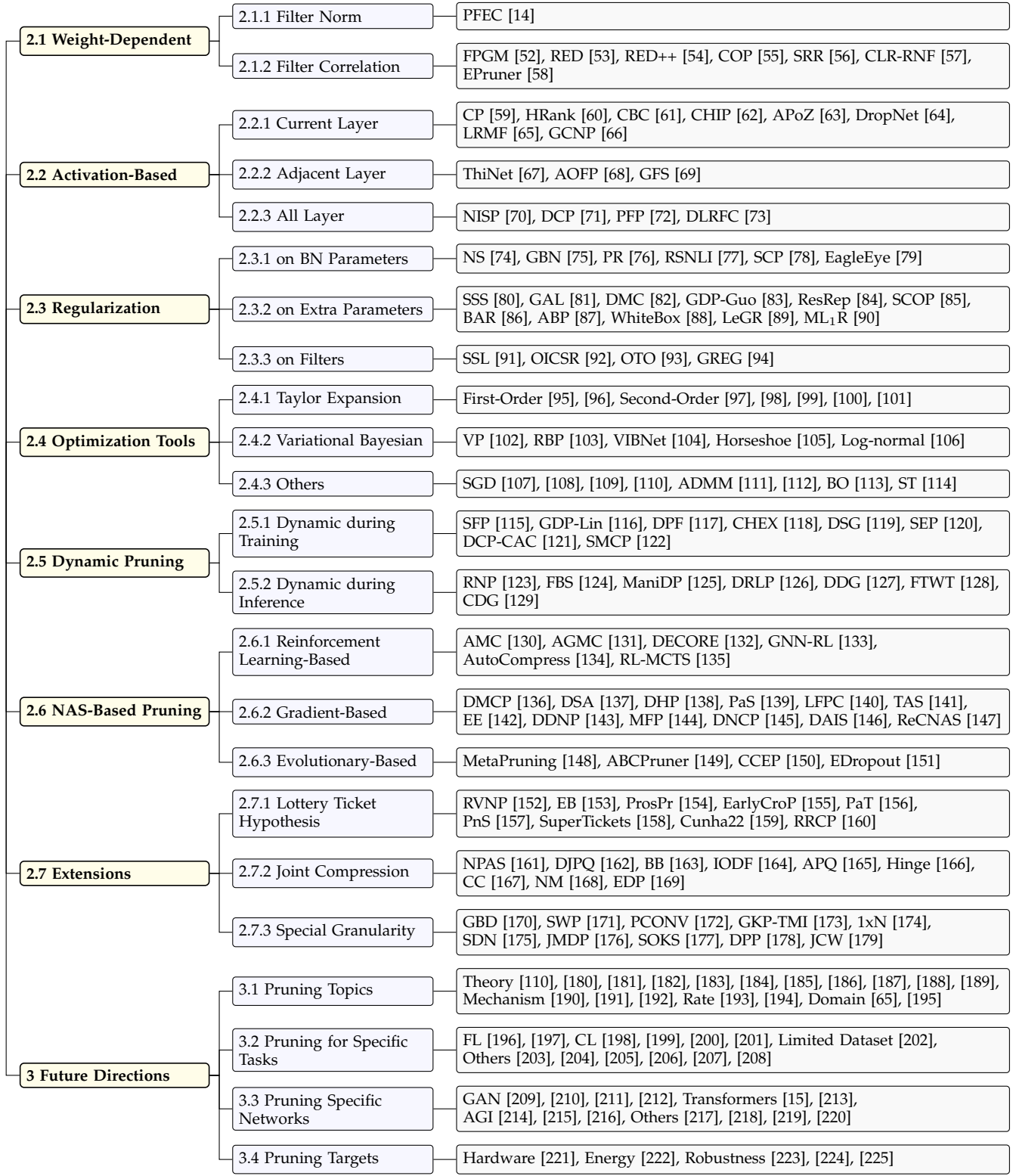

The survey is arranged as follows.
In the taxonomy (Fig.~\ref{fig: taxonomy}), we group the structured pruning methods into different categories.
Each subsection of Section~\ref{METHOD} corresponds to a category of structured pruning methods.
Most methods are first developed in an unstructured manner and then extended to meet structural constraints. While some studies span multiple categories, we place them in the most appropriate categories that serve this survey. 
Section~\ref{FUTURE-DIRECTIONS} then introduces some potential and promising future directions.
Due to the length constraints, only the most representative studies are discussed in detail.

\begin{table}[!t]
    \centering
    \begin{tabular}{l c c c c c c}
    \toprule
        Papers & P. & Q. & D. & KD & NAS \\ \midrule
        \cite{gholami2021survey, guo2018survey, qin2020binary} & ~ & \checkmark & ~ & ~ & ~ \\ \midrule
        \cite{gou2021knowledge, wangKnowledgeDistillationStudentTeacher2022} & ~ & ~ & ~ & \checkmark & ~ \\ \midrule
        \cite{elskenNeuralArchitectureSearch, renComprehensiveSurveyNeural2021, zhouSurveyEvolutionaryConstruction2021, liuSurveyEvolutionaryNeural2023} & ~ & ~ & ~ & ~ & \checkmark \\ \midrule
        \cite{tangSurveySparseRegularization2022, xu2020convolutional, wimmerDimensionalityReducedTraining2022, vaderaMethodsPruningDeep2022, hoeflerSparsityDeepLearning, kulkarniSurveyFilterPruning2022, blalock2020state} & \checkmark & ~ & ~ & ~ & ~ \\ \midrule
        \cite{LIANG2021370, xu2018scaling} & \checkmark & \checkmark & ~ & ~ & ~ \\ \midrule
        \cite{cheng2018model, cheng2018recent, szeEfficientProcessingDeep2017a, zhaoSurveyDeepLearning2022} & \checkmark & \checkmark & \checkmark & \checkmark & ~ \\ \midrule
        \cite{menghani2021efficient, lebedev2018speeding, goel2020survey, ghimire2022survey, capra2020hardware, dengModelCompressionHardware2020} & \checkmark & \checkmark & \checkmark & \checkmark & \checkmark \\
    \bottomrule
    \end{tabular}
    \caption{Categorize existing surveys into Pruning (P.), Quantization (Q.), Decomposition (D.), Knowledge Distillation (KD), and Neural Architecture Search (NAS).}
    \label{tab: survey}
\end{table}

\forestset{
  main'/.style={
    l sep=5mm,
    anchor=west,
  },
  root'/.style={root,
    anchor=west,
    edge path={
     \noexpand\path[\forestoption{edge}]
     ($(!u.east)!.0!(!.west)$) ++(0.5,0) |- (!.west);
    },
  },
  parent'/.style={parent, 
    anchor=west,
    calign=child edge,
    l sep=0.4cm
  },
}

\section{Methods}
\label{METHOD}

\textbf{Preliminaries:} A deep convolutional neural network $\mathcal{N}$ can be parameterized by $\left\{\mathbf{W}^{l} \in \mathbb{R}^{N_{l+1} \times N_l \times K_{l} \times K_{l}}, 1 \leq l \leq L\right\}$. At $l$-th layer, the input tensor $\mathbf{I}^l$ has shape $N_l \times H_l \times W_l$, and the output tensor $\mathbf{O}^l$ has shape $N_{l+1} \times H_{l+1} \times W_{l+1}$. $N_{l}$ and $N_{l+1}$ denote the channel number of input and output tensors in the $l$-th layer, respectively. $\mathbf{W}^{l}$ represents the connections (weights) between the input tensor $\mathbf{I}^l$ and the output tensor (feature map) $\mathbf{O}^l$. The weight matrix $\mathbf{W}^l$ is made of $N_{l+1}$ 3-D filters $\mathbf{F}^l$. Specifically, the $i$-th filter in $l$-th layer can be denoted as $\left\{\mathbf{F}_i^l \in \mathbb{R}^{N_l \times H_l \times W_l}, 1 \leq i \leq N_{l+1}\right\}$. In this paper, we call $\mathbf{K}^l \in \mathbb{R}^{K_l \times K_l}$ 2-D kernels, so a filter has $N_l$ kernels of kernel size $K_l$. 
To express a single weight, we use $\left\{w = \mathbf{F}^l_i(n, k_1, k_2), 1 \leq n \leq N_l, 1 \leq k_1, k_2 \leq K_l\right\}$. The convolution operation for $l$-th layer can be expressed as:
\begin{equation}
    \label{eq:1}
    \mathbf{O}^l_i = \mathbf{I}^l * \mathbf{F}^l_i \textbf{ for } 1 \leq i \leq N_{l+1}
\end{equation}
where $*$ denotes the convolution operator.

Structured pruning, such as filter pruning, aims to:
\begin{equation}
    \begin{aligned}
        \min _{\mathbf{F}} \mathcal{L}(\mathbf{F} ; \mathcal{D})=\min _{\mathbf{F}} \frac{1}{N} \sum_{i=1}^N \mathcal{L}\left(\mathbf{F} ;\left(\mathbf{x}_i, \mathbf{y}_i\right)\right), \\
        \text { s.t. } \operatorname{Card}(\mathbf{F}) \leq \kappa \text {, }
    \end{aligned}
\end{equation}
where $\mathcal{L}(\cdot)$ is the loss function (e.g., cross-entropy loss) and $\mathcal{D}=\left\{\left(\mathbf{x}_i, \mathbf{y}_i\right)\right\}_{i=1}^N$ is a dataset. $\operatorname{Card}(\cdot)$ is the cardinality of the filter set, and $\kappa$ is the target sparsity level such as the number of remaining nonzero filters.

\subsection{Weight-Dependent}
\label{WEIGH-DEPENDENT}

\red{
Weight-dependent criteria are specifically designed to evaluate the importance of filters within a neural network. This is accomplished by assessing the weights of these filters to identify which filters and/or channels are crucial for the model's performance.
}
Compared with activation-based methods, weight-dependent methods do not involve input data. As such, weight-dependent methods are considered straightforward and require lower computational costs. There are two subcategories of weight-dependent criteria: \textbf{filter norm} and \textbf{filter correlation}.
Calculating the norm of a filter is done independently of the norm of other filters, while calculating filter correlation involves multiple filters.

\subsubsection{Filter Norm}
\label{FILTER-NORM}
Unlike unstructured pruning that uses the magnitude of the weights as the metric, structured pruning computes the filter norm values to be the metric. The $\ell_p$-norm of a filter can be written as:

\begin{equation}
    \left\|\mathbf{F}^l_i\right\|_p=\sqrt[p]{\sum_{n=1}^{N_l} \sum_{k_1=1}^{K_l} \sum_{k_2=1}^{K_l}\left|\mathbf{F}^l_i\left(n, k_1, k_2\right)\right|^p},
\end{equation}

\noindent
where $i \in N_{l+1}$ represents the $i$-th filter in $l$-th layer, $N_l$ is the input channel size, and $K_l$ is the kernel size. $p$ is the order of the norm, and the two common norms are $\ell_1$-norm (Manhattan norm) and $\ell_2$-norm (Euclidean norm).

Pruning Filter for Efficient ConvNets (PFEC)~\cite{liPruningFiltersEfficient2017} computes the filter importance based on $\ell_1$-norm. Li~\textit{et al.} consider filters with smaller norms to have a weak activation and contribute less to the final classification decision~\cite{liPruningFiltersEfficient2017}.

Soft Filter Pruning (SFP)~\cite{heSoftFilterPruning2018} empirically finds that $\ell_2$-norm works slightly better than $\ell_1$-norm. This is discussed further in Section~\ref{DYNAMIC-TRAINING}.

\subsubsection{Filter Correlation}
\label{FILTER-CORRELATION}
Filter Pruning via Geometric Median (FPGM)~\cite{heFilterPruningGeometric2019} reveals the ``smaller-norm-less-important" assumption to not always be true, based on the real distribution of the neural networks. Instead of pruning away unimportant filters, it finds redundant filters by exploiting relationships among filters of the same layer. He \textit{et al.} consider the filters close to the geometric median to be redundant because they represent common information shared by all filters in the same layer~\cite{heFilterPruningGeometric2019}. These redundant filters can be removed without significantly influencing the performance.

RED~\cite{yvinecREDLookingRedundancies2021} uses a data-free structured compression method.
It consists of three steps. First, scalar hashing is conducted on weights in each layer. Second, redundant filters are merged based on the relative similarity of the filters. Third, a novel uneven depthwise separation technique is used to prune layers. In RED++~\cite{yvinecREDDataFreePruning2022}, the third step is replaced with an \textit{input-wise splitting} technique to remove redundant operations such as multiplication and addition. The reason behind this is that mathematical operations are more of a bottleneck compared to memory allocation.

Unlike FPGM~\cite{heFilterPruningGeometric2019}, which measures filter importance within layers, Correlation-based Pruning (COP)~\cite{wangCOPCustomizedDeep2019} compares the importance of the cross-layer filters. To determine the redundancy among filters within a layer, COP~\cite{wangCOPCustomizedDeep2019} first conducts a Pearson correlation test. Next, a layer-wise max-normalization is used to address the scaling effect of the correlation-based importance metric in order to rank the filters across layers. Lastly, a cost-aware regularization term is added to the global filter-importance calculation to allow users to have finer control over the budget.

Structural Redundancy Reduction (SRR)~\cite{wangConvolutionalNeuralNetwork2021} exploits the structural redundancy by looking for the most redundant layer, instead of the least ranked filters among all layers. First, filters in each layer are established as a graph. The redundancy of a graph can be evaluated by its two associated properties, i.e., quotient space size and $\ell$-covering number. The two properties with large values indicate a complex, and thus a less redundant, graph. 
Within the most redundant layer, a filter norm can be applied to prune the least important filters.
Finally, the graph of the layer is re-established, and the layers' redundancy is re-evaluated.

\subsection{Activation-Based}
\label{ACTIVATION-BASED}

\red{
Instead of determining filter importance through their weights, activation-based pruning methods harness the activation maps for pruning decisions.
Activation maps, detailed in Eq.~\ref{eq:1}, are produced from the convolutional process between input data and filters. 
}
Channel pruning is another name for filter pruning since removing the channels of activation maps is equivalent to removing the filters. In addition to the effect of the current layer, filter pruning also influences the next layer's filter through feature maps.

To evaluate filters in layer $l$, we can exploit the information on activation maps of:\\
\begin{enumerate}[leftmargin=*]
    \item \textbf{the current layer} - the channel importance can be evaluated by using the reconstruction error~\cite{heChannelPruningAccelerating2017}, the decomposition of the activation map~\cite{linHRankFilterPruning2020}, utilization of channel-independence~\cite{suiCHIPCHannelIndependencebased2021}, and post-activations~\cite{huNetworkTrimmingDataDriven2016, tanDropNetReducingNeural2020};
    \item \textbf{the adjacent layers} - redundant channels can be effectively identified by exploiting the dependency between the current layer and the next layer~\cite{luoThiNetFilterLevel2017,dingApproximatedOracleFilter2019}. In addition, activation maps of the previous layer can also be utilized to guide the pruning decision~\cite{gaoDynamicChannelPruning2019,linRuntimeNeuralPruning2017};
    \item \textbf{all layers} - the holistic effect of removing a filter can be evaluated by minimizing the reconstruction error of the Final Response Layer~\cite{yuNISPPruningNetworks2018a} and considering the discriminative power of all layers~\cite{zhuangDiscriminationawareChannelPruning2018}.
\end{enumerate}

\subsubsection{Current Layer}
\label{CURRENT-LAYER}

Channel Pruning (CP)~\cite{heChannelPruningAccelerating2017} uses layer $l$'s (current) activation maps to guide the pruning of layer $l$'s filters. It models layer-wise channel pruning as an optimization problem that minimizes the reconstruction error of sparse activation maps. Solving the optimization problems involves two alternating steps. (1) To find which channels to prune, CP explicitly solves the LASSO regression by fixing the weights rather than imposing a sparsity regularization to training loss. (2) To minimize the reconstruction error of layer $l$'s feature map, weights are fine-tuned with the fixed pruning decision.

HRank~\cite{linHRankFilterPruning2020} uses the average rank of the current layer's activation maps from a small set of input data as the filter importance. An important finding is that regardless of the data received, a single filter generates activation maps with the same average rank. To find the average rank, the Singular Value Decomposition (SVD) is adopted. The decomposition conducted here is to find the rank rather than to reduce computational cost. After determining the average rank, a layer-wise pruning algorithm is then proposed to retain top-$k$ filters.

Coreset-Based Compression (CBC)~\cite{dubeyCoresetbasedNeuralNetwork2018} adopts filter pruning to pre-process filters for coreset-based compression~\cite{baykal2018data}. The scoring of the filters is based on the mean of activation norms over the entire training set. 
A binary search is then used to find the smallest number of filters that satisfy the accuracy constraint. After pruning, three coreset-based compression techniques are discussed, including \textit{k-Means, Structured Sparse PCA, and Activation-Weighted coresets}. 
Utilizing Deep Compression~\cite{hanDeepCompressionCompressing2016}, \textit{Activation-Weighted coresets} outperforms the rest.

CHannel Independence (CHIP) is used by Sui \textit{et al.} to evaluate channel importance~\cite{suiCHIPCHannelIndependencebased2021}. Channel independence is determined by the cross-channel correlation, indicating whether a channel is linearly dependent on other channels. The greater the independence of the channel, the higher its importance. Channel importance is determined by measuring the activation maps' nuclear norm change.

Average Percentage of Zero (APoZ)~\cite{huNetworkTrimmingDataDriven2016} utilizes the current layer's \textbf{post-activation} maps, which are the activation maps after activation functions such as ReLU. The \textit{average percentage of zeros (APoZ)} in the post-activation maps are used to measure the importance of channels. A small \textit{APoZ} value means that most parts of the activation maps are being activated, so these activation maps contribute more to the final results and are more important.

DropNet~\cite{tanDropNetReducingNeural2020} utilizes the post-activation maps' \textbf{average magnitude} as the metric. Under this metric, a small non-zero activation value, which is considered important by APoZ~\cite{huNetworkTrimmingDataDriven2016}, is no longer important in DropNet~\cite{tanDropNetReducingNeural2020}.
There are two reasons for the use of this metric. First, a small \textbf{average magnitude} indicates the presence of many inactive nodes. Second, the small magnitude also means these nodes are less adaptive to learning.

\subsubsection{Adjacent Layer}
\label{ADJACENT-LAYER}
ThiNet~\cite{luoThiNetFilterLevel2017} uses layer $l + 1$'s (the next layer) activation maps to guide the pruning of the layer $l$ (the current layer). The main idea is to approximate layer $l+1$'s activation maps with \textbf{subsets} of layer $l$'s activation maps. Channels outside these subsets are pruned. To find these subsets, a greedy algorithm is proposed. Specifically, the algorithm greedily adds channels to an initially empty set and measures the reconstruction error. Subsets that have the least reconstruction error and meet the sparsity constraint will be selected.

Approximated Oracle Filter Pruning (AOFP)~\cite{dingApproximatedOracleFilter2019} uses layer $l+1$'s activation maps, and targets at pruning without heuristic knowledge which is often required by \textit{Oracle Pruning} methods~\cite{liPruningFiltersEfficient2017, huNetworkTrimmingDataDriven2016,molchanovPruningConvolutionalNeural2017}. \textbf{Firstly}, the concept of \textit{damage isolation} is introduced to avoid using heuristic importance metrics. \textit{Damage isolation} means the damage caused by pruning layer $l$ is isolated by layer $l+1$, making the damage invisible to $l+2$. \textbf{Secondly}, a multi-path framework is used to benefit from parallel scoring and fine-tuning. \textbf{Thirdly}, the \textit{binary filter search} method is used to solve problems of the multi-path framework.

In addition to using the next layer's activation maps, Runtime Neural Pruning (RNP)~\cite{linRuntimeNeuralPruning2017} and Feature Boosting and Suppression (FBS)~\cite{gaoDynamicChannelPruning2019} utilize the layer $l-1$'s (previous layer) activation maps to guide the pruning of the current layer. Both methods use the global average pooling result of the previous layer as the filter importance. This is further discussed in Section \ref{DYNAMIC-INFERENCE}, since both methods conduct dynamic pruning during inference.

\subsubsection{All Layer}
\label{ALL-LAYER}
Despite the success of existing methods, proponents of Neuron Importance Score Propagation (NISP)~\cite{yuNISPPruningNetworks2018a} argue that most methods did not consider the reconstruction error propagation. NISP proposes to use the \textit{Final Response Layer (FRL)} to determine the neuron importance because reconstruction errors from all previous layers will eventually be propagated to the \textit{FRL}. Initially, the importance score of \textit{FRL} can be determined by any feature ranking technique, i.e., Inf-FS~\cite{roffoInfiniteFeatureSelection2015}. The neuron importance is then propagated backward from \textit{FRL} to the previous layers. Lastly, neurons with low importance scores in the layer are pruned. Pruned neurons will no longer back-propagate scores to the previous layers.

Discrimination-aware Channel Pruning (DCP)~\cite{zhuangDiscriminationawareChannelPruning2018} aims to keep discriminative channels that substantially change the final loss in their absence. However, pruning shallow layers often triggers a smaller decrease in final loss due to the long propagation path. To resolve the problem, Zhuang \textit{et al.} introduce discrimination-aware losses to every last layer of the intermediate layers. A greedy algorithm is then used to select channels based on the discrimination-aware loss and reconstruction loss between the baseline and pruned networks.

\subsection{Regularization}
\label{REGULARIZATION}
Regularization can be used for learning structured sparse networks by adding different sparsity regularizers $R_{s}(\cdot)$. The sparsity regularizer can be applied to \textbf{BN parameters} if the networks contain batch normalization layers. To achieve structured sparsity, BN parameters are used to indicate the pruning decision of structures such as channels or filters. \textbf{Extra parameters} working as learnable gates have been introduced to guide pruning. With these extra parameters, networks no longer require batch normalization layers. Sparsity regularizers can also be directly applied to \textbf{filters}. Group Lasso regularization is commonly used to sparsify filters in a structured manner.

\begin{figure}
    \includegraphics[width=\columnwidth]{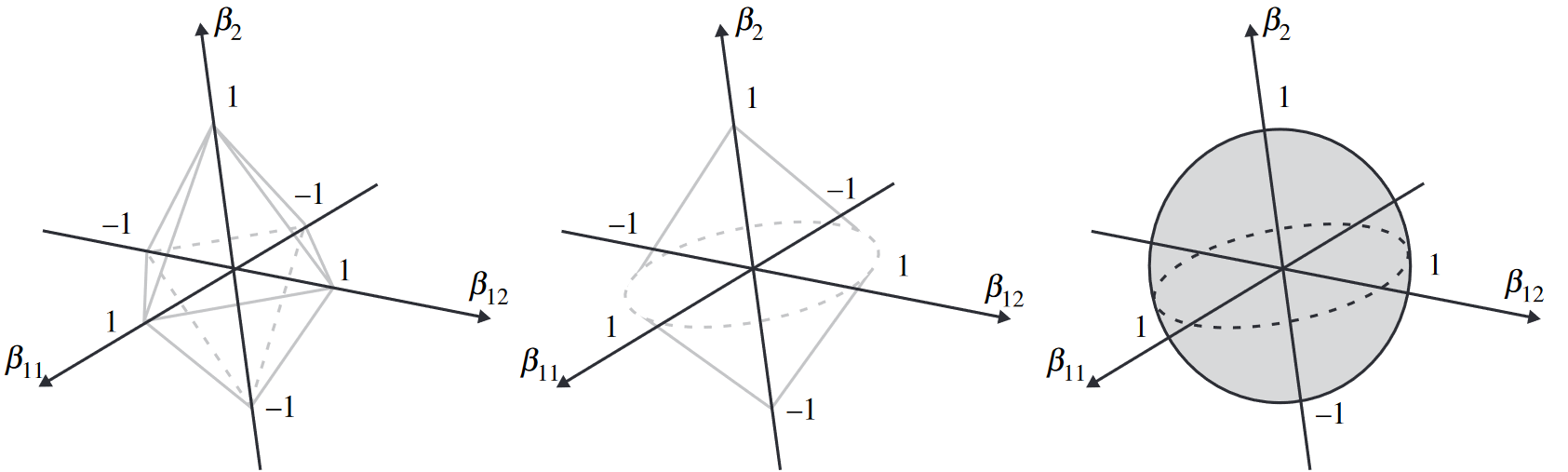}
    \caption{The left image shows the $\ell_1$ penalty used in Lasso. The middle image shows the $\ell_{2,1}$ penalty function used in Group Lasso, and the right image shows the $\ell_2$ penalty function used in Ridge. $\beta_{11}$ and $\beta_{12}$ are grouped together for Group Lasso. (The image is taken from~\cite{yuanModelSelectionEstimation2006}.)}
    \label{fig: reg}
\end{figure}

The general Group Lasso is defined as the solution to the following convex optimization problem~\cite{hoeflerSparsityDeepLearning},
\begin{equation}
\min _{\beta \in \mathbb{R}^p}\left(\left\|\mathbf{y}-\sum_{g=1}^G \mathbf{X}_{\mathbf{g}} \beta_g\right\|_2^2+\lambda \sum_{g=1}^G \sqrt{n_g}\left\|\beta_g\right\|_2\right)
\end{equation}
where the feature matrix is divided into $G$ groups, forming the matrix $\mathbf{X}_g$ that contains only examples of group $g$ as well as the corresponding coefficient vector $\beta_g$. $n_g$ indicates the size of group $g$, and $\lambda \ge 0$ is a tuning parameter. In the context of filter pruning, the first term can be viewed as the reconstruction error of the feature map, and the second term can be rewritten as:
\begin{equation}
    \lambda \left\|\mathbf{F}^l\right\|_{2,1} = 
    \lambda \sum_{i=1}^{N_{l+1}}\left\|\mathbf{F}_i^l\right\|_2
\end{equation}
where groups $G$ are replaced by output channels $N_{l+1}$, and coefficients vectors $\beta_g$ are replaced by filters. In addition, differences exist among the use of $\ell_1$-norm, $\ell_2$-norm and $\ell_{2,1}$-norm as the penalty function~(Fig.~\ref{fig: reg}).

\subsubsection{Regularization on BN Parameters}
\label{REG-ON-BN}

Batch Normalization (BN) layers have been widely used in many modern CNNs to improve model generalization. By normalizing each training mini-batch, the internal covariate shift is addressed. The following equation describes the operations of a BN layer (NS~\cite{liuLearningEfficientConvolutional2017}),
\begin{equation}
    \label{eq:reg-bn}
    \hat{\mathbf{z}}=\frac{\mathbf{z}_\mathrm{in}-\mu_{\mathcal{B}}}{\sqrt{\sigma_{\mathcal{B}}^2+\epsilon}} ; \quad \mathbf{z}_{\mathrm{out}}=\gamma \hat{\mathbf{z}}+\beta
\end{equation}
where $N_{l+1}$-dimension $\mathbf{z}_{\mathrm{in}}$ and $\mathbf{z}_{\mathrm{out}}$ are the inputs and outputs of a BN layer, $\mu_{\mathcal{B}}$ and $\sigma_{\mathcal{B}}$ are the mean and standard deviation over the current mini-batch ${\mathcal{B}}$. $\epsilon$ is a small number to prevent division by zero. $\gamma \in \mathbb{R}^{N_{l+1}}$ and $\beta \in \mathbb{R}^{N_{l+1}}$ are the learnable parameters, indicating scale and shift, respectively. BN parameters are used as gates for filter pruning because the number of learnable parameters is equal to the number of feature maps and filters.

Network Slimming (NS)~\cite{liuLearningEfficientConvolutional2017} directly uses the scaling parameter $\gamma$ in BN to control output channels. The channel-level sparsity-induced $\ell_1$ regularization is then introduced to jointly train the weights together with the $\gamma$. After training, the corresponding channels that have close-to-zero $\gamma$ are pruned. To optimize the non-smooth $\ell_1$ penalty term, a subgradient descent method~\cite{shorMinimizationMethodsNonDifferentiable2012} is used.

Similar to the situation in NS~\cite{liuLearningEfficientConvolutional2017}, Gated Batch Normalization (GBN)~\cite{youGateDecoratorGlobal2019} uses $\gamma$ as the channel-wise gate and $\ell_1$-norm of $\gamma$ as the regularization term. A \textit{Tick-Tock} pruning framework is proposed to boost accuracy by iterative pruning. The \textit{Tick} phase trains the network with little data, and only the gates and final linear layer are allowed to be updated during training for one epoch. Meanwhile, channel importance is computed by the first-order Taylor Expansion for a global filter ranking. The \textit{Tock} phase then fine-tunes the sparse network with sparsity constraints.

Polarization Regularization (PR)~\cite{zhuangNeuronlevelStructuredPruning2020} provides a variant $\ell_1$-based regularizer to polarize the scaling factors $\gamma$. It contends that in most sparsity regularization methods, such as NS~\cite{liuLearningEfficientConvolutional2017}, a naive $\ell_1$ regularizer converges all scaling factors to zero indiscriminately. A more reasonable approach is to push the scaling factors of unimportant neurons to zero and those of important neurons to a large value. 
To achieve the polarization effect, another penalty term is added to the naive $\ell_1$ term, separating $\gamma$ from their mean as far as possible. Similar to NS~\cite{liuLearningEfficientConvolutional2017}, subgradients~\cite{shorMinimizationMethodsNonDifferentiable2012} are used on non-differentiable points to solve the non-smooth regularizer.

Rethinking Smaller-Norm-Less-Informative (RSNLI)~\cite{yeRethinkingSmallernormlessinformativeAssumption2018} is developed from the basis of previous methods tending to suffer from the Model Reparameterization problem and the Transform Invariance problem. As there are doubts whether smaller-norm parameters are less informative; Ye \textit{et al.} propose to train with ISTA~\cite{FastIterativeShrinkageThresholding} to enforce sparsity on $\gamma$. The channel with a $\gamma$ equal to zero is pruned. Then the $\gamma$-$W$ rescaling trick is used on $\gamma$ and weights to quickly start the sparsification process.

Operation-aware Soft Channel Pruning (SCP)~\cite{kangOperationAwareSoftChannel2020} considers both BN and ReLU operations. 
In contrast to 
NS~\cite{liuLearningEfficientConvolutional2017} making decision merely on the channel scaling $\gamma$, Kang \textit{et al.} also consider shifting parameters $\beta$.
Specifically, channels with large negative $\beta$ and large $\gamma$ are considered unimportant since these channels will become zero after ReLU.
To consider BN's large negative mean values, the cumulative distribution function (CDF) of a Gaussian distribution parameterized by $\beta$ and $\gamma$ is used as the indicator function. 
To optimize $\beta$ and $\gamma$, a sparsity-loss-inducing large CDF value is designed to encourage the network to be more sparse.

EagleEye~\cite{liEagleEyeFastSubnet2020} proposes a three-stage pipeline. \textbf{First}, pruning strategies (i.e., layer-wise pruning ratio) are generated by simple random sampling. \textbf{Second}, sub-networks are generated according to the pruning strategies and $\ell_1$-norm of filters. \textbf{Third}, an \textit{adaptive-BN-based candidate evaluation module} is used to evaluate the performance of the sub-networks. Li \textit{et al.} contend that outdated BN statistics are unfair to sub-network evaluations, and that BN statistics for each candidate should be re-calculated on a small part of the training dataset~\cite{liEagleEyeFastSubnet2020}. After evaluating sub-networks with the \textit{adaptive BN statistics}, the best-performing one is selected as the final pruned model.

\subsubsection{Regularization on Extra Parameter}
\label{REG-ON-EXTRA}
Although some studies~\cite{yeRethinkingSmallernormlessinformativeAssumption2018,youGateDecoratorGlobal2019} make special adjustments for networks without BN layers, introducing extra parameters is a more general solution. The extra parameters $\theta$ are trainable and parameterize the gates $g(\theta)$ in determining the pruning results. 

To find a sparse structure, Sparse Structure Selection (SSS)~\cite{huangDataDrivenSparseStructure2018} attempts to force the output of structures to zero. A scaling factor $\theta$ is introduced after each structure, i.e., neuron, group or residual block. When $\theta$ is lower than a threshold, the corresponding structure is removed. The gate $g(\theta)$ function is:
\begin{equation}
    \label{eq:SSS}
    g(\theta)=
    \begin{cases}
        0 , & \theta < \text{Threshold} \\
        1 , & \text{otherwise}          
    \end{cases}
    .
\end{equation}
It adopts a convex relaxation $\ell_1$-norm as the sparsity regularization on the extra parameter $\theta$. To update $\theta$, a modified Accelerated Proximal Gradient~\cite{parikhProximalAlgorithms2014} is used.

Generative Adversarial Learning (GAL)~\cite{linOptimalStructuredCNN2019} jointly prunes structures and adopts a Generative Adversarial Network (GAN) to achieve label-free learning. Extra scaling factors are introduced after each structure in the generator, forming soft masks. During training, a special regularization term that contains three regularizers is proposed: 1) a $\ell_2$ weight decay regularizer on generators, 2) a $\ell_1$ sparsity regularizer on the mask, and 3) an \textit{adversarial regularization} on discriminator. 
Furthermore, FISTA~\cite{FastIterativeShrinkageThresholding} is used to iteratively update the generator and discriminator, and the mask is updated together with the generator.

Discrete Model Compression (DMC)~\cite{gaoDiscreteModelCompression2020} explicitly introduces discrete (binary) gates after the feature map to precisely reflect the pruned channels' impact on the loss function. First, it samples subnetworks with stochastic discrete gates:
\begin{equation}
    \label{eq:DMC}
    g(\theta)=
    \begin{cases}
        1 , & \text{w.p. } \theta   \\
        0 , & \text{w.p. } 1-\theta
    \end{cases}
\end{equation}
where $\text{w.p.}$ stands for ``with probability''. The stochastic nature of the gate ensures that every channel has the chance to be sampled if $\theta \ne 0$, so different subnetworks can be produced. To update the non-differentiable binary gates, a Straight-Through Estimator~\cite{bengioEstimatingPropagatingGradients2013} is adopted. 

Similar to PR~\cite{zhuangNeuronlevelStructuredPruning2020}, Gates with Differentiable Polarization (GDP-Guo)~\cite{guoGDPStabilizedNeural2021} aims to polarize gates. Designing a gate with a polarization effect wields the property of smoothed $\ell_0$ formulation~\cite{xiangNewSmoothedL02019}:
\begin{equation}
    \label{eq:GDP-guo-property}
    g(\theta)= \frac{\theta^2}{\theta^2+\epsilon}
    \quad
    \xrightarrow{\text{has property}}
    \quad
    g(\theta)=
    \begin{cases}
        \approx 1 , & \theta \ne 0 \\
        = 0       , & \theta = 0
    \end{cases}
\end{equation}
where $\epsilon$ is a small positive value to prevent zero division. This property polarizes $\theta$ to either exactly $0$ or values close to $1$. The gate itself is differentiable. However, the sparsity regularization on $\theta$ involves $\ell_0$-norm and renders the objective function non-differentiable. Thus, proximal-SGD~\cite{nitandaStochasticProximalGradient2014} is used to update $\theta$.

Convolutional \textbf{Re-p}arameterization and Gradient \textbf{Res}etting (ResRep)~\cite{dingResRepLosslessCNN2021} re-parameterizes a CNN into two parts. The first ``remembering part'' learns to maintain the model's performance and will not be pruned. The second ``forgetting part'' inserts $1 \times 1$ CONV layers, or the \textit{compactors}, after the BN layers. During training, a modified SGD update rule updates the \textit{compactors} only. Thus, only \textit{compactors} are allowed to forget (forgetting) while other CONV layers are kept untouched (remembering).

Scientific Control Pruning (SCOP)~\cite{tangSCOPScientificControl2020} believes that the importance of a filter may be disturbed by \textit{potential factors} such as input data. For example, the filter importance ranking may vary if input data is slightly changed for data-dependent methods. To minimize the effect of \textit{potential factors}, it prunes under scientific control by creating knockoff counterparts~\cite{candesPanningGoldModelX2018}. Knockoff features are identical to the real features except for not knowing the true label. Two scaling factors $\theta$ and $\tilde{\theta}$ are then introduced to control the participation of real and knockoff features, respectively. The two parameters are complementary that $\theta + \tilde{\theta} = \mathbf{1}$. If $\theta$ cannot suppress $\tilde{\theta}$, the real features are deemed to have little or no association with the true output. Thus, the filter importance score is defined as $\mathcal{I} = \theta - \tilde{\theta}$, and filters with small importance scores are deemed redundant.

To directly control the model budget, Budget-Aware Regularization (BAR)~\cite{lemaireStructuredPruningNeural2019} uses a prior loss and introduces a learnable dropout parameter $\theta$~\cite{kingmaVariationalDropoutLocal2015}. The prior loss is the product of two functions. The first function is an approximation of budgets that is differentiable w.r.t. $\theta$. The second function is a variant log-barrier function~\cite{boydConvexOptimization2004} that employs a sigmoidal schedule. The novel objective function consists of the prior loss, enabling simultaneous training and pruning according to the budget. Knowledge distillation is then used to improve accuracy.

\subsubsection{Regularization on Filters}
\label{REG-ON-FILTER}

Structured Sparsity Learning (SSL)~\cite{wenLearningStructuredSparsity2016} uses Group Lasso to prune channels. Removing layer $l$'s channel will cause the removal of layer $l$'s filters and layer $l+1$'s input channels. Hence, it adds two separate regularization terms for filter-wise and channel-wise pruning:
\begin{equation}
    \label{eq:SSL}
    \sum_{n_l=1}^{N_{l+1}}\left\|\mathbf{W}_{n_l,:,:,:}^{(l)}\right\|_2 ,
    \qquad
    \sum_{c_l=1}^{N_l}\left\|\mathbf{W}_{:, c_l,:,:}^{(l)}\right\|_2 ,
\end{equation}
where $\mathbf{W}^{(l)} \in \mathbb{R}^{N_{l+1} \times N_l \times K_l \times K_l}$.

Out-In-Channel Sparsity Regularization (OICSR)~\cite{liOICSROutInChannelSparsity2019} uses Group Lasso to \textbf{jointly} regularize filters that work cooperatively. The regularization term is:
\begin{equation}
    \label{eq:OICSR}
    \sum_{i=1}^{N}\left\|\mathbf{W}_{i,:}^{(l)} \oplus \mathbf{W}_{:, i}^{(l+1)}\right\|_{2},
\end{equation}
where $\oplus$ denotes concatenation of the out-channel filters $\mathbf{W}^{(l)}_{i,:} \in \mathbb{R}^{N_{l+1} \times (N_l K_l K_l)}$ and the in-channel filters $\mathbf{W}^{(l+1)}_{:,i} \in \mathbb{R}^{(N_{l+2} K_{l+1} K_{l+1}) \times N_{l+1}}$. It uses the premise that out-channel filters of layer $l$ are interdependent of in-channel filters of layer $l+1$, so these filters should be regularized together.

Only Train Once (OTO)~\cite{chenOnlyTrainOnce2021} contends that even if all filter weights are zeros, the activation maps will be non-zero because of three parameters: 1) convolution bias, 2) BN mean, and 3) BN variance. Instead of grouping only filters, this method groups all parameters that cause a non-zero activation into a group named the zero-invariant group. Structured sparsity is then introduced to this group by applying the mixed $\ell_{1}/\ell_{2}$-norm. To solve the non-smooth mixed-norm regularization, a stochastic optimization algorithm named \textit{Half-Space Stochastic Projected Gradient} is used.

Growing Regularization (GREG)~\cite{wangNeuralPruningGrowing2022} exploits regularization under a growing penalty and uses two algorithms. The \textbf{first algorithm} focuses on the pruning schedule and adopts $\ell_1$-norm~\cite{liPruningFiltersEfficient2017} to obtain a mask for pruning. Instead of immediately removing unimportant filters, a growing $\ell_2$ penalty is used to gradually drive them to zero. 
The \textbf{second algorithm} uses a growing regularization to exploit the underlying Hessian information.
The authors observe that the weight discrepancy increases as the regularization parameter increases, and weights will naturally separate. If the discrepancy is large enough, even a simple $\ell_1$-norm can be an accurate criterion.

\subsection{Optimization Tools}
\label{OPTIMIZATION-TOOLS}

\red{
Optimization tools are integrated into the pruning process to find or induce structured sparsity in neural networks. 
For example, Taylor Expansion finds filter importance by approximating the loss function when a specific filter becomes zero. Variational Bayesian methods determine filter importance by exploiting the prior and posterior distributions. SGD-based methods modify the gradient update rule to detect and resolve redundant filters. ADMM-based methods impose structured sparsity constraints and find solutions by using the ADMM optimization algorithm. Bayesian Optimization helps reduce the ``curse of dimensionality problem''~\cite{koppen2000curse} encountered while learning the optimal sparse structures.
}

\subsubsection{Taylor Expansion}
\label{TAYLOR}
Taylor Expansion~\cite{vetterMatrixCalculusOperations1973} expands a function into the Taylor Series, which is an infinite sum of terms. The Taylor Expansion of a function $f(x)$ expands at some point $a$:
\begin{equation}
    \label{eq:TE}
    f(x) = \overbrace{\underbrace{f(a)+\frac{f^{\prime}(a)}{1 !}(x-a)}_{\text{first-order approximation}}+\frac{f^{\prime \prime}(a)}{2 !}(x-a)^2}^{\text{second-order approximation}}
    +\underbrace{R_2(x)}_{\text{remainders}},
\end{equation}
where $f'(\cdot)$ and $f''(\cdot)$ is the first-order derivative and second-order derivative with respect to $x$, respectively.

In structured pruning, Taylor Expansion is used to approximate the change in the loss $\Delta \mathcal{L}$ of pruning structures such as filters or channels. Since pruned weights are set to $0$, the loss function $\mathcal{L}(w)$ of weights $w$ can be evaluated at $a = 0$ using Taylor Expansion. By manipulating Eq.~\ref{eq:TE}, we can get:
\begin{align}
    \mathcal{L}(w) - \mathcal{L}(0) & = \frac{\mathcal{L}^{\prime}(0)}{1 !}(w-0) + \frac{\mathcal{L}^{\prime \prime}(0)}{2 !}(w-0)^2+R_2(w).
\end{align}
Let $\mathcal{L}(w) - \mathcal{L}(0) = \Delta \mathcal{L}$ be the change in the loss of removing some weights, and $w - 0 = \Delta w$ be the change in the weights. We can get the following equation based on~\cite{wangEigenDamageStructuredPruning2019}, 
\begin{equation}
    \label{eq:TE-pruning}
    \Delta \mathcal{L} = \frac{\partial \mathcal{L}^{\top}}{\partial w} \Delta w + \frac{1}{2} \Delta w^{\top} \mathbf{H} \Delta w + R_2(w)
\end{equation}
where $\frac{\partial \mathcal{L}}{\partial w}$ is the first-order gradient of loss function w.r.t. weights, and $\mathbf{H}$ is the Hessian matrix containing second-order derivatives. Compared to regularization-based methods, pruning with Taylor Expansion does not need to wait until the activations are trained sufficiently small~\cite{molchanovImportanceEstimationNeural2019}.

First-order and second-order Taylor expansions have their own characteristics. The second-order expansion contains more information, but it requires calculating the second-degree derivatives that are computationally prohibitive. On the contrary, the first-order expansion can be obtained from backpropagation without requiring additional memory, but this provides less information.

\textbf{First-order-Taylor:}
\label{FIRST-ORDER}
Mol-16~\cite{molchanovPruningConvolutionalNeural2017} uses the first-order information to estimate the change in the loss of pruning activation maps. The higher-order remainders, including the second-order term, are discarded since they are computationally intractable and are encouraged to be small by the widely-used ReLU activation function. Thus, the absolute change in loss approximated by the first-order term is used as the metric for feature map importance:
\begin{equation}
    \left| \frac{1}{M} \sum_{m} \frac{\partial \mathcal{L}}{\partial z_m} z_m \right|,
\end{equation}
where $M$ is the length of a flattened feature map, and $z$ is an activation in a feature map.
After determining the feature map importance, the lowest-ranked maps will be pruned. 

Compared to Mol-16~\cite{molchanovPruningConvolutionalNeural2017}, which fails with skip connections, Mol-19~\cite{molchanovImportanceEstimationNeural2019} proposes a more general method that uses Taylor expansion to approximate the squared change in the final loss. Unlike Mol-16's~\cite{molchanovPruningConvolutionalNeural2017} use of activations that increase memory consumption, Mol-19~\cite{molchanovImportanceEstimationNeural2019} computes the importance $\mathcal{I}$ based on weights:
\begin{equation}
    \label{eq:mol19}
    \mathcal{I}_{\mathcal{S}}(\mathbf{W})
    \triangleq \sum_{s \in \mathcal{S}} \left(\frac{\partial \mathcal{L}}{\partial w_s} w_s\right)^2
    ,
\end{equation}
where $\mathcal{S}$ is a structural set of parameters such as a convolutional filter, $w_s$ is the individual weight of the filter, and $\frac{\partial \mathcal{L}}{\partial w_s}$ represents the gradient.
The first-order expansion performs significantly faster than the second-order expansion with a slightly higher accuracy drop. 

Due to the simplicity and efficiency of computing, the first-order expansion is widely adopted by many methods such as GBN~\cite{youGateDecoratorGlobal2019} and GDP-Lin~\cite{linAcceleratingConvolutionalNetworks2018} which are discussed in other sections.

\textbf{Second-order-Taylor:}
\label{SECOND-ORDER}
In this sub-section, the Hessian matrix $\mathbf{H}$ that contains second-order information from Eq.~\ref{eq:TE-pruning} is exploited. First, pioneering unstructured studies are briefly introduced. Second, the structured pruning methods are addressed.

\textit{Pioneering unstructured studies:} Most current structured pruning methods that use second-order Taylor expansion are based on two pioneering studies on unstructured pruning: Optimal Brain Damage (OBD)~\cite{lecunOptimalBrainDamage1989} and Optimal Brain Surgeon (OBS)~\cite{hassibiSecondOrderDerivatives1992}. OBD~\cite{lecunOptimalBrainDamage1989} assumes that $\mathbf{H}$ is diagonal to ease computation. The diagonal $\mathbf{H}$ is then used to compute the parameter importance. However, OBS~\cite{hassibiSecondOrderDerivatives1992} finds that most of the Hessian matrices $\mathbf{H}$ are strongly non-diagonal. Thus, the full $\mathbf{H}$ is used and the parameter importance is calculated by $\mathbf{H}^{-1}$.

\textit{Structured pruning methods:} With the success of OBD~\cite{lecunOptimalBrainDamage1989} and OBS~\cite{hassibiSecondOrderDerivatives1992} in unstructured pruning, the second-order expansion is applied to structured pruning. As deep CNNs have millions of parameters, computing and storing the Hessian $\mathbf{H}$ become challenging~\cite{pengCollaborativeChannelPruning2019}. Recent methods aim to approximate the Hessian matrix for structured pruning.

Collaborative Channel Pruning (CCP)~\cite{pengCollaborativeChannelPruning2019} approximates the Hessian matrix by only using the first-order derivative of a pre-trained model. The first-order information can be retrieved from backpropagation, and no additional storage is needed. In addition, Peng \textit{et al.} exploit the effect of removing multiple channels instead of a single channel. The non-diagonal element in $\mathbf{H}$ reflects the interaction between two channels and hence exploits the inter-channel dependency. CCP models the channel selection problem as a constrained 0-1 quadratic optimization problem to evaluate the joint impact of pruned and unpruned channels.

Eigen Damage (ED)~\cite{wangEigenDamageStructuredPruning2019} introduces a baseline method that is the naive structured extension of OBD~\cite{lecunOptimalBrainDamage1989} and OBS~\cite{hassibiSecondOrderDerivatives1992}. Two algorithms are then applied to improve the baseline method. The \textbf{baseline method} sums up the individual change in parameters over a filter, raising the granularity to the filter level. However, computing and storing the Hessian $\mathbf{H}$ is intractable. Wang \textit{et al.} propose the use of the \textbf{first algorithm}~\cite{wangEigenDamageStructuredPruning2019} that applies K-FAC~\cite{martensOptimizingNeuralNetworks2015} approximation to decompose filters. As the naive extensions and the first algorithm both fail at capturing the correlation between filters, a \textbf{second algorithm} which decorrelates the weights before pruning is applied. This second algorithm adopts K-FAC~\cite{martensOptimizingNeuralNetworks2015} and projects weight space to a Kronecker-Factored eigenspace (KFE)~\cite{NEURIPS2018_48000647} where there is little correlation.

Group Fisher Pruning (GFP)~\cite{liuGroupFisherPruning2021} addresses the difficulties faced by other pruning methods when channels from multiple layers are coupled and require simultaneous pruning.
First, a layer grouping algorithm is used to automatically identify coupled channels. Second, the Hessian information is used as a unified importance criterion of a single channel and coupled channels. With the help of the Fisher information, the Hessian matrix is transformed into the square of first-order derivatives.

\subsubsection{Variational Bayesian}
\label{BAYESIAN}
Bayesian inference~\cite{fienberg2006did} is a method to infer the posterior probability distribution $p(\theta | x)$ with the known prior probability distribution $p(\theta)$ of parameters $\theta$ and the observed data $x$. The formula to compute the posterior distribution over $\theta$ is given:
\begin{equation}
    \overbrace{p(\theta | x)}^{\text{Posterior}}=\frac{\overbrace{p(x | \theta)}^{\text{Likelihood}} \cdot \overbrace{p(\theta)}^{\text{Piror}}}{\underbrace{p(x)}_{\text{Evidence}}}, \text{where } p(x)=\int p(x | \theta) p(\theta) d \theta.
\end{equation}
However, computing the evidence often requires computationally intractable integrals when large amounts of data are involved. Variational Bayesian (VB) methods~\cite{fox2012tutorial} are used to approximate the posterior distribution $p(\theta | x)$ by a variational distribution $q(\theta)$. Specifically, $q(\theta)$ is optimized by minimizing the Kullback–Leibler (KL) divergence which measures the ``similarity" between $q(\theta)$ and $p(\theta | x)$. Since the computation of KL-divergence involves the intractable posterior distributed $p(\theta | x)$, the optimization problem is solved by equivalently converting it to maximize the Evidence Lower BOund (ELBO).

Variational Pruning (VP)~\cite{zhaoVariationalConvolutionalNeural2019} is based on channel importance being indicated by random variables, as pruning by deterministic channel importance is inherently improper and unstable. Thus, BN's parameter $\gamma$ is used to indicate the channel saliency and model $\gamma$ by Gaussian distribution $\mathcal{N}\left(\mu, \sigma\right)$.
To introduce sparsity, VP utilizes the centrality property of the Gaussian distribution $\mathcal{N}$ and samples $\gamma$ from $\mathcal{N}\left(\mu =0, \sigma\right)$ as the sparse prior distribution.
After optimizing ELBO, distributions of $\gamma$ with close-to-zero mean and small variance are considered safe for pruning, since such distributions are less likely to have salient parameters.

To find redundant channels, Recursive Bayesian Pruning (RBP)~\cite{zhouAccelerateCNNRecursive2019} targets the posterior of redundancy, which assumes an inter-layer dependency among channels. First, each input channel is scaled by a dropout noise $\theta$~\cite{kingmaVariationalDropoutLocal2015, louizos2017bayesian} with a dropout rate $r$. Second, the dropout noises are then modeled across layers as a Markov chain to exploit the inter-layer dependency among channels. To get the posterior of redundancy, a sparsity-inducing Dirac-like prior is chosen. In addition, RBP~\cite{zhouAccelerateCNNRecursive2019} adopts the reparameterization trick~\cite{kingmaVariationalDropoutLocal2015} to scale the noise $\theta$ on the corresponding channels to consider data fitness. As a result, the dropout rate $r$ can be updated along with weights in a gradient-based manner. Pruning is conducted by setting the corresponding channels to zero when $r$ is greater than a threshold.

A few other studies also use the Bayesian point of view. VIBNet~\cite{dai2018compressing} uses a \textit{variational information bottleneck}, which is the informational theoretical measure of redundancy between adjacent layers. Louizos \textit{et al.}~\cite{louizos2017bayesian} utilize the horseshoe prior to efficiently approximate channel redundancy. Neklyudov \textit{et al.}~\cite{neklyudov2017structured} use a log-normal prior, resulting in a tractable and interpretable log-normal posterior.

\subsubsection{Others}
\label{OPTIMIZATION-OTHER}

\textbf{SGD-based:}
Instead of zeroing out filters, Centripetal SGD (C-SGD)~\cite{dingCentripetalSGDPruning2019} makes redundant filters identical and merges identical filters into one filter.
Regularized Modernized Dual Averaging (RMDA)~\cite{huangTrainingStructuredNeural2022} adds momentum to the RDA algorithm~\cite{xiaoDualAveragingMethod2009} and ensures that the trained model has the same structure as the original model. To prune the model, RMDA adopts Group Lasso to promote structured sparsity.

\textbf{ADMM-based:}
\label{ADMM-BASED}
Alternating Direction Method of Multipliers (ADMM)~\cite{boydDistributedOptimizationStatistical2011} is an optimization algorithm used to decompose the initial problem into two smaller, more tractable subproblems.
StructADMM~\cite{zhangStructADMMAchievingUltrahigh2022} studies the solution of different types of structured sparsity such as filter-wise and shape-wise.
Zhang \textit{et al.} use a progressive and multi-step ADMM framework: at each step, it uses ADMM to prune and masks out zero weights, leaving the remaining weights as the optimization space for the next step. 

\textbf{Bayesian Optimization:}
Bayesian optimization (BO)~\cite{frazier2018tutorial} is a sequential design strategy for the global optimization of black-box functions that does not assume any functional forms. 
When pruning a model that involves an extremely large design space, the curse of dimensionality problem occurs~\cite{srinivasGaussian2010}.
Rollback~\cite{fanBayesianOptimizationClustering2022} adopts the RL-style automatic channel pruning~\cite{heAMCAutoMLModel2018} and uses BO to determine the optimal pruning policy. 

\textbf{Soft Threshold:} 
Soft threshold ~\cite{donoho1995noising} is a de-noising operator in the area of digital signal processing.
Soft Threshold Reparameterization (STR)~\cite{kusupatiSoftThresholdWeight2020} adopts this operator to learn a non-uniform sparsity budget which is optimized per layer.

\subsection{Dynamic Pruning}
\label{DYNAMIC}
\red{
Structured pruning can be conducted in a dynamic manner during both training and inference. 
}
\textbf{Dynamic during training} aims to preserve the model's representative capacity by maintaining a dynamic pruning mask during training. It is also called soft pruning to ensure that improper pruning decisions can be recovered later. On the other hand, hard pruning permanently removes weights with a fixed mask. \textbf{Dynamic during inference} indicates the networks are pruned dynamically according to different input samples. For instance, a simple image that contains clear targets requires less model capacity compared to a complex image~\cite{huangTrainingStructuredNeural2022}. Hence, dynamic inference provides better resource-accuracy trade-offs.

\subsubsection{Dynamic during Training}
\label{DYNAMIC-TRAINING}
\textbf{Weight-level dynamic:} The concept of training-time dynamic pruning was first introduced in Dynamic Network Surgery (DNS)~\cite{guoDynamicNetworkSurgery2016}, an unstructured pruning approach. To clarify the difference between other approaches, its formulation is:
\begin{equation}
    \label{eqn: dynamic 1}
    \mathbf{W}^{l}_{i,j} \leftarrow \mathbf{W}^{l}_{i,j}-\eta\frac{\partial \mathcal{L}(\mathbf{W}^{l}\odot \mathbf{T}^{l})}{\partial(\mathbf{W}^{l}_{i,j} \mathbf{T}^{l}_{i,j})}
    ,
\end{equation}
where $\eta$ is the learning rate and $\odot$ denotes the Hadamard Product operator. $\mathbf{W}^{l}$ is the weight matrix in the $l$-th layer with all weights of kernels unfolded and concatenated together. $\mathbf{T}^{l}$ is the binary mask that has the same shape as the weight matrix, indicating the importance of the weights. $\mathcal{L(\cdot)}$ is the loss function. $\mathbf{W}_{i,j}^{l}$ and $\mathbf{T}_{i,j}^{l}$ indicate a single element in $\mathbf{W}^{l}$ and $\mathbf{T}^{l}$. $\mathbf{W}$ and $\mathbf{T}$ are updated alternatively. All the weights are updated, so wrongly pruned parameters have a chance to regrow.

\textbf{Filter-level dynamic:} Soft filter Pruning (SFP)~\cite{heSoftFilterPruning2018} adopts the idea of dynamic pruning in a structured way. Its use is based on the premise that hard pruning that uses a fixed mask throughout the training will reduce the optimization space. Therefore, it dynamically generates the masks based on the $\ell_2$-norm of filters at every epoch. Soft pruning means setting the values of filters to zero instead of removing filters. Previously soft-pruned filters are allowed to be updated at the next epoch, during which masks will be reformed based on new weights. The update rule is:
\begin{equation}
    \mathbf{W}^l \leftarrow \mathbf{W}^l-\eta \frac{\partial \mathcal{L}\left(\mathbf{W}^l \odot \mathbf{m}^l\right)}{\partial\left(\mathbf{W}^l \odot \mathbf{m}^l\right)}
    ,
\end{equation}
where $\mathbf{W}^{l}\in\mathbb{R}^{N_{l+1}\times N_lKK} \text{~and~}\mathbf{m}^l\in\mathbb{R}^{N_{l+1}}$.

Globally Dynamic Pruning (GDP-Lin)~\cite{linAcceleratingConvolutionalNetworks2018} also maintains a binary dynamic mask during training based on the filter importance. GDP-Lin adopts the first-order Taylor expansion to approximate the global discriminative power of each filter. In addition, the authors argue that frequently changing masks cannot effectively guide the pruning. Hence, masks are updated every $e$ iteration, where $e$ is set as a decreasing value to accelerate the convergence.

In addition to maintaining a sparse dynamic mask, Dynamic Pruning with Feedback (DPF)~\cite{linDynamicModelPruning2022} simultaneously maintains a dense model. The premise behind its use is that a sparse model can be considered a dense model with compression error. The error can be used as feedback to ensure the correct direction in the gradient updates. For this goal, gradients computed from the sparse model are used to update the dense model. The advantage of applying gradients on a dense model is that it helps weights to recover from errors.

CHannel EXploration (CHEX)~\cite{houCHEXCHannelEXploration2022} uses two processes to dynamically adjust filter importance. The first process is the channel pruning process with \textit{Column Subset Selection criterion}~\cite{guEfficientAlgorithmsComputing1996}. The second process is the regrowing process, which is based on \textit{orthogonal projection}~\cite{hornMatrixAnalysis2012} to avoid regrowing redundant channels and to explore channel diversity. Regrowing channels are restored to the \textit{most recently used (MRU)} parameters rather than zeros. To better preserve model accuracy,  the remaining channels are re-distributed dynamically among all the layers. 

Dynamic Sparse Graph (DSG)~\cite{liuDynamicSparseGraph2019} activates a small number of critical neurons with the constructed sparse graph at every iteration dynamically. DSG is developed from the argument that direct computation according to output activations is very costly for finding critical neurons. The \textit{dimension reduction search} is formulated to forecast the activation output by computing input and filters in a lower dimension. Furthermore, to prevent BN layers from damaging sparsity, Liu \textit{et al.} introduce the \textit{double-mask selection} that uses the same selection mask before and after BN layers~\cite{liuDynamicSparseGraph2019}.

Other methods, such as SCP~\cite{kangOperationAwareSoftChannel2020} and DMC~\cite{gaoDiscreteModelCompression2020}, also maintain the mask dynamically. These are discussed in Section~\ref{REGULARIZATION}.

\subsubsection{Dynamic during Inference}
\label{DYNAMIC-INFERENCE}
Runtime Neural Pruning (RNP)~\cite{linRuntimeNeuralPruning2017} is based on the premise that the static model fails to exploit the different properties of input images by using the same weights for both easily recognized and complex pattern images. It uses a framework with CNN as the backbone and RNN as a decision network. The network pruning is modeled as a Markov decision process, and models are trained by reinforcement learning (RL). The RL agent evaluates filter importance and performs channel-wise pruning according to the relative difficulty of samples for a task. Thus, when the image is easier for the task, a sparser network is generated.

Feature Boosting and Suppression (FBS)~\cite{gaoDynamicChannelPruning2019} predicts output channel importance based on inputs by introducing a tiny and differentiable auxiliary network. In the auxiliary network, a 2D global average pooling is used to subsample the activation channels in the previous layer to scalars. The advantage of this is that it reduces the computational overhead in the auxiliary network. A salient predictor then uses the subsampled scalars to generate the predicted saliency scores through a fully connected layer followed by an activation function. The top-$k$ salient channels are finally involved in the inference-time calculation.

Deep Reinforcement Learning pruning (DRLP)~\cite{chenStorageEfficientDynamic2020} learns both the runtime (dynamic) and the static importance of channels. The runtime importance measures the importance of channels specific to an input. In contrast, the static importance measures the channel importance for the whole dataset. Like RNP~\cite{linRuntimeNeuralPruning2017}, DRLP uses reinforcement learning (RL). The RL framework contains two parts: one static and one runtime. Each part contains an importance predictor and an agent, providing static/runtime importance and layer-wise pruning ratio, respectively. In addition, a \textit{trade-off pruner} generates a unified pruning decision based on the outputs from the two parts.

Dynamic Dual Gating (DDG)~\cite{liDynamicDualGating2021} uses two separate gating modules (spatial and channel gating) to determine inference-time importance according to inputs. The spatial module consists of an adaptive average pooling followed by a 3x3 CONV layer to extract informative spatial features. By having a spatial mask, only informative features are allowed to pass to the next layer. The channel gating module works in a manner similar to existing methods such as FBS~\cite{gaoDynamicChannelPruning2019} by using global average pooling followed by the application of FC layers. To enable gradient flow, the Gumbel-Softmax reparameterization~\cite{jangCategoricalReparameterizationGumbelSoftmax2022} is used for both spatial and channel gating modules.

Fire Together Wire Together (FTWT)~\cite{elkerdawyFireTogetherWire2022} models the dynamic pruning task as a self-supervised binary classification problem. The proposed framework uses 1) a prediction head to generate learnable binary masks and 2) a ground truth mask to guide the learning after each convolutional layer. The prediction head consists of a global max pooling layer, a 1x1 CONV layer, and a softmax layer. Then, the outputs from the prediction head are rounded to generate a binary mask. To achieve self-supervised learning, ground truth binary masks are needed; by ranking the norm of input activations, ground truth pruning decisions are made to guide the learning.

At the same time, Contrastive Dual Gating (CDG)~\cite{mengContrastiveDualGating2022} implements another self-supervised dynamic pruning framework by using contrastive learning~\cite{hadsellDimensionalityReductionLearning2006}.
Contrastive learning trains a model from the latent contrastiveness of high-level features of two contrastive branches.
Therefore, a similarity-based contrastive loss~\cite{oordRepresentationLearningContrastive2019} is used for gradient-based learning without the labeled data. Empirical studies show that pruning decisions cannot be transferred between contrastive branches. Thus, dual gating is introduced, and the two branches have distinct pruning masks. CDG~\cite{mengContrastiveDualGating2022} uses a framework similar to that of CGNet~\cite{huaChannelGatingNeural2019} for each contrastive branch. Like DDG~\cite{liDynamicDualGating2021}, a 3x3 CONV layer is used to extract spatial features for unpruned features.

\subsection{\red{NAS-Based Pruning}}
\label{NEURAL-ARCHITECTURE-SEARCH}

Given that it is cumbersome to manually determine pruning-related hyperparameters such as the layer-wise pruning ratio, NAS-based Pruning has been proposed to automatically find pruned structures.
Based on the survey of Neural Architecture Search
(NAS)~\cite{elskenNeuralArchitectureSearch}, we categorize NAS for pruning into three methods. NAS can be modeled as:
\begin{enumerate}[leftmargin=*]
    \item \textbf{reinforcement learning (RL) problems}, in which RL agents find sparse subnetworks by searching over the action space such as pruning ratios.
    \item \textbf{gradient-based methods} that modify the gradient update rule to make optimization problems with sparsity constraints differentiable to weights. 
    \item \textbf{evolutionary methods} that adopt evolutionary algorithms to explore and search the sparse subnetworks.
\end{enumerate}

\subsubsection{Reinforcement Learning-Based}
\label{REINFORCEMENT-LEARNING-BASED}
AutoML for Model Compression (AMC)~\cite{heAMCAutoMLModel2018} uses RL to automatically select the appropriate layer-wise pruning ratio without manual sensitivity analysis. To search over a continuous action space, the deep deterministic policy gradient (DDPG) algorithm~\cite{lillicrapContinuousControlDeep2016} is adopted. The actor receives 11 layer-dependent states such as FLOPS and outputs a continuous action about the pruning ratio. For accuracy-guaranteed pruning, the reward for the DDPG agent is modified from Eq.~\ref{eq:reward} to Eq.~\ref{eq:reward-flops}:
\begin{equation}
    \label{eq:reward}
    R_{\mathrm{err}} = - Error
\end{equation}
\begin{equation}
    \label{eq:reward-flops}
    \begin{aligned}
         & R_{\mathrm{FLOPs}}=-Error \cdot \log (\mathrm{ FLOPs })    \\
         & R_{\mathrm{Param}}=-Error \cdot \log (\# \mathrm{ Param })
    \end{aligned}
\end{equation}

Automatic Graph encoder-decoder Model Compression (AGMC)~\cite{yuAutoGraphEncoderDecoder2021} adopts the premise that AMC~\cite{heAMCAutoMLModel2018}  still requires manual selection of the fixed 11 states fed into the RL agent. In addition, the fixed environment states ignore the rich information within the computational graphs. Hence, the DNNs are modeled as computational graphs and fed into a GCN-based graph encoder-decoder~\cite{kipfSemiSupervisedClassificationGraph2017} to automatically learn the RL agent's input states. In contrast to AMC~\cite{heAMCAutoMLModel2018}, AGMC rescales the pruning ratios to compensate for any unmet sparsity constraint rather than applying the FLOPs into the reward function (Eq.~\ref{eq:reward-flops}).

DECORE~\cite{alwaniDECOREDeepCompression2022} adopts multi-agent learning. The presence of multiple agents comes from assigning dedicated agents to all channels in a layer, with each agent only learning one parameter representing a binary decision. In contrast to AMC~\cite{heAMCAutoMLModel2018} and AGMC~\cite{yuAutoGraphEncoderDecoder2021}, DECORE models the state of the channel itself. To learn the action parameters, a higher reward is given when the accuracy retention task and compression task are both well-completed:
\begin{equation}
    \label{eq:DECORE}
    \begin{gathered}
        R^l_\mathrm{comp.} =\sum_{j=1}^{N_{l+1}} 1-a^l_j, \quad
        R_\mathrm{a c c}=\left\{\begin{array}{cc}
            1        & \text { if } \hat{y}=y \\
            -\lambda & \text { else }
        \end{array}\right. \\[2pt]
        R^l=R^l_\mathrm{comp.} * R_\mathrm{a c c}
    \end{gathered}
\end{equation}
where $N_{l+1}$ represents the output channel size of layer $l$, $a^l$ is the action vector of layer $l$, $y$ and $\hat{y}$ are the true label and predicted label, and $-\lambda$ is a large penalty for a wrong prediction. The reward function does not consider the FLOPs; instead, the reward for the compression task simply takes into account the number of channels remaining. 

GNN-RL~\cite{yuTopologyAwareNetworkPruning2022} first models the DNN into a multi-stage graph neural network (GNN) to learn the global topology. The generated hierarchical computational graph is then used as the environmental state of the agent. In contrast to AMC~\cite{heAMCAutoMLModel2018} and AGMC~\cite{yuAutoGraphEncoderDecoder2021} that both use DDPG, this method uses the Proximal Policy Optimization (PPO) algorithm~\cite{schulmanProximalPolicyOptimization2017} as the policy since PPO gives a much better experiment result. In addition, due to the specially designed graph environment, the reward for the RL system does not need to contain sparsity constraints. Accuracy is the only metric for the reward described by Eq.~\ref{eq:reward}.

\subsubsection{Gradient-Based}
\label{GRADIENT-BASED}
To search for the layer-wise sparsity, Differentiable Markov Channel Pruning (DMCP)~\cite{guoDMCPDifferentiableMarkov2020} models channel pruning as a Markov process. Here, the state means retaining the channel during pruning, and transitions between states represent the pruning process. DMCP parameterizes both the transition and budget loss by learnable architecture parameters. The final loss is thus differentiable w.r.t. architecture parameters. Two stages are used to finish pruning in a gradient-based manner. The first stage updates weights of unpruned networks with a variant sandwich rule~\cite{yuUniversallySlimmableNetworks2019}. The second stage updates the architecture parameters.

Differentiable Sparsity Allocation (DSA)~\cite{ningDSAMoreEfficient2020} optimizes the sparsity allocation in continuous space using a gradient-based method that is more effective than a discrete search. The validation loss is adopted as a differentiable surrogate to render the evaluation of validation accuracy differentiable. Next, a probabilistic differentiable pruning process is used as a replacement for the non-differentiable hard pruning process. Finally, sparsity allocation is obtained under a budget constraint. An ADMM-inspired optimization method~\cite{boydDistributedOptimizationStatistical2011} is used to solve the constrained non-convex optimization problem.

Differentiable Hyper Pruning (DHP)~\cite{liDHPDifferentiableMeta2020} uses hypernetworks that generate weights for backbone networks. The designed hypernetwork consists of three layers. \textbf{1) The latent layer} first takes two learnable latent vectors $\{\mathbf{z}^l \text{ and } \mathbf{z}^{l-1}, \mathbf{z}^l \in \mathbb{R}^{N_{l+1}}\}$ and forms the latent matrix. Then \textbf{2) the embedding layer} projects the latent matrix to embedding space, forming embedding vectors. Finally, \textbf{3) the explicit layer} takes embedding vectors and outputs weights that can be explicitly used as weights for CONV layers. Pruning is done by regularizing the latent vectors with $\ell_1$ sparsity regularization, and proximal gradient algorithms~\cite{boydConvexOptimization2004} are used to update the latent vectors. This method is approximately differentiable since the proximal operation step has a closed-form solution.

Learning Filter Pruning Criteria (LFPC)~\cite{heLearningFilterPruning2020} searches for layer-wise pruning criteria instead of pruning ratios. The basis for its use is that filters have differing distributions for extracting coarse and fine-level features. This makes using the same pruning criterion for all layers inappropriate. To explore the criteria space, criteria parameters are introduced to guide the choice of criteria. To enable gradient flow, this method uses the Gumbel-Softmax reparameterization~\cite{jangCategoricalReparameterizationGumbelSoftmax2022} to make the loss differentiable to criteria parameters.

Transformable Architecture Search (TAS)~\cite{dongNetworkPruningTransformable2019} searches the width and depth of a network. Two learnable parameters, $\alpha$ and $\beta$, are introduced to indicate the distribution of the possible number of channels and layers, respectively. Subnetworks can be sampled based on $\alpha$ and $\beta$. Applying the Gumbel-Softmax reparameterization~\cite{jangCategoricalReparameterizationGumbelSoftmax2022} makes the sampling process differentiable. The search objectives are minimizing the validation loss while encouraging smaller computational costs and penalizing the unmet resources budget. Knowledge Distillation is further used to boost the performance of the pruned networks.

Exploration and Estimation (EE)~\cite{zhangExplorationEstimationModel2021} achieves channel pruning in a two-step gradient-based manner. The first step is \textbf{exploration} of subnetworks to allow larger search space with a fast sampling technique (\textit{Stochastic Gradient Hamiltonian Monte Carlo}~\cite{chen2014stochastic}). To bring sparsity, this method applies a FLOPs-aware prior distribution to the exploration process. The second step, \textbf{estimation}, is used to guide the generation of high-quality subnetworks.

\subsubsection{Evolutionary-Based}
\label{EVOLUTIONARY-BASED}

MetaPruning~\cite{liuMetaPruningMetaLearning2019} aims to find the optimal layer-wise channel number with a two-stage framework. The first stage trains a meta network named \textit{PruningNet} to generate various structures. \textit{PruningNet} takes randomly sampled encoding vectors representing structures and learns in an end-to-end manner. In the second stage, an evolutionary search algorithm is deployed to search for the optimal structures under constraints. No fine-tuning is needed at search time since \textit{PruningNet} predicts the weights for all the pruned nets.

In contrast to MetaPruning~\cite{liuMetaPruningMetaLearning2019}, ABCPruner~\cite{linChannelPruningAutomatic2020} looks for the optimal layer-wise channel number with a one-stage approach and requires no extra supporting network. In addition, it drastically reduces the combinations of pruned structures by limiting the preserved channels to a given space. To search for the optimal pruned structures, an evolutionary algorithm based on the \textit{Artificial Bee Colony algorithm (ABC)}~\cite{karabogaIDEABASEDHONEY} is applied.

Instead of using evolutionary algorithms (EA) on the entire network, Cooperative CoEvolution algorithm for Pruning (CCEP)~\cite{shangNeuralNetworkPruning2022} uses \textit{cooperative coevolution}~\cite{potter2000cooperative} that groups the network by layer and applies EA on each layer. By decomposing the networks into multiple groups, the search space is drastically decreased. The overall framework uses an iterative prune-and-fine-tune strategy. In each iteration, multiple individual candidates (parents) are generated by randomly pruning filters. Offsprings are generated by conducting bit-wise mutations on parents, where each bit represents the presence of a filter. Individuals are then evaluated with accuracy and FLOPs, and only top-$k$ are retained for the next iteration.

\subsection{Extensions}
\label{COMPRESSION-EXTENSIONS}

\red{
Structured pruning can be extended with the Lottery Ticket Hypothesis, with orthogonal network compression techniques and with different granularity of structured sparsity. 
}

\subsubsection{Lottery Ticket Hypothesis}
\label{LTH}
The Lottery Ticket Hypothesis (LTH)~\cite{frankleLotteryTicketHypothesis2019} claims that ``dense, randomly-initialized, feed-forward networks contain subnetworks (winning tickets) that — when trained in isolation — reach test accuracy comparable to the original network in a similar number of iterations." 
\textit{Weight rewinding} is often used in LTH-based papers.
In the study by Frankle \& Carbin~\cite{frankleLotteryTicketHypothesis2019}, the weights and the learning rate schedule are rewound to the values at epoch $k = 0$, in order to find the sparse subnetworks at initialization. Frankle~\textit{et al.}~\cite{frankle2020linear} propose to rewind the weights and learning rate schedule to $k > 0$ epoch to deal with SGD randomness (noise). 
Different means of achieving unstructured LTH have been proposed:
SNIP~\cite{LeeSNIPSingleshotNetwork2019} preserves the training loss;
GraSP~\cite{wangPickingWinningTickets2019} preserves the \textit{gradient flow};
Neural Tangent Kernel (NTK)~\cite{jacotNeuralTangentKernel2018} captures the \textit{training dynamics};
GF~\cite{lubanaGradientFlowFramework2020} prunes weights that cause the least change to the \textit{gradient flow}.
 
Renda \textit{et al.}~\cite{rendaComparingRewindingfinetuning2020} propose to rewind the learning rate schedule but not the weight value (\textit{learning rate rewinding}). Experiments show that rewinding techniques consistently outperform fine-tuning, in which \textit{learning rate rewinding} outperforms or matches \textit{weight rewinding} in all scenarios.

Rethinking the Value of Network Pruning (RVNP)~\cite{liuRethinkingValueNetwork2019} re-evaluates the value of network pruning. The argument is that traditional fine-tuning techniques work no better than pruning from scratch. Extending LTH in a structured pruning setting for large-scale datasets fails to yield the lottery ticket. 

Early Bird (EB)~\cite{youDrawingEarlyBirdTickets2020} proposes to find the winning ticket in the early stages of training, rather than fully training the dense network.
To find the EB ticket, the mask for the filters is computed based on the Hamming distance between two subnetworks pruned from the same model.

Prospect Pruning (ProsPr)~\cite{alizadehProspectPruningFinding2022} argues that the \textit{trainability} of the pruned networks should be considered. \textit{Trainability} exists as the model is going to be trained after pruning.
Thus, \textit{meta-gradients} (gradient-of-gradients) have been proposed as a measure of \textit{trainability}. 
Instead of estimating the changes in the loss at initialization, the effect of pruning on the loss over several steps of gradient descent at the beginning of training is estimated in ProsPr.

Early Compression via Gradient Flow Preservation (EarlyCroP)~\cite{rachwanWinningLotteryAhead2022} achieves early pruning by solving three key problems. The problem \textbf{1) why to prune} is solved by extending a GF-based pruning criterion~\cite{lubanaGradientFlowFramework2020} to structured pruning. \textbf{2) How to prune} is solved by preserving the training dynamics using a connection between the \textit{gradient flow} and NTK~\cite{jacotNeuralTangentKernel2018}. Lastly, \textbf{3) when to prune} is solved by finding the \textit{lazy kernel regime}, which is the phase where pruning has little effect on the \textit{training dynamics}.  

Pruning-aware Training (PaT)~\cite{shenWhenPrunePolicy2022} tries to resolve the problem of \textit{when early pruning should begin}? PaT is based on the premise that sub-networks having the same number of remaining neurons can have very different architectures. To determine when the architectures are stable, a novel metric called the \textit{early pruning indicator (EPI)} that computes the structure similarity of two subnetworks has been proposed.

\begin{figure*}[h]
    \centering
    \includegraphics[width=\textwidth]{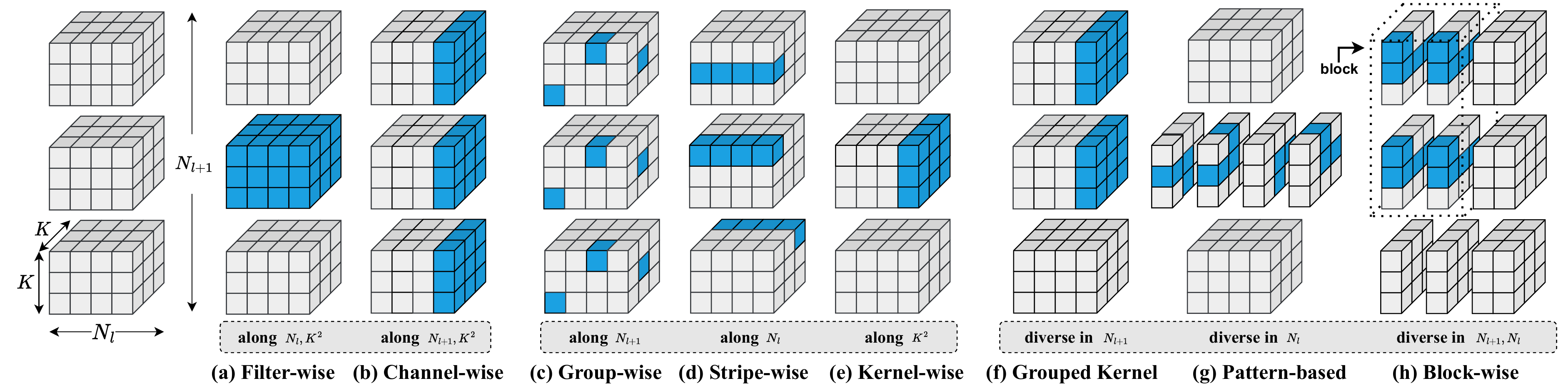}
    \caption{
    Explain different pruning granularities in three basic dimensions: the output dimension $N_{l+1}$, the input dimension $N_{l}$, and the kernel dimension $K^{2}$.
    ``Along" means the same pruning decision made to all weights at the same position in this dimension. In contrast, ``diverse" indicates that pruning results are different in this dimension.
    \textit{(a) Filter-wise} and \textit{(b) Channel-wise} represent well-known structured pruning. Remaining \textit{(c) - (h)} are special granularities. \textit{(c) Group-wise pruning} groups weights along the output dimension $N_{l+1}$. \textit{(d) Stripe-wise pruning} groups weights along the input dimension $N_{l}$. 
    \textit{(e) Kernel-wise pruning} groups weights along the kernel dimension $K^2$. \textit{(f) Grouped kernel pruning} groups kernels across several different filters in $N_{l+1}$ dimension. \textit{(g) Pattern-based pruning} utilizes pre-defined patterns to group weights in $N_{l}$ dimension. 
    \textit{(h) Block-wise pruning} groups kernels with the same pattern in both $N_{l+1}$ and $N_{l}$ dimensions.
    }
    \label{fig: granularity}
\end{figure*}

\subsubsection{Joint Compression}
\label{JOINT-COMPRESSION}

Some prevailing techniques, i.e., architecture search, decomposition and quantization, are orthogonal to pruning methods for neural network compression. Applying these techniques in sequence may seem like a natural extension but can lead to sub-optimal solutions due to different optimization objectives~\cite{wangAPQJointSearch2020}. Therefore, researchers have proposed to apply these techniques jointly.

\textbf{Pruning and NAS}: NPAS~\cite{liNPASCompilerAwareFramework2021} develops a framework of joint network pruning and architecture search. This method is compiler-aware and replaces computational activation functions (such as the sigmoid function) with compiler-friendly ones. 

\textbf{Pruning and Quantization:} DJPQ~\cite{wangDifferentiableJointPruning2020} jointly performs structured pruning and mixed-bit precision quantization in a gradient-based manner. This structured pruning is based on the variational information bottleneck~\cite{dai2018compressing}, and mixed-bit precision quantization is extended to power-of-two bit-restricted quantization.
Bayesian Bits (BB)~\cite{vanbaalenBayesianBitsUnifying2020} unifies the view of mixed precision quantization and pruning, and introduces mixed precision gates on filters for structured pruning.
IODF~\cite{wangFastLosslessNeural2022} utilizes integer-only arithmetic based on 8-bit quantization and has learnable binary gates to eliminate redundant filters during inference.

\textbf{Pruning, NAS, and Quantization:} APQ~\cite{wangAPQJointSearch2020} performs these tasks jointly. A once-for-all network~\cite{caiOnceforAllTrainOne2020} supporting a large search space for channel number is trained. Next, a quantization-aware accuracy predictor is designed to evaluate the accuracy of the selected structure with mixed-precision quantization~\cite{wangHAQHardwareAwareAutomated2019}. This is followed by the use of an evolutionary-based architecture search~\cite{guoSinglePathOneShot2020} to find the best-accuracy model under latency or energy constraints.

\textbf{Pruning and Decomposition:} Some studies have exploited structured pruning and low-rank decomposition within a unified framework. The authors of Hinge~\cite{liGroupSparsityHinge2020} contend that pruning techniques cannot deal with the last CONV layer in a residual block that low-rank decomposition is able to. The two techniques are hinged by introducing \textit{sparsity-inducing matrices} after filters. Imposing group sparsity on the columns and rows of \textit{sparsity-inducing matrices} achieves filter pruning and decomposition, respectively. Collaborative Compression (CC)~\cite{liCompactCNNsCollaborative2021} simultaneously learns the model sparsity and low-rankness to achieve channel pruning and tensor decomposition jointly.

\subsubsection{Special Granularity}
\label{PRUNING-FOR-SPECIAL-GRANULARITY}

Apart from unstructured pruning at the weight level and structured pruning at the filter level, there are also pruning methods at other granularities. 
To categorize these methods, we define three basic dimensions: output dimension $N_{l+1}$, input dimension $N_{l}$, and kernel dimension $K^2$. Filter pruning can be viewed as pruning along dimension $N_{l} \times K^2 $, and channel pruning is along dimension $N_{l+1} \times K^2 $. They are similar since pruning filters in the $l$-th layer will remove the corresponding channels in the $l+1$-th layer. More different pruning granularities are shown in Fig.~\ref{fig: granularity}.

\textbf{Group along basic dimensions:}
GBD~\cite{lebedevFastConvNetsUsing2016} and SSL~\cite{wenLearningStructuredSparsity2016} introduce \textit{group-wise pruning} (Fig.~\ref{fig: granularity}c) which prunes weights located at the same position among all filters in a layer, that is, along the output dimension $N_{l+1}$. SWP~\cite{mengPruningFilterFilter2020} prunes filters in a \textit{stripe-wise} (Fig.~\ref{fig: granularity}d) manner along input dimension $N_{l}$. 
Compared to \textit{group-wise pruning}, \textit{stripe-wise pruning} maintains filter independence, for there is no pattern across filters.
PCONV~\cite{maPCONVMissingDesirable2020} uses \textit{kernel-wise pruning} (Fig.~\ref{fig: granularity}e) to prune weights along kernel dimension $K^2$.

\textbf{Diverse in basic dimensions:}
Some researchers find making the same pruning decision along basic dimensions may not be optimal, and they propose to look into the basic dimensions. 
GKP-TMI~\cite{zhongRevisitKernelPruning2022} uses \textit{grouped kernel pruning} (Fig.~\ref{fig: granularity}f). 
On top of \textit{kernel-wise pruning}, this method groups different filters and makes diverse pruning decisions in the output dimension $N_{l+1}$.
The remaining kernels are re-permuted to output a densely structured pruned network that benefits from parallel computing. 
1xN~\cite{lin1xNPatternPruning2022} (Fig.~\ref{fig: granularity}f) uses a \textit{1xN pruning pattern} which treats $N$ output kernels sharing the same input channel index as the basic pruning granularity.
PCONV~\cite{maPCONVMissingDesirable2020} uses \textit{pattern-based pruning} (Fig.~\ref{fig: granularity}g) by looking into the kernels and making different pruning decisions in the kernel dimension $K^2$.
\textit{Block-wise pruning}~\cite{liNPASCompilerAwareFramework2021} (Fig.~\ref{fig: granularity}h) generalizes \textit{pattern-based pruning}~\cite{maPCONVMissingDesirable2020} to share a same pattern for a kernel group, where kernels are grouped in both the output dimension $N_{l+1}$ and the input dimension $N_{l}$.
\textbf{Other possible granularities.} 
There are some more possible structured granularities, such as grouped stripe (extension of Fig.~\ref{fig: granularity}d) and grouped pattern (extension of Fig.~\ref{fig: granularity}g). 
Multi-granularity~\cite{zhaoMultigranularityPruningModel2022} also deserves investigation in structure pruning.

\textbf{Layer-level:}
Besides breaking down filters into smaller granularities, Shallowing Deep Networks (SDN)~\cite{chenShallowingDeepNetworks2019} uses \textit{layer-wise pruning} to prune entire layers. Specifically, SDN places \textit{a linear classifier probe}~\cite{alain2016understanding} after each layer to evaluate the effectiveness of the layer. The network is retrained with the knowledge distillation technique after pruning unimportant layers.

\section{Future Directions}
\label{FUTURE-DIRECTIONS}

\subsection{Pruning Topics}
\label{PRUNING-TOPICS}
 Every subsection in Section~\ref{METHOD} has the potential for further development in the future. Besides, structured pruning will continue to borrow ideas from unstructured pruning. In this section, we will discuss some promising topics directly related to pruning.
 
\textbf{Pruning theory:} Apart from utilizing optimization tools for pruning in Section~\ref{OPTIMIZATION-TOOLS}, some works view pruning from the perspective of synaptic flow~\cite{tanakaPruningNeuralNetworks2020}, signal propagation~\cite{lee2019signal}, and graph theory~\cite{linCompactConvNetsStructureSparsity2020}.
Besides, the pruning process can be guided by leveraging the interpretations of a model~\cite{ganjdaneshInterpretationsSteeredNetwork2022}, the loss landscape~\cite{leeEnsembleKnowledgeGuided2022}, generalization-stability trade-off~\cite{bartoldson2020generalization} and entropy of a model~\cite{luo2017entropy}.
Moreover, the theory behind LTH is investigated with Logarithmic Pruning~\cite{orseau2020logarithmic}.
In addition, different training methods~\cite{yangProxSGDTrainingStructured2020, dingGlobalSparseMomentum2019, barsbeyHeavyTailsSGD2021} can be combined with pruning.
The above directions have the potential in structure pruning.

\textbf{Pruning mechanism:} Researchers have new views on the prevailing three-stage training-pruning-retraining mechanism.
First, the Lottery Ticket Hypothesis, which is proposed on unstructured pruning, is expected to extend on structured pruning. 
Second, single-shot pruning~\cite{miaoLearningPruningFriendlyNetworks2022, zhangOneshotPruningRecurrent2019} prunes only once to get the pruned models. Structured pruning can potentially benefit from this mechanism.
Third, AC/DC training~\cite{peste2021ac} is able to co-training the dense and sparse models. Therefore, handling multiple models during pruning and training is another promising direction for structured pruning.

\textbf{Pruning Rate:} Structured pruning can also extend the current weight pruning strategies on researching the layer-wise pruning ratios~\cite{heFilterPruningSwitching2022,leeLayeradaptiveSparsityMagnitudebased2021,liebenwein2021compressing}.

\textbf{Pruning Domain:} 
Utilizing representation in the frequency domain~\cite{liu2018frequency, zhangFilterPruningLearned2021} to guide pruning is another interesting direction.

\subsection{Pruning for Specific Tasks}
\label{PRUNING-FOR-SPECIFIC-TASK}
Pruning techniques can be applied to other tasks to achieve high computational efficiency. Here are some straightforward examples:
super-resolution~\cite{zhanAchievingOnMobileRealTime2021}, personal re-id~\cite{wangSoftPersonReidentification2022,wang2023progressive}, medical imaging diagnostic~\cite{fernandesAutomaticSearchingPruning2021}, face attribute classification~\cite{linFairGRAPEFairnessAwareGRAdient2022} and ensemble learning~\cite{bianSubarchitectureEnsemblePruning2022,whitaker2022prune}.
Apart from the tasks mentioned above, some emerging directions are still in the early stages but are promising in the future.

\textbf{Pruning for Federated Learning:}
Federated learning~\cite{federatedSurvey} aims to solve the problem of training a model without transferring data to a central location. 
Pruning~\cite{zhangFedDUAPFederatedLearning2022, jiangModelPruningEnables2022} helps alleviate communication costs required between devices and servers. Zhang \textit{et al.}~\cite{zhangFedDUAPFederatedLearning2022} contend that the non-Independent and Identically Distributed (non-IID) degree for each device and server is different, so it proposes to compute the different pruning ratio from each device and an aggregated expected pruning ratio for the server.

\textbf{Pruning for Continual Learning:}
Continual Learning tackles the problem of catastrophic forgetting~\cite{pengOvercomingLongTermCatastrophic2022}. 
Some pioneering works~\cite{golkar2019continual,yan2022learning ,mallyaPackNetAddingMultiple2018} use the pruned filters of the sparsified network to train a new task, so the training process does not cause deterioration to the performance of previous tasks.

\textbf{Limited Dataset:}
Dataset compression and pruning~\cite{yang2023dataset} are emerging. The idea is to utilize part of the images from the training dataset to train the network. 
Structured pruning~\cite{tangRebornFiltersPruning2020} has lots of topics to investigate along with this new trend.

\subsection{Pruning Specific Networks}
\label{PRUNING-SPECIFIC-NETWORKS}
Besides the prevailing CNNs, pruning is beneficial to other type of neural networks such as MLPs~\cite{wangWeightNoiseInjectionBased2019}, rectifier neural networks~\cite{serra2021scaling}, spiking neural networks~\cite{kim2022exploring, chowdhury2022towards}, and Generative adversarial network (GAN)~\cite{shuCoEvolutionaryCompressionUnpaired2019,liuContentawareGANCompression2021,songSPGANSelfGrowingPruning2021,liRevisitingDiscriminatorGAN2021}.

\red{\textbf{Pruning CNN-Based Transformers:}
There has been a growing trend to incorporate the architectural design of CNNs into Transformer models~\cite{zhang2022bootstrapping, d2021convit, chen2022dearkd, zhang2023vitaev2, ren2022co}. Therefore, adopting these structured pruning techniques to compress these novel architecture designs is meaningful~\cite{liu2022convnet,chavan2022vision}.
}

\red{
\textbf{Pruning Transformer-Based Architecture:}
The attention mechanism of Transformer architecture fundamentally operates using fully connected layers, which can be perceived as a unique manifestation of convolutional layers when the kernel size is set to 1~\cite{lecun2015convnets}.
In recent years, foundation models~\cite{bommasani2021opportunities}
such as GPT-3~\cite{brown2020language} and generalist agents such as Gato~\cite{reed2022a} are the possible way to Artificial general intelligence (AGI).
The efficiency of attention components of these huge models can benefit from the research of structured pruning.
}

\subsection{Pruning Targets}
\label{PRUNING-TARGET}

During past years, the target of pruning has shifted from reducing the number of parameters in unstructured pruning to minimizing the FLOPs in structured pruning. 
Recently, the target of pruning has been evolving to meet the practical demands of real-world scenarios.

\textbf{Hardware:} Incorporating pruning into the hardware process is an emerging trend. 
Hardware compiler-aware pruning~\cite{kimCPruneCompilerInformedModel2022} conducts pruning based on the structural information of subgraphs constructed during compiler tuning.

\textbf{Energy:} 
As the deployment of AI models expands, the rising energy consumption of AI models needs greater attention. Considering energy during pruning should be investigated. Energy-aware pruning~\cite{yangDesigningEnergyefficientConvolutional2017} greedily prunes the layer that consumes the most energy, for minimizing the MACs may not necessarily reduce the most energy consumption.

\textbf{Robustness:}
The robustness of a network describes how easily a network is fooled to make a wrong prediction under attacks~\cite{guiModelCompressionAdversarial2019}. Considering the robustness and computational cost of models together during the model design process is a new trend. Researchers find that sparsity brought by pruning can improve the adversarial robustness~\cite{guoSparseDNNsImproved2018}.
Linearity Grafting (Grafting)~\cite{chenLinearityGraftingRelaxed2022} operates on the premise that network robustness favors linear functions and proposes the \textit{linearity grafting} method. 
ANP-VS~\cite{madaanAdversarialNeuralPruning2020} identifies the latent features with high vulnerability and proposes a Bayesian framework to prune these features by minimizing adversarial loss and feature-level vulnerability.






%


\bibliographystyle{IEEEtran}
\normalem
\bibliography{structured-pruning-survey-pages}

%


\vspace{ -7mm}
\begin{IEEEbiography}
[{\includegraphics[width=1in,height=1.25in,clip,keepaspectratio]{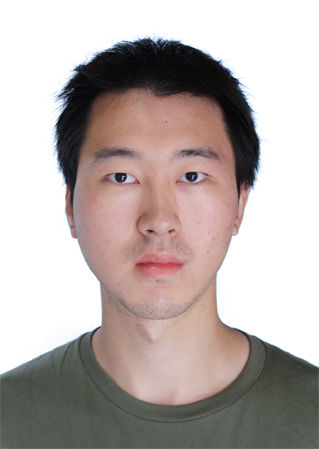}}]
{Yang He}
received the B.S. degree and M.Sc. from the University of Science and Technology of China, Hefei, China, in 2014 and 2017, respectively. He received his Ph.D. degree from the University of Technology Sydney in 2022. He currently is a scientist with the Centre for Frontier AI Research (CFAR), Agency for Science, Technology and Research (A*STAR), Singapore, and also with the Institute of High Performance Computing (IHPC), Agency for Science, Technology and Research (A*STAR), Singapore.
His research interests include deep learning, computer vision, and filter pruning.
\end{IEEEbiography}

\vspace{ -7mm}
\begin{IEEEbiography}
[{\includegraphics[width=1in,height=1.25in,clip,keepaspectratio]{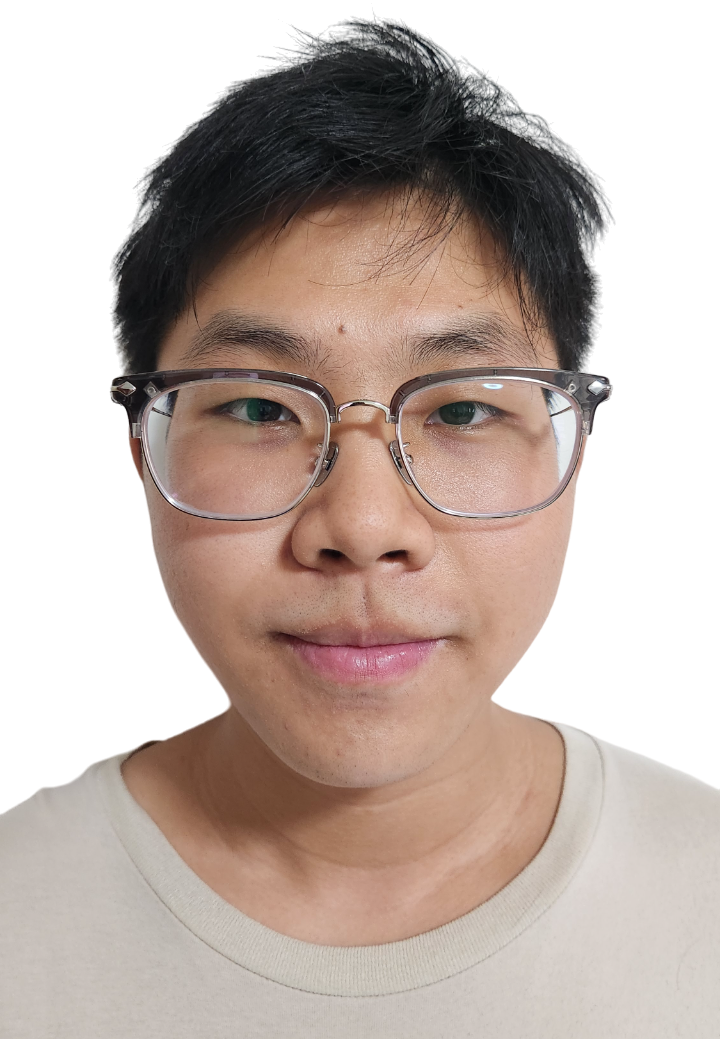}}]
{Lingao Xiao} is an intern student with the Centre for Frontier AI Research (CFAR), Agency for Science Technology and Research (A*STAR), Singapore, and also with the Institute of High Performance Computing (IHPC), Agency for Science, Technology and Research (A*STAR), Singapore.
He is pursuing his bachelor's degree in computer engineering at Nanyang Technological University, Singapore. 
His research interests include network compression and filter pruning.
\end{IEEEbiography}










\section*{Supplementary material}

# Supplementary for Structured Pruning for Deep Convolutional Neural Networks: A survey

Yang He, Lingao Xiao

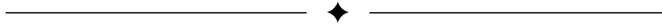

## 1 RELATED WORKS

**Unstructured Pruning (Weight Pruning):** The idea of pruning is first introduced by LeCun *et al.* [1] who propose *saliency* to measure the importance of each weight. The *saliency* is computed by the diagonal elements of the Hessian matrix. Han *et al.* [2] propose to iteratively prune weights with $\ell_1$ regularization and retrain. Guo *et al.* [3] propose *connection splicing* which restores the wrongly pruned weights. In addition, Guo *et al.* incorporate the *connection splicing* into the training process to dynamically learn the weights [3]. Han *et al.* [4] adopt a 3-stage pipeline to prune weights and achieve a 49× compression ratio with merely accuracy loss on VGG-16 [5], bringing back the research focus on network pruning.

**Quantization:** Quantization in model compression describes the process of approximating full precision weights (*32-bit floating point*) with lower precision weights (e.g., *8-bit integer*) to reduce computation and storage cost. The low-bit quantization includes binarization [6], [7] and ternarization [8], [9]. These methods can be roughly divided into two categories. First, heuristic quantization methods quantize weights by using mapping functions [7], [8]. Second, non-heuristic methods formulate quantization as an optimization problem [6], [9]. In addition, the quantization process also includes clustering and parameter sharing [10], and quantization can be combined with pruning [4], [11], [12].

**Decomposition:** Decomposition is referring to matrix decomposition and tensor decomposition, where a matrix is just a 2-D tensor. Specifically, a tensor is decomposed or factorized into products of several low-rank tensors that approximate the original tensor. To decompose 2-D tensors (matrices), *full-rank decomposition* [13], *singular value decomposition (SVD)* [14], and *QR* [15] are arguably the most popular low-rank matrix decomposition techniques. To deal with 4-D tensors, methods or concepts such as *Tucker* [16], *classical prolongation (CP)* [17], and *tensor networks* [18] are applied. Furthermore, decomposition is orthogonal to pruning, and these two techniques are applied together [19], [20], [21].

**Knowledge Distillation:** The idea of transferring knowledge from a usually large network to a small network is first introduced by Bucilua *et al.* [22], and the idea is later well-known as Knowledge Distillation [23]. The distillation process includes *(1) offline distillation* [23] which takes a pretrained teacher model. Distillation can also be *(2) online distillation* [24] which simultaneously updates the teacher model and the student model in an end-to-end manner. Furthermore, *(3) self-distillation* [25] requires only one model and distills the knowledge from deeper sections to shallow sections. In addition, knowledge distillation can work with pruning to achieve few-shot compression which requires few data [26], [27], [28].

**Neural Architecture Search:** The objective of neural architecture search [29] is to minimize human intervention in the search for neural network designs. In this regard, three factors play a significant role. (1) The search space defines potential candidates. Selecting candidates with prior knowledge reduces the frequently vast search space, but human bias may hinder performance. (2) Search strategies are the search space exploration algorithms. Simple taxonomy involves Bayesian optimization [30], evolutionary methods [31], gradient-based methods [32], and reinforcement learning [33]. (3) The evaluation strategy is used to assess the predictive ability of the model on unseen data. To address the prohibitive computational cost of the model's evaluation, efficient methods are developed.

## 2 EXPERIMENTS

We summarize 800+ experiment results from 100+ papers into **21 tables** over the three most popular datasets: CIFAR-10, CIFAR-100, and ImageNet-1K. The detailed numbers of covered methods, models and results are summarized in Tab. 1. The 21 tables are easily accessible via the links in Tab. 2.

In each table, we specify their corresponding sections and the years they were published in each table. Note that various methods adopt distinct settings for baselines and fine-tuning. To offer a comprehensive contrast, we present nine metrics for these methods:

1) Baseline Accuracy
2) Accuracy After Pruning
3) Accuracy Drop (%)
4) Baseline FLOPs
5) FLOPs After Pruning
6) FLOPs Drop (%)
7) Baseline Parameters
8) Parameters After Pruning
9) Parameters Drop (%)

| Section | 2.1.1 | 2.1.2 | 2.2.1 | 2.2.2 | 2.2.3 | 2.3.1 | 2.3.2 | 2.3.3 | 2.4.1 | 2.4.2 | 2.4.3 | 2.5.1 | 2.5.2 | 2.6.1 | 2.6.2 | 2.6.3 | 2.7.1 | 2.7.2 | 2.7.3 | Total |
|---|---|---|---|---|---|---|---|---|---|---|---|---|---|---|---|---|---|---|---|---|
| #. methods | 1 | 5 | 4 | 3 | 4 | 6 | 9 | 3 | 6 | 4 | 4 | 7 | 6 | 6 | 11 | 4 | 6 | 6 | 7 | 102 |
| #. models | 4 | 11 | 9 | 9 | 13 | 9 | 12 | 7 | 12 | 8 | 10 | 13 | 10 | 13 | 13 | 11 | 9 | 13 | 10 | 27* |
| #. results | 7 | 30 | 53 | 30 | 36 | 44 | 72 | 26 | 62 | 18 | 19 | 71 | 39 | 51 | 99 | 45 | 33 | 38 | 35 | 808 |

TABLE 1: Summary of experiments in 21 tables. * denotes the unique models.

| | | |
|---|---|---|
| CIFAR-10 | VGG | VGG-16, VGG-19 |
| | ResNet (Small) | ResNet-20, ResNet-32, ResNet-44, ResNet-56, ResNet-110, ResNet-164 |
| | ResNet (Large) | ResNet-18, ResNet-34, ResNet-50, ResNet-101 |
| | MobileNet | MobileNet-V1, MobileNet-V2 |
| | Other | DenseNet-40, GoogLeNet, PreResNet-29, PreResNet-101, ResNeXt-20, ResNeXt-164, NAS |
| CIFAR-100 | VGG | VGG-16, VGG-19 |
| | ResNet (Small) | ResNet-20, ResNet-32, ResNet-44, ResNet-56, ResNet-110, ResNet-164 |
| | ResNet (Large) | ResNet-18, ResNet-34, ResNet-50, ResNet-101 |
| | MobileNet | MobileNet-V1, MobileNet-V2 |
| | Other | DenseNet-40, PreResNet-29, PreResNet-101, ResNeXt-20, ResNeXt-164 |
| ImageNet-1K | AlexNet | AlexNet |
| | VGG | VGG-11, VGG-16 |
| | ResNet | ResNet-18, ResNet-34, ResNet-50, ResNet-101, ResNet-152 |
| | MobileNet | MobileNet-V1, MobileNet-V2, MobileNet-V3-Small |
| | Other | ProxylessNet-Mobile, GoogLeNet, ResNeXt-50, NAS |

TABLE 2: Indexing of the 21 tables.

For papers that only report the FLOPs drop ratio without a baseline, we determine the pruned model's absolute FLOPs using our calculation of the FLOPs baseline. We apply a similar approach to the parameter count. Tables are sorted according to the FLOPs of pruned models.

## 3 OUR WEBSITE

To better visualize and analyze the experiment results, we have developed a dedicated website that offers a more interactive and dynamic platform for comparing structured pruning methods. The website is available at: https://huggingface.co/spaces/he-yang/Structured-Pruning-Survey.

We provide an illustrative example in Fig. 1 to show how the query works. If a user wants to find methods that satisfy the following:

1) Select Dataset: ImageNet-1K
2) Select Model: ResNet-50
3) Select Pruning Method: Regularization-based Pruning in section 2.3
4) Target 1: Accuracy after pruning > 75%
5) Target 2: Pruned FLOPs > 40%
6) Target 3: Model size after pruning < 30M

By simply entering the requirements into the corresponding query box, we can narrow down the results to five data points from three methods, including GBN, SCOP, and OTO. If we click the method name, such as SCOP, the corresponding paper title, link, code, venues and BibTeX for this method will be generated.

Fig. 1: An illustrative example sourced from our website: https://huggingface.co/spaces/he-yang/Structured-Pruning-Survey.

| Section | Year | Method | Baseline Top-1 Acc.(%) | Pruned Top-1 Acc.(%) | Top-1 Acc. ↓(%) | Baseline FLOPs (M) | Pruned FLOPs (M) | FLOPs ↓(%) | Baseline Params (M) | Pruned Params (M) | Params ↓(%) |
|---|---|---|---|---|---|---|---|---|---|---|---|
| VGG-16 | | | | | | | | | | | |
| 2.6.1 | 2020 | AutoCompress [34] | 93.70 | 93.21 | -0.49 | 314.59 | 35.75 | 88.64 | 14.73 | - | - |
| 2.6.1 | 2022 | DECORE [35] | 93.96 | 91.68 | -2.28 | 314.59 | 36.95 | 88.25 | 14.73 | 0.26 | 98.26 |
| 2.4.3 | 2022 | StructADMM [36] | - | 93.10 | - | 314.59 | 37.90 | 87.95 | 14.73 | - | - |
| 2.2.3 | 2020 | PFP [37] | 92.89 | 92.39 | -0.50 | 314.59 | 47.09 | 85.03 | 14.73 | 0.84 | 94.32 |
| 2.3.3 | 2021 | OTO [38] | 91.60 | 91.00 | -0.60 | 314.59 | 51.28 | 83.70 | 14.73 | 0.37 | 97.50 |
| 2.6.1 | 2022 | DECORE [35] | 93.96 | 92.44 | -1.52 | 314.59 | 51.34 | 83.68 | 14.73 | 0.50 | 96.60 |
| 2.3.2 | 2021 | ABP [39] | 93.96 | 92.65 | -1.31 | 314.59 | 52.81 | 83.21 | 14.73 | 1.50 | 89.79 |
| 2.7.2 | 2021 | EDP [40] | 93.60 | 93.52 | -0.08 | 314.59 | 62.57 | 80.11 | 14.73 | 0.65 | 95.59 |
| 2.6.1 | 2022 | RL-MCTS [41] | 93.51 | 93.72 | 0.21 | 314.59 | 63.23 | 79.90 | 14.73 | - | - |
| 2.2.1 | 2021 | CHIP [42] | 93.96 | 93.18 | -0.78 | 314.59 | 67.32 | 78.60 | 14.73 | 1.87 | 87.30 |
| 2.4.2 | 2018 | VIBNet [43] | - | 91.50 | - | 314.59 | 70.99 | 77.43 | 14.73 | 0.78 | 94.70 |
| 2.2.3 | 2022 | DLRFC [44] | 93.25 | 93.64 | -0.39 | 314.59 | 72.51 | 76.95 | 14.73 | 0.83 | 94.38 |
| 2.2.1 | 2020 | HRank [45] | 93.96 | 91.23 | -2.73 | 314.59 | 73.90 | 76.51 | 14.73 | 1.75 | 88.12 |
| 2.3.2 | 2022 | WhiteBox [46] | 93.02 | 93.47 | 0.45 | 314.59 | 74.24 | 76.40 | 14.73 | - | - |
| 2.1.2 | 2022 | EPruner [47] | 93.02 | 93.08 | 0.06 | 314.59 | 74.42 | 76.34 | 14.73 | 1.65 | 88.80 |
| 2.2.2 | 2019 | AOFP [48] | 93.38 | 93.28 | -0.10 | 314.59 | 77.39 | 75.40 | 14.73 | - | - |
| 2.1.2 | 2022 | CLR-RNF [49] | 93.02 | 93.32 | 0.30 | 314.59 | 81.48 | 74.10 | 14.73 | 0.74 | 95.00 |
| 2.6.3 | 2020 | ABCPruner [50] | 93.02 | 93.08 | 0.06 | 314.59 | 82.81 | 73.68 | 14.73 | 1.67 | 88.66 |
| 2.1.2 | 2019 | COP [51] | 93.56 | 93.31 | -0.25 | 314.59 | 83.37 | 73.50 | 14.73 | 1.06 | 92.80 |
| 2.3.3 | 2021 | OTO [38] | 93.20 | 93.30 | 0.10 | 314.59 | 84.31 | 73.20 | 14.73 | 0.81 | 94.50 |
| 2.2.1 | 2022 | GCNP [52] | 93.10 | 93.08 | -0.02 | 314.59 | 84.72 | 73.07 | 14.73 | 1.02 | 93.06 |
| 2.5.2 | 2022 | FTWT [53] | 93.82 | 93.19 | -0.63 | 314.59 | 84.94 | 73.00 | 14.73 | - | - |
| 2.4.2 | 2018 | VIBNet [43] | - | 93.90 | - | 314.59 | 87.26 | 72.26 | 14.73 | 0.80 | 94.55 |
| 2.7.3 | 2022 | SOKS [54] | 93.53 | 94.01 | 0.48 | 314.59 | 87.46 | 72.20 | 14.73 | 3.19 | 78.33 |
| 2.4.2 | 2019 | RBP [55] | - | 91.00 | - | 314.59 | 89.88 | 71.43 | 14.73 | - | - |
| 2.7.3 | 2020 | SWP [56] | 93.25 | 93.65 | 0.40 | 314.59 | 90.73 | 71.16 | 14.73 | 1.08 | 92.66 |
| 2.3.2 | 2021 | GDP-Guo [57] | 93.89 | 93.99 | 0.10 | 314.59 | 96.11 | 69.45 | 14.73 | - | - |
| 2.2.1 | 2021 | CHIP [42] | 93.96 | 93.72 | -0.24 | 314.59 | 105.07 | 66.60 | 14.73 | 2.46 | 83.30 |
| 2.3.1 | 2020 | SCP [58] | 93.85 | 93.79 | 0.06 | 314.59 | 106.24 | 66.23 | 14.73 | 1.02 | 93.05 |
| 2.3.2 | 2021 | ABP [39] | 93.96 | 93.50 | -0.46 | 314.59 | 106.59 | 66.12 | 14.73 | 2.66 | 81.96 |
| 2.2.2 | 2019 | AOFP [48] | 93.38 | 93.47 | 0.09 | 314.59 | 108.55 | 65.50 | 14.73 | - | - |
| 2.2.1 | 2020 | HRank [45] | 93.96 | 92.34 | -1.62 | 314.59 | 108.91 | 65.38 | 14.73 | 2.60 | 82.38 |
| 2.5.2 | 2022 | FTWT [53] | 93.82 | 93.55 | -0.27 | 314.59 | 110.11 | 65.00 | 14.73 | - | - |
| 2.6.1 | 2022 | DECORE [35] | 93.96 | 93.56 | -0.40 | 314.59 | 110.81 | 64.78 | 14.73 | 1.63 | 88.92 |
| 2.6.3 | 2022 | CCEP [59] | 93.71 | 93.52 | -0.19 | 314.59 | 115.77 | 63.20 | 14.73 | - | - |
| 2.4.2 | 2018 | VIBNet [43] | - | 94.20 | - | 314.59 | 116.59 | 62.94 | 14.73 | 0.85 | 94.21 |
| 2.2.3 | 2022 | DLRFC [44] | 93.25 | 93.93 | -0.68 | 314.59 | 121.97 | 61.23 | 14.73 | 1.05 | 92.86 |
| 2.7.2 | 2021 | CC [60] | 93.70 | 94.09 | 0.39 | 314.59 | 123.62 | 60.70 | 14.73 | 4.02 | 72.69 |
| 2.7.3 | 2022 | SOKS [54] | 93.53 | 94.11 | 0.58 | 314.59 | 124.23 | 60.51 | 14.73 | 4.50 | 69.47 |
| 2.2.2 | 2019 | AOFP [48] | 93.38 | 93.84 | 0.46 | 314.59 | 124.63 | 60.38 | 14.73 | - | - |
| 2.2.1 | 2021 | CHIP [42] | 93.96 | 93.86 | -0.10 | 314.59 | 131.81 | 58.10 | 14.73 | 2.71 | 81.60 |
| 2.2.1 | 2022 | GCNP [52] | 93.10 | 93.27 | 0.17 | 314.59 | 134.36 | 57.29 | 14.73 | 2.14 | 85.50 |
| 2.6.2 | 2021 | EE [61] | 93.36 | 93.63 | 0.27 | 314.59 | 136.53 | 56.60 | 14.73 | - | - |
| 2.5.2 | 2022 | FTWT [53] | 93.82 | 93.73 | -0.09 | 314.59 | 138.42 | 56.00 | 14.73 | - | - |
| 2.3.1 | 2020 | PR [62] | 93.88 | 93.92 | -0.04 | 314.59 | 144.71 | 54.00 | 14.73 | - | - |
| 2.2.1 | 2020 | HRank [45] | 93.96 | 93.43 | -0.53 | 314.59 | 146.01 | 53.59 | 14.73 | 2.47 | 83.24 |
| 2.3.2 | 2021 | ABP [39] | 93.96 | 93.75 | -0.21 | 314.59 | 146.19 | 53.53 | 14.73 | 2.44 | 83.46 |
| 2.7.2 | 2021 | CC [60] | 93.70 | 94.15 | 0.45 | 314.59 | 154.78 | 50.80 | 14.73 | 5.02 | 65.90 |
| 2.6.1 | 2022 | RL-MCTS [41] | 93.51 | 93.90 | 0.39 | 314.59 | 171.45 | 45.50 | 14.73 | - | - |
| 2.3.2 | 2019 | GAL [63] | 93.96 | 90.78 | -3.18 | 314.59 | 172.36 | 45.21 | 14.73 | 2.63 | 82.18 |
| 2.3.2 | 2019 | GAL [63] | 93.96 | 93.42 | -0.54 | 314.59 | 172.36 | 45.21 | 14.73 | 2.63 | 82.18 |
| 2.7.1 | 2020 | EB [64] | - | 92.49 | - | 314.59 | 174.77 | 44.44 | 14.73 | 4.42 | 70.00 |
| 2.2.2 | 2019 | AOFP [48] | 93.38 | 94.03 | 0.65 | 314.59 | 186.94 | 40.58 | 14.73 | - | - |
| 2.3.2 | 2019 | GAL [63] | 93.96 | 93.77 | -0.19 | 314.59 | 190.01 | 39.60 | 14.73 | 3.30 | 77.57 |
| 2.3.2 | 2019 | GAL [63] | 93.96 | 92.03 | -1.93 | 314.59 | 190.01 | 39.60 | 14.73 | 3.30 | 77.57 |
| 2.4.2 | 2019 | VP [65] | 93.25 | 93.18 | -0.07 | 314.59 | 190.97 | 39.30 | 14.73 | 3.93 | 73.35 |
| 2.7.2 | 2020 | Hinge [66] | 94.02 | 93.59 | -0.43 | 314.59 | 191.68 | 39.07 | 14.73 | 2.94 | 80.05 |
| 2.7.3 | 2019 | SDN [67] | 93.50 | 93.47 | -0.03 | 314.59 | 192.21 | 38.90 | 14.73 | 1.78 | 87.90 |
| 2.2.1 | 2021 | LRMF [68] | 93.58 | 93.70 | 0.12 | 314.59 | 201.65 | 35.90 | 14.73 | - | - |
| 2.2.1 | 2021 | LRMF [68] | 93.58 | 93.20 | -0.38 | 314.59 | 201.65 | 35.90 | 14.73 | - | - |
| 2.6.1 | 2022 | DECORE [35] | 93.96 | 94.02 | 0.06 | 314.59 | 203.64 | 35.27 | 14.73 | 5.45 | 63.02 |
| 2.1.1 | 2017 | PFEC [69] | 93.25 | 93.40 | 0.15 | 314.59 | 207.05 | 34.19 | 14.73 | 5.30 | 64.00 |
| 2.2.2 | 2019 | AOFP [48] | 93.38 | 93.81 | 0.43 | 314.59 | 216.09 | 31.31 | 14.73 | - | - |
| 2.7.1 | 2020 | EB [64] | - | 93.90 | - | 314.59 | 224.71 | 28.57 | 14.73 | 7.37 | 50.00 |
| 2.7.1 | 2020 | EB [64] | - | 93.91 | - | 314.59 | 262.16 | 16.67 | 14.73 | 10.31 | 30.00 |
| 2.7.1 | 2022 | EarlyCroP [70] | 90.20 | 93.00 | 2.80 | 314.59 | - | - | 14.73 | 0.29 | 98.00 |
| 2.7.3 | 2022 | DPP [71] | 93.50 | 93.60 | 0.10 | 314.59 | - | - | 14.73 | 1.58 | 89.27 |

TABLE 3: VGG-16 on CIFAR-10.

| Section | Year | Method | Baseline Top-1 Acc.(%) | Pruned Top-1 Acc.(%) | Top-1 Acc. ↓(%) | Baseline FLOPs (M) | Pruned FLOPs (M) | FLOPs ↓(%) | Baseline Params (M) | Pruned Params (M) | Params ↓(%) |
|---|---|---|---|---|---|---|---|---|---|---|---|
| VGG-19 | | | | | | | | | | | |
| 2.6.1 | 2022 | DECORE [35] | 93.76 | 91.65 | -2.11 | 398.00 | 43.85 | 88.98 | 20.04 | 0.30 | 98.50 |
| 2.4.1 | 2019 | ED [72] | 94.17 | 92.29 | -1.88 | 398.00 | 53.69 | 86.51 | 20.04 | 0.57 | 97.15 |
| 2.5.1 | 2021 | SEP [73] | 93.66 | 93.40 | -0.26 | 398.00 | 59.80 | 84.97 | 20.04 | 4.01 | 79.99 |
| 2.4.1 | 2019 | ED [72] | 93.71 | 91.79 | -1.92 | 398.00 | 60.42 | 84.82 | 20.04 | 0.63 | 96.84 |
| 2.3.1 | 2020 | SCP [58] | 93.84 | 93.82 | 0.02 | 398.00 | 103.24 | 74.06 | 20.04 | 0.96 | 95.21 |
| 2.2.3 | 2018 | DCP [74] | 93.99 | 94.57 | 0.58 | 398.00 | 139.16 | 65.03 | 20.04 | 1.29 | 93.58 |
| 2.4.1 | 2022 | SOSP [75] | 94.18 | 93.73 | -0.45 | 398.00 | 168.19 | 57.74 | 20.04 | 2.55 | 87.29 |
| 2.5.1 | 2021 | DCP-CAC [76] | 92.47 | 93.19 | 0.72 | 398.00 | 195.00 | 51.01 | 20.04 | 5.51 | 72.53 |
| 2.3.1 | 2017 | NS [77] | 93.66 | 93.80 | 0.14 | 398.00 | 196.98 | 50.51 | 20.04 | 2.30 | 88.52 |
| 2.2.3 | 2018 | DCP [74] | 93.99 | 94.16 | 0.17 | 398.00 | 199.00 | 50.00 | 20.04 | 10.44 | 47.92 |
| 2.4.1 | 2022 | SOSP [75] | 94.18 | 93.99 | -0.19 | 398.00 | 215.08 | 45.96 | 20.04 | 2.86 | 85.75 |
| 2.4.1 | 2019 | ED [72] | 93.71 | 93.88 | 0.17 | 398.00 | 239.44 | 39.84 | 20.04 | 4.11 | 79.50 |
| 2.4.1 | 2019 | ED [72] | 94.17 | 93.98 | -0.19 | 398.00 | 250.22 | 37.13 | 20.04 | 4.37 | 78.18 |
| 2.7.1 | 2022 | ProsPr [78] | 93.60 | 93.61 | 0.01 | 398.00 | - | - | 20.04 | 4.01 | 80.00 |
| 2.7.1 | 2022 | ProsPr [78] | 93.60 | 93.64 | 0.04 | 398.00 | - | - | 20.04 | 2.00 | 90.00 |
| 2.7.1 | 2022 | ProsPr [78] | 93.60 | 93.32 | -0.28 | 398.00 | - | - | 20.04 | 1.00 | 95.00 |

TABLE 4: VGG-19 on CIFAR-10.

| Section | Year | Method | Baseline Top-1 Acc.(%) | Pruned Top-1 Acc.(%) | Top-1 Acc. ↓(%) | Baseline FLOPs (M) | Pruned FLOPs (M) | FLOPs ↓(%) | Baseline Params (M) | Pruned Params (M) | Params ↓(%) |
|---|---|---|---|---|---|---|---|---|---|---|---|
| ResNet-20 | | | | | | | | | | | |
| 2.6.2 | 2020 | DSA [79] | 92.17 | 90.24 | -1.93 | 40.81 | 13.26 | 67.50 | 0.27 | - | - |
| 2.7.3 | 2022 | SOKS [54] | 92.05 | 90.78 | -1.27 | 40.81 | 15.49 | 62.04 | 0.27 | 0.14 | 48.15 |
| 2.3.2 | 2020 | SCOP [80] | 92.22 | 90.75 | -1.47 | 40.81 | 18.08 | 55.70 | 0.27 | 0.12 | 56.30 |
| 2.7.2 | 2020 | Hinge [66] | 92.54 | 91.84 | -0.70 | 40.81 | 18.57 | 54.50 | 0.27 | 0.12 | 55.45 |
| 2.5.2 | 2021 | ManiDP [81] | 92.22 | 92.05 | -0.17 | 40.81 | 18.69 | 54.20 | 0.27 | - | - |
| 2.2.1 | 2021 | LRMF [68] | 92.20 | 90.47 | -1.73 | 40.81 | 18.77 | 54.00 | 0.27 | - | - |
| 2.1.2 | 2019 | FPGM [82] | 92.20 | 91.99 | -0.21 | 40.81 | 18.77 | 54.00 | 0.27 | - | - |
| 2.5.2 | 2021 | DDG [83] | 92.24 | 91.90 | -0.34 | 40.81 | 19.43 | 52.40 | 0.27 | - | - |
| 2.6.2 | 2022 | DAIS [84] | 93.25 | 92.89 | -0.36 | 40.81 | 19.96 | 51.10 | 0.27 | - | - |
| 2.6.1 | 2022 | GNN-RL [85] | 91.73 | 91.31 | -0.42 | 40.81 | 20.00 | 51.00 | 0.27 | - | - |
| 2.2.1 | 2022 | GCNP [52] | 92.25 | 91.58 | -0.67 | 40.81 | 20.18 | 50.54 | 0.27 | 0.17 | 38.51 |
| 2.6.2 | 2020 | DSA [79] | 92.17 | 91.38 | -0.79 | 40.81 | 20.28 | 50.30 | 0.27 | - | - |
| 2.6.1 | 2021 | AGMC [86] | 91.73 | 91.42 | -0.31 | 40.81 | 20.41 | 50.00 | 0.27 | - | - |
| 2.3.2 | 2021 | ABP [39] | 92.15 | 91.03 | -1.12 | 40.81 | 21.34 | 47.70 | 0.27 | 0.15 | 45.10 |
| 2.1.2 | 2021 | SRR [87] | 92.27 | 92.48 | 0.21 | 40.81 | 22.12 | 45.80 | 0.27 | - | - |
| 2.2.3 | 2020 | PFP [37] | 91.40 | 90.91 | -0.49 | 40.81 | 22.26 | 45.46 | 0.27 | 0.10 | 62.67 |
| 2.6.2 | 2019 | TAS [88] | 92.88 | 92.88 | 0.00 | 40.81 | 22.45 | 45.00 | 0.27 | - | - |
| 2.7.3 | 2022 | GKP-TMI [89] | 92.35 | 92.01 | -0.34 | 40.81 | 23.30 | 42.90 | 0.27 | 0.15 | 43.40 |
| 2.5.2 | 2021 | DDG [83] | 92.24 | 92.27 | 0.03 | 40.81 | 23.38 | 42.70 | 0.27 | - | - |
| 2.5.1 | 2018 | SFP [90] | 92.20 | 90.83 | -1.37 | 40.81 | 23.59 | 42.20 | 0.27 | - | - |
| 2.2.1 | 2021 | LRMF [68] | 92.20 | 91.04 | -1.16 | 40.81 | 23.59 | 42.20 | 0.27 | - | - |
| 2.7.3 | 2022 | SOKS [54] | 92.05 | 91.83 | -0.22 | 40.81 | 23.63 | 42.09 | 0.27 | 0.16 | 40.74 |
| 2.2.1 | 2022 | GCNP [52] | 92.25 | 92.22 | -0.03 | 40.81 | 28.38 | 30.47 | 0.27 | 0.20 | 27.42 |
| 2.5.1 | 2018 | SFP [90] | 92.20 | 91.20 | -1.00 | 40.81 | 28.85 | 29.30 | 0.27 | - | - |
| 2.6.2 | 2020 | DSA [79] | 92.17 | 92.10 | -0.07 | 40.81 | 30.20 | 26.00 | 0.27 | - | - |
| 2.4.2 | 2019 | VP [65] | 92.01 | 91.66 | -0.35 | 40.81 | 34.39 | 15.73 | 0.27 | 0.22 | 19.05 |
| ResNet-32 | | | | | | | | | | | |
| 2.4.1 | 2019 | ED [72] | 95.30 | 93.05 | -2.25 | 69.12 | 3.64 | 94.74 | 0.47 | 0.02 | 96.05 |
| 2.4.1 | 2019 | ED [72] | 95.30 | 95.17 | -0.13 | 69.12 | 20.56 | 70.25 | 0.47 | 0.13 | 71.99 |
| 2.4.1 | 2022 | SOSP [75] | 95.30 | 95.22 | -0.08 | 69.12 | 22.22 | 67.85 | 0.47 | 0.13 | 72.85 |
| 2.4.1 | 2022 | SOSP [75] | 95.30 | 95.06 | -0.24 | 69.12 | 22.56 | 67.36 | 0.47 | 0.13 | 72.33 |
| 2.5.2 | 2021 | ManiDP [81] | 92.66 | 92.15 | -0.51 | 69.12 | 25.44 | 63.20 | 0.47 | - | - |
| 2.3.2 | 2020 | SCOP [80] | 92.66 | 92.13 | -0.53 | 69.12 | 30.55 | 55.80 | 0.47 | 0.21 | 56.20 |
| 2.7.3 | 2022 | SOKS [54] | 92.82 | 92.02 | -0.80 | 69.12 | 31.39 | 54.58 | 0.47 | 0.21 | 54.35 |
| 2.5.2 | 2021 | DDG [83] | 93.22 | 92.96 | -0.26 | 69.12 | 31.52 | 54.40 | 0.47 | - | - |
| 2.6.2 | 2022 | DAIS [84] | 92.92 | 93.49 | 0.57 | 69.12 | 31.86 | 53.90 | 0.47 | - | - |
| 2.1.2 | 2019 | COP [51] | 92.64 | 91.97 | -0.67 | 69.12 | 31.86 | 53.90 | 0.47 | 0.20 | 57.50 |
| 2.6.2 | 2022 | MFP [91] | 92.63 | 91.85 | -0.78 | 69.12 | 32.35 | 53.20 | 0.47 | - | - |
| 2.2.1 | 2021 | LRMF [68] | 92.63 | 92.08 | -0.55 | 69.12 | 32.35 | 53.20 | 0.47 | - | - |
| 2.2.1 | 2021 | LRMF [68] | 92.63 | 92.08 | -0.55 | 69.12 | 32.35 | 53.20 | 0.47 | - | - |
| 2.1.2 | 2019 | FPGM [82] | 92.63 | 92.82 | 0.19 | 69.12 | 32.35 | 53.20 | 0.47 | - | - |
| 2.6.2 | 2020 | LFPC [92] | 92.63 | 92.12 | -0.51 | 69.12 | 32.76 | 52.60 | 0.47 | - | - |
| 2.6.1 | 2022 | GNN-RL [85] | 92.63 | 92.58 | -0.05 | 69.12 | 33.87 | 51.00 | 0.47 | - | - |
| 2.5.1 | 2021 | DCP-CAC [76] | 92.36 | 92.21 | -0.15 | 69.12 | 34.49 | 50.10 | 0.47 | 0.25 | 47.83 |
| 2.6.1 | 2021 | AGMC [86] | 92.63 | 90.96 | -1.67 | 69.12 | 34.56 | 50.00 | 0.47 | - | - |
| 2.6.2 | 2019 | TAS [88] | 93.89 | 93.16 | -0.73 | 69.12 | 34.97 | 49.40 | 0.47 | - | - |
| 2.7.3 | 2022 | SOKS [54] | 92.82 | 92.44 | -0.38 | 69.12 | 36.74 | 46.85 | 0.47 | 0.27 | 43.48 |
| 2.3.2 | 2021 | ABP [39] | 92.63 | 92.55 | -0.08 | 69.12 | 37.12 | 46.30 | 0.47 | 0.27 | 43.60 |
| 2.5.2 | 2021 | DDG [83] | 93.22 | 93.21 | -0.01 | 69.12 | 39.12 | 43.40 | 0.47 | - | - |
| 2.7.3 | 2022 | GKP-TMI [89] | 92.82 | 93.04 | 0.22 | 69.12 | 39.33 | 43.10 | 0.47 | 0.27 | 43.40 |
| 2.5.1 | 2018 | SFP [90] | 92.63 | 90.08 | -2.55 | 69.12 | 40.44 | 41.50 | 0.47 | - | - |
| 2.5.1 | 2021 | DCP-CAC [76] | 92.36 | 92.85 | 0.49 | 69.12 | 48.35 | 30.04 | 0.47 | 0.35 | 26.09 |
| 2.5.1 | 2018 | SFP [90] | 92.63 | 90.63 | -2.00 | 69.12 | 49.21 | 28.80 | 0.47 | - | - |
| ResNet-44 | | | | | | | | | | | |
| 2.5.1 | 2021 | DCP-CAC [76] | 92.45 | 92.42 | -0.03 | 97.15 | 48.54 | 50.04 | 0.66 | 0.35 | 47.69 |
| 2.6.1 | 2021 | AGMC [86] | 93.10 | 92.28 | -0.82 | 97.15 | 48.58 | 50.00 | 0.66 | - | - |
| 2.5.1 | 2021 | DCP-CAC [76] | 92.45 | 93.26 | 0.81 | 97.15 | 67.98 | 30.03 | 0.66 | 0.46 | 30.77 |

TABLE 5: ResNet-20/32/44 on CIFAR-10.

| Section | Year | Method | Baseline Top-1 Acc.(%) | Pruned Top-1 Acc.(%) | Top-1 Acc. ↓(%) | Baseline FLOPs (M) | Pruned FLOPs (M) | FLOPs ↓(%) | Baseline Params (M) | Pruned Params (M) | Params ↓(%) |
|---|---|---|---|---|---|---|---|---|---|---|---|
| ResNet-56 | | | | | | | | | | | |
| 2.2.3 | 2020 | PFP [37] | 92.95 | 92.67 | -0.28 | 125.75 | 19.59 | 84.42 | 0.86 | 0.09 | 88.98 |
| 2.6.1 | 2022 | DECORE [35] | 93.26 | 90.85 | -2.41 | 125.75 | 23.27 | 81.50 | 0.86 | 0.13 | 84.71 |
| 2.3.2 | 2021 | ResRep [93] | 93.71 | 92.66 | 1.05 | 125.75 | 27.88 | 77.83 | 0.86 | - | - |
| 2.2.1 | 2022 | GCNP [52] | 93.72 | 92.75 | -0.97 | 125.75 | 28.65 | 77.22 | 0.86 | 0.25 | 70.50 |
| 2.7.2 | 2020 | Hinge [66] | 92.95 | 92.65 | -0.30 | 125.75 | 30.18 | 76.00 | 0.86 | 0.18 | 79.20 |
| 2.7.3 | 2020 | SWP [56] | 93.10 | 92.98 | -0.12 | 125.75 | 30.68 | 75.60 | 0.86 | 0.19 | 77.70 |
| 2.2.1 | 2020 | HRank [45] | 93.26 | 90.72 | -2.54 | 125.75 | 32.59 | 74.09 | 0.86 | 0.27 | 68.24 |
| 2.2.1 | 2021 | CHIP [42] | 93.26 | 92.05 | -1.21 | 125.75 | 34.83 | 72.30 | 0.86 | 0.24 | 71.80 |
| 2.6.2 | 2022 | DAIS [84] | 92.53 | 93.53 | 1.00 | 125.75 | 36.59 | 70.90 | 0.86 | - | - |
| 2.3.1 | 2019 | GBN [94] | - | - | 0.03 | 125.75 | 37.35 | 70.30 | 0.86 | 0.29 | 66.70 |
| 2.6.2 | 2020 | DSA [79] | 93.12 | 92.20 | -0.92 | 125.75 | 40.99 | 67.40 | 0.86 | - | - |
| 2.5.2 | 2022 | FTWT [53] | 93.66 | 92.63 | -1.03 | 125.75 | 42.75 | 66.00 | 0.86 | - | - |
| 2.3.2 | 2021 | GDP-Guo [57] | 93.90 | 93.55 | -0.35 | 125.75 | 43.21 | 65.64 | 0.86 | - | - |
| 2.4.3 | 2022 | EKG [95] | 93.84 | 93.69 | -0.15 | 125.75 | 43.87 | 65.11 | 0.86 | - | - |
| 2.6.3 | 2022 | CCEP [59] | 93.48 | 93.24 | -0.24 | 125.75 | 46.00 | 63.42 | 0.86 | - | - |
| 2.5.2 | 2021 | ManiDP [81] | 93.70 | 93.64 | -0.06 | 125.75 | 47.28 | 62.40 | 0.86 | - | - |
| 2.1.2 | 2022 | EPruner [47] | 93.26 | 93.18 | -0.08 | 125.75 | 48.63 | 61.33 | 0.86 | 0.39 | 54.12 |
| 2.4.3 | 2019 | C-SGD [96] | 93.39 | 93.44 | 0.05 | 125.75 | 49.23 | 60.85 | 0.86 | - | - |
| 2.3.3 | 2021 | GREG [97] | 93.36 | 93.36 | 0.00 | 125.75 | 49.32 | 60.78 | 0.86 | - | - |
| 2.3.3 | 2021 | GREG [97] | 93.36 | 93.18 | -0.18 | 125.75 | 49.32 | 60.78 | 0.86 | - | - |
| 2.3.2 | 2019 | GAL [63] | 93.26 | 91.58 | -1.68 | 125.75 | 50.09 | 60.16 | 0.86 | 0.29 | 65.88 |
| 2.3.2 | 2019 | GAL [63] | 93.26 | 90.36 | -2.90 | 125.75 | 50.09 | 60.16 | 0.86 | 0.29 | 65.88 |
| 2.3.1 | 2019 | GBN [94] | - | - | -0.33 | 125.75 | 50.17 | 60.10 | 0.86 | 0.40 | 53.50 |
| 2.7.2 | 2021 | EDP [40] | 93.61 | 93.61 | 0.00 | 125.75 | 53.18 | 57.71 | 0.86 | 0.39 | 54.18 |
| 2.1.2 | 2022 | CLR-RNF [49] | 93.26 | 93.27 | 0.01 | 125.75 | 53.70 | 57.30 | 0.86 | 0.38 | 55.50 |
| 2.3.2 | 2020 | SCOP [80] | 93.70 | 93.64 | -0.06 | 125.75 | 55.33 | 56.00 | 0.86 | 0.38 | 56.30 |
| 2.6.2 | 2021 | EE [61] | 93.62 | 93.68 | 0.06 | 125.75 | 55.33 | 56.00 | 0.86 | - | - |
| 2.3.2 | 2022 | WhiteBox [46] | 93.26 | 93.54 | 0.28 | 125.75 | 55.83 | 55.60 | 0.86 | - | - |
| 2.6.1 | 2022 | RL-MCTS [41] | 93.20 | 93.56 | 0.36 | 125.75 | 56.59 | 55.00 | 0.86 | - | - |
| 2.6.3 | 2020 | ABCPruner [50] | 93.26 | 93.23 | -0.03 | 125.75 | 57.68 | 54.13 | 0.86 | 0.39 | 54.12 |
| 2.6.1 | 2022 | GNN-RL [85] | 93.39 | 93.49 | 0.10 | 125.75 | 57.84 | 54.00 | 0.86 | - | - |
| 2.5.2 | 2022 | FTWT [53] | 93.66 | 92.28 | -1.38 | 125.75 | 57.84 | 54.00 | 0.86 | - | - |
| 2.1.2 | 2021 | SRR [87] | 93.38 | 93.75 | 0.37 | 125.75 | 58.10 | 53.80 | 0.86 | - | - |
| 2.3.2 | 2021 | GDP-Guo [57] | 93.90 | 93.97 | 0.07 | 125.75 | 58.66 | 53.35 | 0.86 | - | - |
| 2.3.2 | 2020 | LeGR [98] | 93.90 | 93.70 | -0.20 | 125.75 | 59.10 | 53.00 | 0.86 | - | - |
| 2.3.2 | 2021 | ResRep [93] | 93.71 | 93.71 | 0.00 | 125.75 | 59.22 | 52.91 | 0.86 | - | - |
| 2.6.2 | 2020 | LFPC [92] | 93.59 | 93.34 | -0.25 | 125.75 | 59.23 | 52.90 | 0.86 | - | - |
| 2.6.2 | 2020 | LFPC [92] | 93.59 | 93.24 | -0.35 | 125.75 | 59.23 | 52.90 | 0.86 | - | - |
| 2.6.2 | 2019 | TAS [88] | 94.46 | 93.69 | -0.77 | 125.75 | 59.48 | 52.70 | 0.86 | - | - |
| 2.6.2 | 2022 | MFP [91] | 93.59 | 93.56 | -0.03 | 125.75 | 59.61 | 52.60 | 0.86 | - | - |
| 2.6.2 | 2022 | MFP [91] | 93.59 | 92.76 | -0.83 | 125.75 | 59.61 | 52.60 | 0.86 | - | - |
| 2.5.1 | 2018 | SFP [90] | 93.59 | 93.35 | -0.24 | 125.75 | 59.61 | 52.60 | 0.86 | - | - |
| 2.5.1 | 2018 | SFP [90] | 93.59 | 92.26 | -1.33 | 125.75 | 59.61 | 52.60 | 0.86 | - | - |
| 2.4.1 | 2019 | CCP [99] | 93.50 | - | 0.08 | 125.75 | 59.61 | 52.60 | 0.86 | - | - |
| 2.2.1 | 2021 | LRMF [68] | 93.59 | 93.25 | -0.34 | 125.75 | 59.61 | 52.60 | 0.86 | - | - |
| 2.2.1 | 2021 | LRMF [68] | 93.59 | 93.29 | -0.30 | 125.75 | 59.61 | 52.60 | 0.86 | - | - |
| 2.1.2 | 2019 | FPGM [82] | 93.59 | 93.49 | -0.10 | 125.75 | 59.61 | 52.60 | 0.86 | - | - |
| 2.2.3 | 2022 | DLRFC [44] | 93.06 | 93.57 | -0.51 | 125.75 | 59.63 | 52.58 | 0.86 | 0.38 | 55.63 |
| 2.6.2 | 2020 | DSA [79] | 93.12 | 92.91 | -0.21 | 125.75 | 60.11 | 52.20 | 0.86 | - | - |
| 2.7.2 | 2021 | CC [60] | 93.33 | 93.64 | 0.31 | 125.75 | 60.36 | 52.00 | 0.86 | 0.45 | 48.24 |
| 2.7.3 | 2022 | SOKS [54] | 93.06 | 93.08 | 0.02 | 125.75 | 60.70 | 51.73 | 0.86 | 0.39 | 54.12 |
| 2.3.1 | 2020 | SCP [58] | 93.69 | 93.23 | 0.46 | 125.75 | 60.99 | 51.50 | 0.86 | 0.44 | 48.47 |
| 2.6.2 | 2022 | DDNP [100] | 93.62 | 93.83 | 0.21 | 125.75 | 61.62 | 51.00 | 0.86 | - | - |
| 2.3.1 | 2020 | EagleEye [101] | - | 94.66 | - | 125.75 | 62.23 | 50.51 | 0.86 | - | - |
| 2.5.1 | 2021 | DCP-CAC [76] | 92.88 | 93.10 | 0.22 | 125.75 | 62.82 | 50.04 | 0.86 | 0.44 | 49.41 |
| 2.2.1 | 2020 | HRank [45] | 93.26 | 93.17 | -0.09 | 125.75 | 62.85 | 50.02 | 0.86 | 0.50 | 42.35 |
| 2.3.2 | 2020 | DMC [102] | 93.62 | 93.69 | 0.07 | 125.75 | 62.88 | 50.00 | 0.86 | - | - |
| 2.6.1 | 2018 | AMC [103] | 92.80 | 90.10 | -2.70 | 125.75 | 62.88 | 50.00 | 0.86 | - | - |
| 2.4.3 | 2022 | EKG [95] | 93.84 | 94.09 | 0.25 | 125.75 | 62.88 | 50.00 | 0.86 | - | - |
| 2.6.1 | 2021 | AGMC [86] | 93.39 | 92.76 | -0.63 | 125.75 | 62.88 | 50.00 | 0.86 | - | - |
| 2.7.2 | 2020 | Hinge [66] | 92.95 | 93.69 | 0.74 | 125.75 | 62.88 | 50.00 | 0.86 | 0.42 | 51.27 |
| 2.6.1 | 2018 | AMC [103] | 92.80 | 91.90 | -0.90 | 125.75 | 62.88 | 50.00 | 0.86 | - | - |
| 2.4.3 | 2022 | RollBack [104] | - | 92.53 | - | 125.75 | 62.88 | 50.00 | 0.86 | - | - |
| 2.6.1 | 2022 | DECORE [35] | 93.26 | 93.26 | 0.00 | 125.75 | 63.06 | 49.85 | 0.86 | 0.44 | 49.41 |
| 2.2.3 | 2018 | DCP [74] | 93.80 | 93.49 | -0.31 | 125.75 | 63.19 | 49.75 | 0.86 | 0.44 | 49.24 |
| 2.7.1 | 2022 | RRCP [105] | - | 93.48 | - | 125.75 | 64.17 | 48.97 | 0.86 | 0.47 | 44.92 |
| 2.2.1 | 2022 | GCNP [52] | 93.72 | 93.85 | 0.13 | 125.75 | 65.00 | 48.31 | 0.86 | 0.56 | 35.01 |
| 2.5.1 | 2021 | SEP [73] | 93.04 | 93.85 | 0.81 | 125.75 | 65.99 | 47.52 | 0.86 | 0.62 | 27.91 |
| 2.2.1 | 2021 | CHIP [42] | 93.26 | 94.16 | 0.90 | 125.75 | 66.14 | 47.40 | 0.86 | 0.49 | 42.80 |
| 2.6.2 | 2020 | LFPC [92] | 93.59 | 93.72 | 0.13 | 125.75 | 66.52 | 47.10 | 0.86 | - | - |
| Continue on the next page... | | | | | | | | | | | |

TABLE 6: ResNet-56 on CIFAR-10. (Part 1)

| Section | Year | Method | Baseline Top-1 Acc.(%) | Pruned Top-1 Acc.(%) | Top-1 Acc. ↓(%) | Baseline FLOPs (M) | Pruned FLOPs (M) | FLOPs ↓(%) | Baseline Params (M) | Pruned Params (M) | Params ↓(%) |
|---|---|---|---|---|---|---|---|---|---|---|---|
| ResNet-56 (cont.) | | | | | | | | | | | |
| 2.6.2 | 2020 | LFPC [92] | 93.59 | 93.56 | -0.03 | 125.75 | 66.52 | 47.10 | 0.86 | - | - |
| 2.2.3 | 2018 | DCP [74] | 93.80 | 93.81 | 0.01 | 125.75 | 66.53 | 47.09 | 0.86 | 0.26 | 70.33 |
| 2.4.1 | 2019 | CCP [99] | 93.50 | - | -0.19 | 125.75 | 66.65 | 47.00 | 0.86 | - | - |
| 2.3.1 | 2020 | PR [62] | 93.80 | 93.83 | -0.03 | 125.75 | 66.65 | 47.00 | 0.86 | - | - |
| 2.3.2 | 2021 | ABP [39] | 93.41 | 93.10 | -0.31 | 125.75 | 68.91 | 45.20 | 0.86 | 0.47 | 45.70 |
| 2.2.3 | 2018 | NISP [106] | - | - | 0.03 | 125.75 | 70.91 | 43.61 | 0.86 | 0.49 | 42.60 |
| 2.7.3 | 2022 | GKP-TMI [89] | 93.78 | 94.00 | 0.22 | 125.75 | 71.39 | 43.23 | 0.86 | 0.49 | 43.49 |
| 2.7.2 | 2021 | CC [60] | 93.33 | 93.87 | 0.54 | 125.75 | 72.43 | 42.40 | 0.86 | 0.55 | 36.47 |
| 2.5.1 | 2018 | SFP [90] | 93.59 | 93.78 | 0.19 | 125.75 | 74.07 | 41.10 | 0.86 | - | - |
| 2.5.1 | 2018 | SFP [90] | 93.59 | 93.10 | -0.49 | 125.75 | 74.07 | 41.10 | 0.86 | - | - |
| 2.4.1 | 2022 | HAP [107] | 93.88 | 93.55 | 0.33 | 125.75 | 74.57 | 40.70 | 0.86 | - | - |
| 2.3.2 | 2019 | GAL [63] | 93.26 | 93.38 | 0.12 | 125.75 | 78.46 | 37.60 | 0.86 | 0.76 | 11.76 |
| 2.3.2 | 2019 | GAL [63] | 93.26 | 92.98 | -0.28 | 125.75 | 78.46 | 37.60 | 0.86 | 0.76 | 11.76 |
| 2.7.3 | 2022 | SOKS [54] | 93.06 | 93.22 | 0.16 | 125.75 | 80.60 | 35.91 | 0.86 | 0.52 | 40.00 |
| 2.7.3 | 2019 | SDN [67] | 93.03 | 93.29 | 0.26 | 125.75 | 81.99 | 34.80 | 0.86 | 0.50 | 42.30 |
| 2.5.1 | 2021 | SEP [73] | 93.04 | 94.05 | 1.01 | 125.75 | 82.89 | 34.08 | 0.86 | 0.62 | 27.91 |
| 2.5.1 | 2021 | DCP-CAC [76] | 92.88 | 93.47 | 0.59 | 125.75 | 88.01 | 30.01 | 0.86 | 0.60 | 30.59 |
| 2.3.2 | 2020 | LeGR [98] | 93.90 | 94.10 | 0.20 | 125.75 | 88.02 | 30.00 | 0.86 | - | - |
| 2.2.1 | 2020 | HRank [45] | 93.26 | 93.52 | 0.26 | 125.75 | 88.90 | 29.30 | 0.86 | 0.72 | 16.47 |
| 2.6.2 | 2020 | DSA [79] | 93.12 | 93.08 | -0.04 | 125.75 | 88.91 | 29.30 | 0.86 | - | - |
| 2.1.1 | 2017 | PFEC [69] | 93.04 | 93.06 | 0.02 | 125.75 | 91.04 | 27.60 | 0.86 | 0.74 | 13.73 |
| 2.6.1 | 2022 | DECORE [35] | 93.26 | 93.34 | 0.08 | 125.75 | 92.67 | 26.30 | 0.86 | 0.65 | 24.71 |
| 2.4.1 | 2022 | HAP [107] | 93.88 | 92.92 | 0.96 | 125.75 | 95.70 | 23.90 | 0.86 | - | - |
| 2.4.2 | 2019 | VP [65] | 93.04 | 92.26 | -0.78 | 125.75 | 100.37 | 20.18 | 0.86 | 0.69 | 19.30 |
| 2.1.1 | 2017 | PFEC [69] | 93.04 | 93.10 | 0.06 | 125.75 | 112.67 | 10.40 | 0.86 | 0.78 | 9.41 |

TABLE 6: ResNet-56 on CIFAR-10. (Part 2)

| Section | Year | Method | Baseline Top-1 Acc.(%) | Pruned Top-1 Acc.(%) | Top-1 Acc. ↓(%) | Baseline FLOPs (M) | Pruned FLOPs (M) | FLOPs ↓(%) | Baseline Params (M) | Pruned Params (M) | Params ↓(%) |
|---|---|---|---|---|---|---|---|---|---|---|---|
| ResNet-110 | | | | | | | | | | | |
| 2.2.3 | 2020 | PFP [37] | 93.57 | 93.21 | -0.36 | 253.15 | 25.92 | 89.76 | 1.73 | 0.14 | 92.07 |
| 2.6.1 | 2022 | DECORE [35] | 93.50 | 92.71 | -0.79 | 253.15 | 58.43 | 76.92 | 1.73 | 0.35 | 79.65 |
| 2.2.1 | 2021 | CHIP [42] | 93.50 | 93.63 | 0.13 | 253.15 | 71.89 | 71.60 | 1.73 | 0.55 | 68.30 |
| 2.2.1 | 2020 | HRank [45] | 93.50 | 92.65 | -0.85 | 253.15 | 79.38 | 68.64 | 1.73 | 0.53 | 69.19 |
| 2.5.2 | 2021 | DDG [83] | 94.25 | 93.87 | -0.38 | 253.15 | 83.03 | 67.20 | 1.73 | - | - |
| 2.6.3 | 2022 | CCEP [59] | 93.68 | 93.46 | -0.22 | 253.15 | 83.31 | 67.09 | 1.73 | - | - |
| 2.3.2 | 2022 | WhiteBox [46] | 93.50 | 94.12 | 0.62 | 253.15 | 86.07 | 66.00 | 1.73 | - | - |
| 2.1.2 | 2022 | CLR-RNF [49] | 93.57 | 93.71 | 0.14 | 253.15 | 86.07 | 66.00 | 1.73 | 0.53 | 69.10 |
| 2.1.2 | 2022 | EPruner [47] | 93.50 | 93.62 | 0.12 | 253.15 | 86.31 | 65.91 | 1.73 | 0.41 | 76.30 |
| 2.6.3 | 2020 | ABCPruner [50] | 93.50 | 93.58 | 0.08 | 253.15 | 88.49 | 65.04 | 1.73 | 0.56 | 67.63 |
| 2.2.1 | 2021 | LRMF [68] | 93.68 | 93.61 | -0.07 | 253.15 | 94.17 | 62.80 | 1.73 | - | - |
| 2.6.2 | 2022 | MFP [91] | 93.68 | 93.38 | -0.30 | 253.15 | 94.17 | 62.80 | 1.73 | - | - |
| 2.6.1 | 2022 | DECORE [35] | 93.50 | 93.50 | 0.00 | 253.15 | 96.76 | 61.78 | 1.73 | 0.61 | 64.53 |
| 2.4.3 | 2019 | C-SGD [96] | 94.38 | 94.44 | 0.06 | 253.15 | 99.01 | 60.89 | 1.73 | - | - |
| 2.6.2 | 2020 | LFPC [92] | 93.68 | 93.07 | -0.61 | 253.15 | 100.50 | 60.30 | 1.73 | - | - |
| 2.6.2 | 2020 | LFPC [92] | 93.68 | 93.79 | 0.11 | 253.15 | 100.50 | 60.30 | 1.73 | - | - |
| 2.6.2 | 2022 | DAIS [84] | 95.62 | 95.02 | -0.60 | 253.15 | 101.26 | 60.00 | 1.73 | - | - |
| 2.3.2 | 2021 | ResRep [93] | 94.64 | 94.62 | 0.02 | 253.15 | 105.79 | 58.21 | 1.73 | - | - |
| 2.5.2 | 2021 | DDG [83] | 94.25 | 94.33 | 0.08 | 253.15 | 111.13 | 56.10 | 1.73 | - | - |
| 2.6.2 | 2022 | MFP [91] | 93.68 | 93.31 | -0.37 | 253.15 | 120.75 | 52.30 | 1.73 | - | - |
| 2.6.2 | 2022 | MFP [91] | 93.68 | 93.69 | 0.01 | 253.15 | 120.75 | 52.30 | 1.73 | - | - |
| 2.6.2 | 2019 | TAS [88] | 94.97 | 94.33 | -0.64 | 253.15 | 120.75 | 52.30 | 1.73 | - | - |
| 2.2.1 | 2021 | LRMF [68] | 93.68 | 93.88 | 0.20 | 253.15 | 120.75 | 52.30 | 1.73 | - | - |
| 2.1.2 | 2019 | FPGM [82] | 93.68 | 93.85 | 0.17 | 253.15 | 120.75 | 52.30 | 1.73 | - | - |
| 2.2.1 | 2021 | LRMF [68] | 93.68 | 93.73 | 0.05 | 253.15 | 120.75 | 52.30 | 1.73 | - | - |
| 2.2.1 | 2021 | CHIP [42] | 93.50 | 94.44 | 0.94 | 253.15 | 121.26 | 52.10 | 1.73 | 0.89 | 48.30 |
| 2.6.1 | 2022 | GNN-RL [85] | 93.68 | 94.31 | 0.63 | 253.15 | 121.51 | 52.00 | 1.73 | - | - |
| 2.5.1 | 2021 | DCP-CAC [76] | 93.38 | 93.86 | 0.48 | 253.15 | 126.45 | 50.02 | 1.73 | 0.89 | 48.84 |
| 2.6.1 | 2021 | AGMC [86] | 93.68 | 93.08 | -0.60 | 253.15 | 126.58 | 50.00 | 1.73 | - | - |
| 2.3.2 | 2019 | GAL [63] | 93.50 | 92.74 | -0.76 | 253.15 | 130.33 | 48.52 | 1.73 | 0.96 | 44.77 |
| 2.3.2 | 2019 | GAL [63] | 93.50 | 92.55 | -0.95 | 253.15 | 130.33 | 48.52 | 1.73 | 0.96 | 44.77 |
| 2.3.2 | 2021 | ABP [39] | 93.63 | 93.95 | 0.32 | 253.15 | 136.19 | 46.20 | 1.73 | 0.95 | 44.90 |
| 2.2.3 | 2018 | NISP [106] | - | - | 0.18 | 253.15 | 142.32 | 43.78 | 1.73 | 0.98 | 43.25 |
| 2.7.3 | 2022 | GKP-TMI [89] | 94.26 | 94.90 | 0.64 | 253.15 | 143.51 | 43.31 | 1.73 | 0.98 | 43.52 |
| 2.2.1 | 2020 | HRank [45] | 93.50 | 94.23 | 0.73 | 253.15 | 148.85 | 41.20 | 1.73 | 1.05 | 39.53 |
| 2.5.1 | 2018 | SFP [90] | 93.68 | 93.38 | -0.30 | 253.15 | 149.86 | 40.80 | 1.73 | - | - |
| 2.5.1 | 2018 | SFP [90] | 93.68 | 93.86 | 0.18 | 253.15 | 149.86 | 40.80 | 1.73 | - | - |
| 2.1.1 | 2017 | PFEC [69] | 93.53 | 93.30 | -0.23 | 253.15 | 155.43 | 38.60 | 1.73 | 1.17 | 32.40 |
| 2.4.2 | 2019 | VP [65] | 93.21 | 92.96 | -0.25 | 253.15 | 160.61 | 36.56 | 1.73 | 1.02 | 41.07 |
| 2.6.1 | 2022 | DECORE [35] | 93.50 | 93.88 | 0.38 | 253.15 | 163.47 | 35.43 | 1.73 | 1.12 | 35.47 |
| 2.5.1 | 2021 | DCP-CAC [76] | 93.38 | 94.11 | 0.73 | 253.15 | 176.97 | 30.09 | 1.73 | 1.23 | 29.07 |
| 2.5.1 | 2018 | SFP [90] | 93.68 | 93.93 | 0.25 | 253.15 | 181.76 | 28.20 | 1.73 | - | - |
| 2.3.2 | 2019 | GAL [63] | 93.50 | 93.59 | 0.09 | 253.15 | 205.91 | 18.66 | 1.73 | 1.66 | 4.07 |
| 2.3.2 | 2019 | GAL [63] | 93.50 | 92.55 | -0.95 | 253.15 | 205.91 | 18.66 | 1.73 | 1.66 | 4.07 |
| 2.1.1 | 2017 | PFEC [69] | 93.53 | 93.55 | 0.02 | 253.15 | 213.13 | 15.81 | 1.73 | 1.69 | 2.33 |
| ResNet-164 | | | | | | | | | | | |
| 2.4.3 | 2019 | C-SGD [96] | 94.83 | 94.75 | -0.08 | 247.65 | 96.81 | 60.91 | 1.70 | - | - |
| 2.7.2 | 2020 | Hinge [66] | 95.18 | 94.60 | -0.58 | 247.65 | 110.01 | 55.58 | 1.70 | 0.86 | 49.47 |
| 2.4.2 | 2019 | VP [65] | 93.58 | 93.16 | -0.42 | 247.65 | 126.21 | 49.04 | 1.70 | 0.74 | 56.55 |
| 2.3.1 | 2017 | NS [77] | 94.58 | 94.73 | 0.15 | 247.65 | 136.48 | 44.89 | 1.70 | 1.10 | 35.29 |
| 2.6.2 | 2019 | TAS [88] | 95.47 | 94.00 | -1.47 | 247.65 | 178.06 | 28.10 | 1.70 | - | - |
| 2.3.1 | 2017 | NS [77] | 94.58 | 94.92 | 0.34 | 247.65 | 189.09 | 23.65 | 1.70 | 1.44 | 15.29 |

TABLE 7: ResNet-110/164 on CIFAR-10.

| Section | Year | Method | Baseline Top-1 Acc.(%) | Pruned Top-1 Acc.(%) | Top-1 Acc. ↓(%) | Baseline FLOPs (M) | Pruned FLOPs (M) | FLOPs ↓(%) | Baseline Params (M) | Pruned Params (M) | Params ↓(%) |
|---|---|---|---|---|---|---|---|---|---|---|---|
| ResNet-18 | | | | | | | | | | | |
| 2.3.3 | 2019 | OICSR [108] | 94.46 | 92.44 | -2.02 | 557.21 | 30.65 | 94.50 | 11.17 | - | - |
| 2.4.3 | 2022 | StructADMM [36] | - | 93.80 | - | 557.21 | 45.67 | 91.80 | 11.17 | - | - |
| 2.6.1 | 2020 | AutoCompress [34] | 93.90 | 93.81 | -0.09 | 557.21 | 45.67 | 91.80 | 11.17 | - | - |
| 2.3.3 | 2019 | OICSR [108] | 94.46 | 94.27 | -0.19 | 557.21 | 75.22 | 86.50 | 11.17 | - | - |
| 2.6.2 | 2022 | DNCP [109] | 95.35 | 94.97 | -0.38 | 557.21 | 140.56 | 74.77 | 11.17 | - | - |
| 2.5.2 | 2022 | CDG [110] | 91.26 | 90.37 | -0.89 | 557.21 | 257.97 | 53.70 | 11.17 | - | - |
| 2.6.2 | 2022 | DNCP [109] | 95.35 | 95.31 | -0.04 | 557.21 | 314.25 | 43.60 | 11.17 | - | - |
| 2.3.3 | 2019 | OICSR [108] | 94.46 | 95.10 | 0.64 | 557.21 | 338.78 | 39.20 | 11.17 | - | - |
| 2.6.3 | 2022 | EDropout [111] | 92.81 | 90.96 | -1.85 | 557.21 | - | - | 11.17 | 5.49 | 50.89 |
| 2.7.1 | 2022 | EarlyCroP [70] | 91.50 | 91.00 | -0.50 | 557.21 | - | - | 11.17 | 0.55 | 95.10 |
| 2.7.3 | 2022 | DPP [71] | 93.60 | 93.14 | -0.46 | 557.21 | - | - | 11.17 | 4.13 | 63.06 |
| ResNet-34 | | | | | | | | | | | |
| 2.6.3 | 2022 | EDropout [111] | 92.80 | 88.21 | -4.59 | 1160 | - | - | 21.60 | 8.42 | 61.03 |
| ResNet-50 | | | | | | | | | | | |
| 2.3.3 | 2021 | OTO [38] | 93.50 | 94.40 | 0.90 | 1307 | 167.30 | 87.20 | 23.52 | 2.07 | 91.20 |
| 2.6.3 | 2022 | EDropout [111] | 92.21 | 85.30 | -6.91 | 1307 | - | - | 23.52 | 10.91 | 53.62 |
| ResNet-101 | | | | | | | | | | | |
| 2.7.1 | 2020 | EB [64] | - | 92.49 | - | 7595.00 | 3452.27 | 54.55 | 44.54 | 13.36 | 70.00 |
| 2.7.1 | 2020 | EB [64] | - | 93.90 | - | 7595.00 | 5425.00 | 28.57 | 44.54 | 22.27 | 50.00 |
| 2.7.1 | 2020 | EB [64] | - | 93.91 | - | 7595.00 | 6904.55 | 9.09 | 44.54 | 31.18 | 30.00 |
| 2.6.3 | 2022 | EDropout [111] | 92.66 | 86.57 | -6.09 | 2520.00 | - | - | 42.51 | 19.20 | 54.82 |

TABLE 8: ResNet-18/34/50/101 on CIFAR-10.

| Section | Year | Method | Baseline Top-1 Acc.(%) | Pruned Top-1 Acc.(%) | Top-1 Acc. ↓(%) | Baseline FLOPs (M) | Pruned FLOPs (M) | FLOPs ↓(%) | Baseline Params (M) | Pruned Params (M) | Params ↓(%) |
|---|---|---|---|---|---|---|---|---|---|---|---|
| MobileNet-V1 | | | | | | | | | | | |
| 2.3.1 | 2020 | EagleEye [101] | - | 88.01 | - | 34.20 | 3.30 | 90.35 | 4.20 | - | - |
| 2.5.2 | 2022 | FTWT [53] | 90.89 | 91.21 | 0.32 | 34.20 | 7.52 | 78.00 | 4.20 | - | - |
| 2.6.2 | 2022 | DAIS [84] | 90.87 | 91.87 | 1.00 | 34.20 | 11.42 | 66.60 | 4.20 | - | - |
| 2.3.1 | 2020 | EagleEye [101] | - | 91.44 | - | 34.20 | 12.10 | 64.62 | 4.20 | - | - |
| 2.2.3 | 2018 | DCP [74] | 93.96 | 94.37 | 0.41 | 34.20 | 19.54 | 42.86 | 4.20 | 2.94 | 30.07 |
| 2.3.1 | 2020 | EagleEye [101] | - | 91.89 | - | 34.20 | 26.50 | 22.51 | 4.20 | - | - |
| 2.7.3 | 2022 | DPP [71] | 93.78 | 93.96 | 0.18 | 34.20 | - | - | 4.20 | 1.57 | 62.70 |
| MobileNet-V2 | | | | | | | | | | | |
| 2.4.3 | 2022 | StructADMM [36] | - | 95.10 | - | 89.00 | 22.82 | 74.36 | 3.50 | - | - |
| 2.6.2 | 2022 | DNCP [109] | 94.15 | 93.30 | -0.85 | 89.00 | 25.00 | 71.91 | 3.50 | - | - |
| 2.4.3 | 2022 | EKG [95] | 94.21 | 94.52 | 0.31 | 89.00 | 44.50 | 50.00 | 3.50 | - | - |
| 2.3.2 | 2021 | GDP-Guo [57] | 94.89 | 95.15 | 0.26 | 89.00 | 47.86 | 46.22 | 3.50 | - | - |
| 2.6.2 | 2022 | DDNP [100] | 94.58 | 94.81 | 0.23 | 89.00 | 50.73 | 43.00 | 3.50 | - | - |
| 2.6.2 | 2022 | DNCP [109] | 94.15 | 93.71 | -0.44 | 89.00 | 52.00 | 41.57 | 3.50 | - | - |
| 2.3.2 | 2020 | SCOP [80] | 94.48 | 94.24 | -0.24 | 89.00 | 53.13 | 40.30 | 3.50 | 2.24 | 36.10 |
| 2.3.2 | 2020 | DMC [102] | 94.23 | 94.49 | 0.26 | 89.00 | 53.40 | 40.00 | 3.50 | - | - |
| 2.3.2 | 2022 | WhiteBox [46] | 95.02 | 95.28 | 0.26 | 89.00 | 63.01 | 29.20 | 3.50 | - | - |
| 2.2.3 | 2018 | DCP [74] | 94.47 | 94.69 | 0.22 | 89.00 | 65.44 | 26.47 | 3.50 | 2.67 | 23.66 |

TABLE 9: MobileNet-V1/V2 on CIFAR-10.

| Section | Year | Method | Baseline Top-1 Acc.(%) | Pruned Top-1 Acc.(%) | Top-1 Acc. ↓(%) | Baseline FLOPs (M) | Pruned FLOPs (M) | FLOPs ↓(%) | Baseline Params (M) | Pruned Params (M) | Params ↓(%) |
|---|---|---|---|---|---|---|---|---|---|---|---|
| DenseNet-40 | | | | | | | | | | | |
| 2.3.2 | 2019 | GAL [63] | 94.81 | 93.23 | -1.58 | 283.00 | 80.91 | 71.41 | 1.04 | 0.26 | 75.00 |
| 2.3.2 | 2019 | GAL [63] | 94.81 | 91.90 | -2.91 | 283.00 | 80.91 | 71.41 | 1.04 | 0.26 | 75.00 |
| 2.3.1 | 2020 | SCP [58] | 94.39 | 93.77 | 0.62 | 283.00 | 82.72 | 70.77 | 1.04 | 0.26 | 75.41 |
| 2.2.1 | 2020 | HRank [45] | 94.81 | 93.68 | -1.13 | 283.00 | 110.54 | 60.94 | 1.04 | 0.48 | 53.85 |
| 2.7.2 | 2021 | CC [60] | 94.81 | 94.40 | -0.41 | 283.00 | 112.00 | 60.42 | 1.04 | 0.37 | 64.42 |
| 2.4.3 | 2019 | C-SGD [96] | 93.81 | 94.44 | 0.63 | 283.00 | 113.06 | 60.05 | 1.04 | - | - |
| 2.3.1 | 2017 | NS [77] | 93.89 | 94.35 | 0.46 | 283.00 | 127.43 | 54.97 | 1.04 | 0.36 | 65.69 |
| 2.3.2 | 2019 | GAL [63] | 94.81 | 94.50 | -0.31 | 283.00 | 128.15 | 54.72 | 1.04 | 0.45 | 56.73 |
| 2.3.2 | 2019 | GAL [63] | 94.81 | 93.53 | -1.28 | 283.00 | 128.15 | 54.72 | 1.04 | 0.45 | 56.73 |
| 2.6.1 | 2022 | DECORE [35] | 94.81 | 94.04 | -0.77 | 283.00 | 128.17 | 54.71 | 1.04 | 0.37 | 64.42 |
| 2.7.2 | 2021 | CC [60] | 94.81 | 94.67 | -0.14 | 283.00 | 150.00 | 47.00 | 1.04 | 0.50 | 51.92 |
| 2.4.2 | 2019 | VP [65] | 94.11 | 93.16 | -0.95 | 283.00 | 156.55 | 44.68 | 1.04 | 0.42 | 59.62 |
| 2.7.2 | 2020 | Hinge [66] | 94.74 | 94.67 | -0.07 | 283.00 | 157.35 | 44.40 | 1.04 | 0.75 | 27.54 |
| 2.2.1 | 2020 | HRank [45] | 94.81 | 94.24 | -0.57 | 283.00 | 168.00 | 40.63 | 1.04 | 0.66 | 36.54 |
| 2.6.1 | 2022 | DECORE [35] | 94.81 | 94.59 | -0.22 | 283.00 | 171.41 | 39.43 | 1.04 | 0.56 | 46.15 |
| 2.3.2 | 2019 | GAL [63] | 94.81 | 94.61 | -0.20 | 283.00 | 182.97 | 35.35 | 1.04 | 0.67 | 35.58 |
| 2.3.2 | 2019 | GAL [63] | 94.81 | 94.29 | -0.52 | 283.00 | 182.97 | 35.35 | 1.04 | 0.67 | 35.58 |
| 2.3.1 | 2017 | NS [77] | 93.89 | 94.81 | 0.92 | 283.00 | 202.29 | 28.52 | 1.04 | 0.67 | 35.29 |
| 2.4.1 | 2022 | SOSP [75] | 94.58 | 94.41 | -0.17 | 283.00 | 209.02 | 26.14 | 1.04 | 0.68 | 34.78 |
| 2.4.1 | 2022 | SOSP [75] | 94.58 | 94.42 | -0.16 | 283.00 | 220.66 | 22.03 | 1.04 | 0.71 | 32.21 |
| 2.6.1 | 2022 | DECORE [35] | 94.81 | 94.85 | 0.04 | 283.00 | 229.02 | 19.07 | 1.04 | 0.83 | 20.19 |
| GoogLeNet | | | | | | | | | | | |
| 2.6.1 | 2022 | DECORE [35] | 95.05 | 94.33 | -0.72 | 1535.00 | 232.27 | 84.87 | 6.17 | 0.86 | 86.02 |
| 2.6.1 | 2022 | DECORE [35] | 95.05 | 94.51 | -0.54 | 1535.00 | 333.26 | 78.29 | 6.17 | 1.17 | 80.98 |
| 2.2.1 | 2020 | HRank [45] | 95.05 | 94.07 | -0.98 | 1535.00 | 454.44 | 70.39 | 6.17 | 1.87 | 69.76 |
| 2.1.2 | 2022 | CLR-RNF [49] | 95.03 | 94.85 | -0.18 | 1535.00 | 492.73 | 67.90 | 6.17 | 2.18 | 64.70 |
| 2.1.2 | 2022 | EPruner [47] | 95.05 | 94.99 | -0.06 | 1535.00 | 501.02 | 67.36 | 6.17 | 2.22 | 64.02 |
| 2.6.3 | 2020 | ABCPruner [50] | 95.05 | 94.84 | -0.21 | 1535.00 | 513.34 | 66.56 | 6.17 | 2.46 | 60.13 |
| 2.7.2 | 2021 | CC [60] | 95.05 | 94.88 | -0.17 | 1535.00 | 616.02 | 59.87 | 6.17 | 2.27 | 63.25 |
| 2.2.1 | 2020 | HRank [45] | 95.05 | 94.53 | -0.52 | 1535.00 | 696.81 | 54.61 | 6.17 | 2.75 | 55.45 |
| 2.7.2 | 2021 | CC [60] | 95.05 | 95.18 | 0.13 | 1535.00 | 767.50 | 50.00 | 6.17 | 2.84 | 53.98 |
| 2.3.2 | 2019 | GAL [63] | 95.05 | 93.93 | -1.12 | 1535.00 | 949.28 | 38.16 | 6.17 | 3.13 | 49.27 |
| 2.3.2 | 2019 | GAL [63] | 95.05 | 94.56 | -0.49 | 1535.00 | 949.28 | 38.16 | 6.17 | 3.13 | 49.27 |
| 2.6.1 | 2022 | DECORE [35] | 95.05 | 95.20 | 0.15 | 1535.00 | 1232.04 | 19.74 | 6.17 | 4.75 | 23.09 |
| PreResNet-29 | | | | | | | | | | | |
| 2.4.1 | 2019 | ED [72] | 94.42 | 89.10 | -5.32 | 41.28 | 3.85 | 90.67 | 0.27 | 0.02 | 93.45 |
| 2.4.1 | 2019 | ED [72] | 94.42 | 93.80 | -0.62 | 41.28 | 15.22 | 63.13 | 0.27 | 0.08 | 70.09 |
| PreResNet-101 | | | | | | | | | | | |
| 2.7.1 | 2020 | EB [64] | - | 92.49 | - | 7595.00 | 3452.27 | 54.55 | 44.54 | 13.36 | 70.00 |
| 2.7.1 | 2020 | EB [64] | - | 93.90 | - | 7595.00 | 5425.00 | 28.57 | 44.54 | 22.27 | 50.00 |
| 2.7.1 | 2020 | EB [64] | - | 93.91 | - | 7595.00 | 6904.55 | 9.09 | 44.54 | 31.18 | 30.00 |
| ResNeXt-20 | | | | | | | | | | | |
| 2.7.2 | 2020 | Hinge [66] | 92.54 | 91.96 | -0.58 | 36.00 | 21.24 | 41.00 | 0.21 | 0.13 | 36.05 |
| ResNeXt-164 | | | | | | | | | | | |
| 2.7.2 | 2020 | Hinge [66] | 95.18 | 94.87 | -0.31 | 210.00 | 93.28 | 55.58 | 1.70 | 0.86 | 49.47 |
| NAS | | | | | | | | | | | |
| 2.6.2 | 2022 | ReCNAS [112] | - | 97.48 | - | - | 300 | - | - | 4.10 | - |
| 2.6.2 | 2022 | ReCNAS [112] | - | 97.71 | - | - | 500 | - | - | 5.20 | - |

TABLE 10: Other models on CIFAR-10, including DenseNet-40, GoogLeNet, PreResNet-29/101, ResNeXt-20/164, and models searched by Neural Architecture Search (NAS).

| Section | Year | Method | Baseline Top-1 Acc.(%) | Pruned Top-1 Acc.(%) | Top-1 Acc. ↓(%) | Baseline FLOPs (M) | Pruned FLOPs (M) | FLOPs ↓(%) | Baseline Params (M) | Pruned Params (M) | Params ↓(%) |
|---|---|---|---|---|---|---|---|---|---|---|---|
| VGG-16 | | | | | | | | | | | |
| 2.3.1 | 2020 | SCP [58] | 73.51 | 69.96 | 3.55 | 314.59 | 65.31 | 79.24 | 14.73 | 0.88 | 94.01 |
| 2.2.1 | 2022 | GCNP [52] | 72.00 | 71.64 | -0.36 | 314.59 | 99.91 | 68.24 | 14.73 | 1.70 | 88.46 |
| 2.2.1 | 2022 | GCNP [52] | 72.00 | 72.00 | 0.00 | 314.59 | 136.59 | 56.58 | 14.73 | 2.87 | 80.49 |
| 2.3.1 | 2020 | SCP [58] | 73.51 | 73.86 | -0.35 | 314.59 | 152.73 | 51.45 | 14.73 | 2.93 | 80.14 |
| 2.2.3 | 2022 | DLRFC [44] | 73.54 | 74.09 | -0.55 | 314.59 | 178.06 | 43.40 | 14.73 | 2.58 | 82.50 |
| 2.1.2 | 2019 | COP [51] | 72.59 | 71.77 | -0.82 | 314.59 | 179.00 | 43.10 | 14.73 | 3.95 | 73.20 |
| 2.3.1 | 2020 | PR [62] | 73.83 | 74.25 | -0.42 | 314.59 | 179.32 | 43.00 | 14.73 | - | - |
| 2.2.1 | 2021 | LRMF [68] | - | - | 0.50 | 314.59 | 201.65 | 35.90 | 14.73 | - | - |
| 2.7.1 | 2020 | EB [64] | - | 71.28 | - | 314.59 | 209.73 | 33.33 | 14.73 | 7.37 | 50.00 |
| 2.7.3 | 2019 | SDN [67] | 72.38 | 73.43 | 1.05 | 314.59 | 211.09 | 32.90 | 14.73 | 3.11 | 78.90 |
| 2.4.2 | 2019 | VP [65] | 73.26 | 73.33 | 0.07 | 314.59 | 257.30 | 18.21 | 14.73 | 9.15 | 37.87 |
| 2.7.1 | 2020 | EB [64] | - | 72.17 | - | 314.59 | 262.16 | 16.67 | 14.73 | 10.31 | 30.00 |
| 2.7.1 | 2020 | EB [64] | - | 71.81 | - | 314.59 | 285.99 | 9.09 | 14.73 | 13.26 | 10.00 |
| 2.7.1 | 2022 | EarlyCroP [70] | - | 62.20 | - | 314.59 | - | - | 14.73 | 0.31 | 97.90 |
| 2.7.3 | 2022 | DPP [71] | 70.32 | 70.40 | 0.08 | 314.59 | - | - | 14.73 | 2.77 | 81.20 |
| VGG-19 | | | | | | | | | | | |
| 2.3.3 | 2021 | GREG [97] | 74.02 | 67.55 | -6.47 | 398.00.00 | 45.01 | 88.69 | 20.04 | - | - |
| 2.3.3 | 2021 | GREG [97] | 74.02 | 67.75 | -6.27 | 398.00 | 45.01 | 88.69 | 20.04 | - | - |
| 2.4.1 | 2019 | ED [72] | 73.08 | 64.91 | -8.17 | 398.00 | 45.17 | 88.65 | 20.04 | - | - |
| 2.4.1 | 2019 | ED [72] | 73.34 | 65.18 | -8.16 | 398.00 | 45.25 | 88.63 | 20.04 | - | - |
| 2.5.1 | 2021 | SEP [73] | 73.16 | 73.47 | 0.31 | 398.00 | 117.00 | 70.60 | 20.04 | 5.95 | 70.31 |
| 2.3.1 | 2020 | SCP [58] | 72.56 | 72.15 | 0.41 | 398.00 | 151.48 | 61.94 | 20.04 | 2.13 | 89.37 |
| 2.4.1 | 2022 | SOSP [75] | 73.45 | 73.11 | -0.34 | 398.00 | 192.59 | 51.61 | 20.04 | - | - |
| 2.4.1 | 2022 | SOSP [75] | 73.45 | 73.17 | -0.28 | 398.00 | 219.42 | 44.87 | 20.04 | - | - |
| 2.3.1 | 2020 | SCP [58] | 72.56 | 72.99 | -0.43 | 398.00 | 235.14 | 40.92 | 20.04 | 4.50 | 77.52 |
| 2.4.1 | 2019 | ED [72] | 73.08 | 73.01 | -0.07 | 398.00 | 248.91 | 37.46 | 20.04 | - | - |
| 2.4.1 | 2019 | ED [72] | 73.34 | 72.90 | -0.44 | 398.00 | 249.15 | 37.40 | 20.04 | - | - |
| 2.3.1 | 2017 | NS [77] | 73.26 | 73.48 | 0.22 | 398.00 | 250.19 | 37.14 | 20.04 | 4.99 | 75.10 |
| 2.5.1 | 2021 | DCP-CAC [76] | 70.17 | 72.28 | 2.11 | 398.00 | 251.00 | 36.93 | 20.04 | 8.93 | 55.44 |
| 2.7.1 | 2022 | ProsPr [78] | 72.50 | 72.29 | -0.21 | 398.00 | - | - | 20.04 | 4.01 | 80.00 |
| 2.7.1 | 2022 | ProsPr [78] | 73.50 | 71.12 | -2.38 | 398.00 | - | - | 20.04 | 2.00 | 90.00 |
| 2.7.1 | 2022 | ProsPr [78] | 74.50 | 68.03 | -6.47 | 398.00 | - | - | 20.04 | 1.00 | 95.00 |

TABLE 11: VGG-16/19 on CIFAR-100.

| Section | Year | Method | Baseline Top-1 Acc.(%) | Pruned Top-1 Acc.(%) | Top-1 Acc. ↓(%) | Baseline FLOPs (M) | Pruned FLOPs (M) | FLOPs ↓(%) | Baseline Params (M) | Pruned Params (M) | Params ↓(%) |
|---|---|---|---|---|---|---|---|---|---|---|---|
| ResNet-20 | | | | | | | | | | | |
| 2.7.2 | 2020 | Hinge [66] | 68.83 | 66.34 | -2.49 | 40.81 | 13.44 | 67.06 | 0.27 | 0.09 | 66.36 |
| 2.2.1 | 2022 | GCNP [52] | 68.38 | 66.14 | -2.24 | 40.81 | 16.38 | 59.87 | 0.27 | 0.14 | 49.42 |
| 2.6.2 | 2019 | TAS [88] | 68.69 | 68.90 | 0.21 | 40.81 | 22.45 | 45.00 | 0.27 | - | - |
| 2.2.1 | 2022 | GCNP [52] | 68.38 | 68.95 | 0.57 | 40.81 | 34.84 | 14.64 | 0.27 | 0.25 | 8.16 |
| ResNet-32 | | | | | | | | | | | |
| 2.4.1 | 2019 | ED [72] | 78.17 | 65.72 | -12.45 | 69.12 | 3.72 | 94.62 | 0.47 | - | - |
| 2.4.1 | 2022 | SOSP [75] | 76.80 | 75.33 | -1.47 | 69.12 | 17.78 | 74.28 | 0.47 | - | - |
| 2.4.1 | 2019 | ED [72] | 78.17 | 75.51 | -2.66 | 69.12 | 19.62 | 71.62 | 0.47 | - | - |
| 2.4.1 | 2022 | SOSP [75] | 76.80 | 75.52 | -1.28 | 69.12 | 19.63 | 71.60 | 0.47 | - | - |
| 2.5.1 | 2021 | DCP-CAC [76] | 68.48 | 68.11 | -0.37 | 69.12 | 34.49 | 50.10 | 0.47 | 0.25 | 47.83 |
| 2.6.2 | 2022 | DAIS [84] | 73.24 | 72.20 | -1.04 | 69.12 | 39.47 | 42.90 | 0.47 | - | - |
| 2.6.2 | 2019 | TAS [88] | 70.61 | 72.41 | 1.80 | 69.12 | 42.51 | 38.50 | 0.47 | - | - |
| 2.1.2 | 2019 | COP [51] | 68.74 | 68.29 | -0.45 | 69.12 | 45.48 | 34.20 | 0.47 | 0.30 | 35.20 |
| 2.5.1 | 2021 | DCP-CAC [76] | 68.48 | 69.51 | 1.03 | 69.12 | 48.35 | 30.04 | 0.47 | 0.35 | 26.09 |
| ResNet-44 | | | | | | | | | | | |
| 2.5.1 | 2021 | DCP-CAC [76] | 69.87 | 69.82 | -0.05 | 97.15 | 48.54 | 50.04 | 0.66 | 0.35 | 47.69 |
| 2.5.1 | 2021 | DCP-CAC [76] | 69.87 | 70.85 | 0.98 | 97.15 | 67.97 | 30.04 | 0.66 | 0.46 | 30.77 |
| ResNet-56 | | | | | | | | | | | |
| 2.3.3 | 2019 | OICSR [108] | 75.87 | 73.10 | -2.77 | 125.75 | 3.52 | 97.20 | 0.86 | - | - |
| 2.3.3 | 2019 | OICSR [108] | 75.87 | 75.75 | -0.12 | 125.75 | 17.35 | 86.20 | 0.86 | - | - |
| 2.6.2 | 2022 | DAIS [84] | 71.76 | 72.57 | 0.81 | 125.75 | 58.35 | 53.60 | 0.86 | - | - |
| 2.2.1 | 2021 | LRMF [68] | - | - | -0.44 | 125.75 | 59.61 | 52.60 | 0.86 | - | - |
| 2.2.1 | 2022 | GCNP [52] | 72.86 | 72.22 | -0.64 | 125.75 | 60.08 | 52.22 | 0.86 | 0.52 | 39.83 |
| 2.6.2 | 2020 | LFPC [92] | 70.25 | 70.83 | 0.58 | 125.75 | 60.86 | 51.60 | 0.86 | - | - |
| 2.6.2 | 2019 | TAS [88] | 73.18 | 72.25 | -0.93 | 125.75 | 61.24 | 51.30 | 0.86 | - | - |
| 2.5.1 | 2021 | DCP-CAC [76] | 70.90 | 70.10 | -0.80 | 125.75 | 62.82 | 50.04 | 0.86 | 0.44 | 49.41 |
| 2.4.2 | 2022 | EKG [95] | 72.62 | 72.93 | 0.31 | 125.75 | 62.88 | 50.00 | 0.86 | - | - |
| 2.2.1 | 2022 | GCNP [52] | 72.86 | 72.86 | 0.00 | 125.75 | 64.42 | 48.77 | 0.86 | 0.52 | 39.24 |
| 2.3.3 | 2019 | OICSR [108] | 75.87 | 76.23 | 0.36 | 125.75 | 77.34 | 38.50 | 0.86 | - | - |
| 2.7.3 | 2019 | SDN [67] | 70.01 | 69.78 | -0.23 | 125.75 | 77.59 | 38.30 | 0.86 | 0.55 | 36.10 |
| 2.5.1 | 2021 | DCP-CAC [76] | 70.90 | 71.31 | 0.41 | 125.75 | 88.01 | 30.01 | 0.86 | 0.60 | 30.59 |
| 2.2.3 | 2022 | DLRFC [44] | 71.14 | 71.41 | -0.27 | 125.75 | 93.68 | 25.50 | 0.86 | 0.64 | 25.90 |
| 2.3.1 | 2020 | PR [62] | 72.49 | 72.46 | 0.06 | 125.75 | 94.31 | 25.00 | 0.86 | - | - |
| ResNet-110 | | | | | | | | | | | |
| 2.6.2 | 2022 | DAIS [84] | 75.34 | 74.69 | -0.65 | 253.15 | 109.61 | 56.70 | 1.73 | - | - |
| 2.6.2 | 2019 | TAS [88] | 75.06 | 73.16 | -1.90 | 253.15 | 119.99 | 52.60 | 1.73 | - | - |
| 2.5.1 | 2021 | DCP-CAC [76] | 72.53 | 72.16 | -0.37 | 253.15 | 126.52 | 50.02 | 1.73 | 0.89 | 48.84 |
| 2.5.1 | 2021 | DCP-CAC [76] | 72.53 | 72.79 | 0.26 | 253.15 | 176.97 | 30.09 | 1.73 | 1.23 | 29.07 |
| ResNet-164 | | | | | | | | | | | |
| 2.3.1 | 2020 | SCP [58] | 77.24 | 75.05 | 2.19 | 247.65 | 86.85 | 64.93 | 1.70 | 0.79 | 53.30 |
| 2.3.1 | 2017 | NS [77] | 76.63 | 76.09 | -0.54 | 247.65 | 122.34 | 50.60 | 1.70 | 1.19 | 30.06 |
| 2.3.1 | 2020 | SCP [58] | 77.24 | 76.62 | 0.62 | 247.65 | 135.32 | 45.36 | 1.70 | 1.21 | 28.89 |
| 2.7.2 | 2020 | Hinge [66] | 76.78 | 76.88 | 0.10 | 247.65 | 137.00 | 44.68 | 1.70 | 1.30 | 23.43 |
| 2.3.1 | 2017 | NS [77] | 76.63 | 77.13 | 0.50 | 247.65 | 164.93 | 33.40 | 1.70 | 1.43 | 15.61 |
| 2.6.2 | 2019 | TAS [88] | 78.29 | 77.76 | -0.53 | 247.65 | 171.13 | 30.90 | 1.70 | - | - |
| 2.4.2 | 2019 | VP [65] | 75.56 | 73.76 | -1.80 | 247.65 | 180.18 | 27.24 | 1.70 | 1.40 | 17.86 |

TABLE 12: ResNet-20/32/44/56/110/164 on CIFAR-100.

| Section | Year | Method | Baseline Top-1 Acc.(%) | Pruned Top-1 Acc.(%) | Top-1 Acc. ↓(%) | Baseline FLOPs (M) | Pruned FLOPs (M) | FLOPs ↓(%) | Baseline Params (M) | Pruned Params (M) | Params ↓(%) |
|---|---|---|---|---|---|---|---|---|---|---|---|
| ResNet-18 | | | | | | | | | | | |
| 2.5.2 | 2022 | CDG [110] | 67.78 | 65.94 | -1.84 | 557.21 | 293.27 | 47.37 | 11.17 | - | - |
| 2.6.3 | 2022 | EDropout [111] | 69.03 | 67.06 | -1.97 | 557.21 | - | - | 11.17 | 5.39 | 51.79 |
| ResNet-34 | | | | | | | | | | | |
| 2.6.3 | 2022 | EDropout [111] | 69.96 | 64.79 | -5.17 | 1160.00 | - | - | 21.60 | 10.65 | 50.70 |
| ResNet-50 | | | | | | | | | | | |
| 2.6.2 | 2022 | DNCP [109] | 79.01 | 79.22 | 0.21 | 1307.00 | 739.62 | 43.41 | 23.52 | - | - |
| 2.6.3 | 2022 | EDropout [111] | 71.22 | 61.91 | -9.31 | 1307.00 | - | - | 23.52 | 10.82 | 54.01 |
| ResNet-101 | | | | | | | | | | | |
| 2.7.1 | 2020 | EB [64] | - | 72.29 | - | 7595.00 | 3038.00 | 60.00 | 44.54 | 13.36 | 70.00 |
| 2.7.1 | 2020 | EB [64] | - | 73.15 | - | 7595.00 | 5425.00 | 28.57 | 44.54 | 22.27 | 50.00 |
| 2.7.1 | 2020 | EB [64] | - | 73.52 | - | 7595.00 | 6329.17 | 16.67 | 44.54 | 31.18 | 30.00 |
| 2.6.3 | 2022 | EDropout [111] | 71.19 | 61.92 | -9.27 | 2520.00 | - | - | 42.51 | 18.56 | 56.34 |

TABLE 13: ResNet-18/34/50/101 on CIFAR-100.

| Section | Year | Method | Baseline Top-1 Acc.(%) | Pruned Top-1 Acc.(%) | Top-1 Acc. ↓(%) | Baseline FLOPs (M) | Pruned FLOPs (M) | FLOPs ↓(%) | Baseline Params (M) | Pruned Params (M) | Params ↓(%) |
|---|---|---|---|---|---|---|---|---|---|---|---|
| MobileNet-V1 | | | | | | | | | | | |
| 2.7.3 | 2022 | DPP [71] | 72.35 | 72.50 | -0.15 | - | - | - | - | - | 60 |
| MobileNet-V2 | | | | | | | | | | | |
| 2.4.2 | 2022 | EKG [95] | 76.07 | 76.29 | 0.22 | 89.00 | 44.50 | 50.00 | 3.50 | - | - |
| 2.6.2 | 2022 | DNCP [109] | 75.91 | 75.69 | -0.22 | 89.00 | 52.00 | 41.57 | 3.50 | - | - |

TABLE 14: MobileNet-V1/V2 on CIFAR-100.

| Section | Year | Method | Baseline Top-1 Acc.(%) | Pruned Top-1 Acc.(%) | Top-1 Acc. ↓(%) | Baseline FLOPs (M) | Pruned FLOPs (M) | FLOPs ↓(%) | Baseline Params (M) | Pruned Params (M) | Params ↓(%) |
|---|---|---|---|---|---|---|---|---|---|---|---|
| DenseNet-40 | | | | | | | | | | | |
| 2.3.1 | 2020 | SCP [58] | 74.24 | 73.17 | 1.07 | 283.00 | 91.07 | 67.82 | 1.04 | 0.26 | 74.86 |
| 2.3.1 | 2017 | NS [77] | 74.64 | 74.28 | -0.36 | 283.00 | 149.20 | 47.28 | 1.04 | 0.45 | 56.60 |
| 2.3.1 | 2020 | SCP [58] | 74.24 | 73.84 | 0.40 | 283.00 | 152.11 | 46.25 | 1.04 | 0.47 | 55.22 |
| 2.3.1 | 2017 | NS [77] | 74.64 | 74.72 | 0.08 | 283.00 | 196.98 | 30.39 | 1.04 | 0.65 | 37.74 |
| 2.4.1 | 2022 | SOSP [75] | 74.11 | 73.46 | -0.65 | 283.00 | 198.16 | 29.98 | 1.04 | - | - |
| 2.4.1 | 2022 | SOSP [75] | 74.11 | 73.60 | -0.51 | 283.00 | 203.11 | 28.23 | 1.04 | - | - |
| 2.4.2 | 2019 | VP [65] | 74.64 | 72.19 | -2.45 | 283.00 | 218.77 | 22.70 | 1.04 | 0.65 | 37.50 |
| PreResNet-29 | | | | | | | | | | | |
| 2.4.1 | 2019 | ED [72] | 75.70 | 65.11 | -10.59 | 41.28 | 3.91 | 90.52 | 0.27 | - | - |
| 2.4.1 | 2019 | ED [72] | 75.70 | 73.62 | -2.08 | 41.28 | 15.33 | 62.86 | 0.27 | - | - |
| PreResNet-101 | | | | | | | | | | | |
| 2.7.1 | 2020 | EB [64] | - | 72.29 | - | 7595 | 3038.00 | 60.00 | 44.54 | 13.36 | 70 |
| 2.7.1 | 2020 | EB [64] | - | 73.15 | - | 7595 | 5425.00 | 28.57 | 44.54 | 22.27 | 50 |
| 2.7.1 | 2020 | EB [64] | - | 73.52 | - | 7595 | 6329.17 | 16.67 | 44.54 | 31.18 | 30 |
| ResNeXt-20 | | | | | | | | | | | |
| 2.7.2 | 2020 | Hinge [66] | 71.95 | 71.26 | -0.69 | 36 | 19.29 | 46.41 | 0.21 | 0.14 | 34.76 |
| ResNeXt-164 | | | | | | | | | | | |
| 2.7.2 | 2020 | Hinge [66] | 76.87 | 77.44 | 0.57 | 210 | 100.28 | 52.25 | 1.70 | 0.99 | 41.51 |

TABLE 15: Other models on CIFAR-100, including DenseNet-40, PreResNet-29/101, and ResNeXt-20/164.

| Section | Year | Method | Baseline Top-1 Acc.(%) | Pruned Top-1 Acc.(%) | Top-1 Acc. ↓(%) | Baseline FLOPs (M) | Pruned FLOPs (M) | FLOPs ↓(%) | Baseline Params (M) | Pruned Params (M) | Params ↓(%) |
|---|---|---|---|---|---|---|---|---|---|---|---|
| AlexNet | | | | | | | | | | | |
| 2.3.3 | 2019 | OICSR [108] | 56.98 | 53.78 | -3.20 | 7270.00 | 2304.59 | 68.30 | 17.69 | - | - |
| 2.2.3 | 2018 | NISP [106] | - | - | 1.43 | 7270.00 | 2337.30 | 67.85 | 17.69 | 11.72 | 33.77 |
| 2.5.1 | 2018 | GDP-Lin [113] | 56.60 | 54.82 | -1.78 | 7270.00 | 2621.26 | 63.94 | 17.69 | - | - |
| 2.2.3 | 2018 | NISP [106] | - | - | 0.97 | 7270.00 | 2712.44 | 62.69 | 17.69 | 17.34 | 1.96 |
| 2.3.3 | 2019 | OICSR [108] | 56.98 | 56.83 | -0.15 | 7270.00 | 3344.20 | 54.00 | 17.69 | - | - |
| 2.2.3 | 2018 | NISP [106] | - | - | 0.54 | 7270.00 | 3366.01 | 53.70 | 17.69 | 17.18 | 2.91 |
| 2.5.1 | 2018 | GDP-Lin [113] | 56.60 | 55.83 | -0.77 | 7270.00 | 3469.12 | 52.28 | 17.69 | - | - |
| 2.2.2 | 2019 | AOFP [48] | 55.71 | 56.17 | 0.46 | 7270.00 | 4268.31 | 41.29 | 17.69 | - | - |
| 2.2.3 | 2018 | NISP [106] | - | - | 0.00 | 7270.00 | 4353.28 | 40.12 | 17.69 | 9.36 | 47.09 |
| 2.5.1 | 2018 | GDP-Lin [113] | 56.60 | 56.46 | -0.14 | 7270.00 | 4535.16 | 37.62 | 17.69 | - | - |
| 2.2.2 | 2019 | AOFP [48] | 55.71 | 56.54 | 0.83 | 7270.00 | 5014.39 | 31.03 | 17.69 | - | - |
| 2.3.3 | 2019 | OICSR [108] | 56.98 | 57.87 | 0.89 | 7270.00 | 5568.82 | 23.40 | 17.69 | - | - |

TABLE 16: AlexNet on ImageNet-1K.

| Section | Year | Method | Baseline Top-1 Acc.(%) | Pruned Top-1 Acc.(%) | Top-1 Acc. ↓(%) | Baseline FLOPs (M) | Pruned FLOPs (M) | FLOPs ↓(%) | Baseline Params (M) | Pruned Params (M) | Params ↓(%) |
|---|---|---|---|---|---|---|---|---|---|---|---|
| VGG-11 | | | | | | | | | | | |
| 2.1.2 | 2019 | COP [51] | 63.60 | 62.38 | -1.22 | 7630.00 | 4211.76 | 44.80 | 132.86 | 22.32 | 83.20 |
| 2.6.2 | 2021 | EE [61] | 70.84 | 71.15 | 0.31 | 7630.00 | 5035.80 | 34.00 | 132.86 | - | - |
| 2.4.1 | 2019 | Mol-19 [114] | 70.84 | 70.65 | -0.19 | 7630.00 | 6948.21 | 8.94 | 132.86 | 31.79 | 76.07 |
| VGG-16 | | | | | | | | | | | |
| 2.6.1 | 2022 | GNN-RL [85] | 70.50 | 70.99 | 0.49 | 15500.00 | 3100.00 | 80.00 | 138.40 | - | - |
| 2.4.2 | 2019 | RBP [55] | - | 69.20 | - | 15500.00 | 3100.00 | 80.00 | 138.40 | - | - |
| 2.6.1 | 2018 | AMC [103] | 70.50 | 69.10 | -1.40 | 15500.00 | 3100.00 | 80.00 | 138.40 | - | - |
| 2.6.1 | 2021 | AGMC [86] | 70.50 | 70.35 | -0.15 | 15500.00 | 3100.00 | 80.00 | 138.40 | - | - |
| 2.5.1 | 2018 | GDP-Lin [113] | 70.32 | 67.51 | -2.81 | 15500.00 | 3800.00 | 75.48 | 138.40 | - | - |
| 2.6.2 | 2022 | DAIS [84] | - | - | 2.91 | 15500.00 | 3875.00 | 75.00 | 138.40 | - | - |
| 2.2.2 | 2017 | ThiNet [115] | 68.34 | 67.34 | -1.00 | 15500.00 | 4679.06 | 69.81 | 138.40 | 8.32 | 93.99 |
| 2.2.2 | 2017 | ThiNet [115] | 68.34 | 69.80 | 1.46 | 15500.00 | 4799.29 | 69.04 | 138.40 | 131.50 | 4.99 |
| 2.5.1 | 2019 | DSG [116] | - | 71.44 | - | 15500.00 | 5747.40 | 62.92 | 138.40 | - | - |
| 2.5.1 | 2018 | GDP-Lin [113] | 70.32 | 68.80 | -1.52 | 15500.00 | 6400.00 | 58.71 | 138.40 | - | - |
| 2.7.2 | 2021 | CC [60] | 71.59 | 68.81 | -2.78 | 15500.00 | 7379.52 | 52.39 | 138.40 | 8.37 | 93.95 |
| 2.5.1 | 2018 | GDP-Lin [113] | 70.32 | 69.88 | -0.44 | 15500.00 | 7500.00 | 51.61 | 138.40 | - | - |
| 2.6.2 | 2022 | DAIS [84] | - | - | 1.64 | 15500.00 | 7750.00 | 50.00 | 138.40 | - | - |
| 2.3.2 | 2018 | SSS [117] | - | 68.53 | - | 15500.00 | - | - | 138.40 | - | - |
| 2.7.1 | 2022 | EarlyCroP [70] | - | 61.43 | - | 15500.00 | - | - | 138.40 | 69.20 | 50.00 |
| 2.7.2 | 2020 | NM [118] | 73.36 | 61.18 | -12.18 | 15500.00 | - | - | 138.40 | 85.25 | 38.41 |
| 2.7.2 | 2020 | NM [118] | 73.36 | 54.13 | -19.23 | 15500.00 | - | - | 138.40 | 55.16 | 60.14 |
| 2.7.2 | 2020 | NM [118] | 73.36 | 43.99 | -29.37 | 15500.00 | - | - | 138.40 | 64.19 | 53.62 |

TABLE 17: VGG-11/16 on ImageNet-1K.

| Section | Year | Method | Baseline Top-1 Acc.(%) | Pruned Top-1 Acc.(%) | Top-1 Acc. ↓(%) | Baseline FLOPs (M) | Pruned FLOPs (M) | FLOPs ↓(%) | Baseline Params (M) | Pruned Params (M) | Params ↓(%) |
|---|---|---|---|---|---|---|---|---|---|---|---|
| ResNet-18 | | | | | | | | | | | |
| 2.5.2 | 2021 | DDG [83] | 69.76 | 69.38 | -0.38 | 1814.00 | 758.25 | 58.20 | 11.69 | - | - |
| 2.5.2 | 2021 | ManiDP [81] | 69.76 | 68.35 | -1.41 | 1814.00 | 814.49 | 55.10 | 11.69 | - | - |
| 2.7.3 | 2022 | SOKS [54] | 70.42 | 69.16 | -1.26 | 1814.00 | 817.30 | 54.95 | 11.69 | 6.27 | 46.36 |
| 2.7.3 | 2020 | SWP [56] | 69.76 | 69.72 | -0.04 | 1814.00 | 823.92 | 54.58 | 11.69 | - | - |
| 2.2.1 | 2021 | LRMF [68] | 70.28 | 67.87 | -2.41 | 1814.00 | 847.14 | 53.30 | 11.69 | - | - |
| 2.6.2 | 2022 | MFP [91] | 70.28 | 67.11 | -3.17 | 1814.00 | 874.35 | 51.80 | 11.69 | - | - |
| 2.5.2 | 2022 | FTWT [53] | 69.76 | 67.49 | -2.27 | 1814.00 | 878.70 | 51.56 | 11.69 | - | - |
| 2.6.1 | 2022 | GNN-RL [85] | 69.76 | 68.66 | -1.10 | 1814.00 | 888.86 | 51.00 | 11.69 | - | - |
| 2.5.2 | 2021 | ManiDP [81] | 69.76 | 68.88 | -0.88 | 1814.00 | 888.86 | 51.00 | 11.69 | - | - |
| 2.7.3 | 2020 | SWP [56] | 69.76 | 69.98 | 0.22 | 1814.00 | 898.29 | 50.48 | 11.69 | - | - |
| 2.4.3 | 2022 | EKG [95] | 70.38 | 69.39 | -0.99 | 1814.00 | 905.19 | 50.10 | 11.69 | - | - |
| 2.5.2 | 2019 | FBS [119] | 70.71 | 68.17 | -2.54 | 1814.00 | 916.16 | 49.49 | 11.69 | - | - |
| 2.5.2 | 2021 | DDG [83] | 69.76 | 70.12 | 0.36 | 1814.00 | 917.88 | 49.40 | 11.69 | - | - |
| 2.5.2 | 2020 | DRLP [120] | 69.76 | 68.73 | -1.03 | 1814.00 | 935.05 | 48.45 | 11.69 | - | - |
| 2.6.3 | 2020 | ABCPruner [50] | 69.66 | 67.80 | -1.86 | 1814.00 | 962.55 | 46.94 | 11.69 | 9.50 | 18.73 |
| 2.6.2 | 2021 | EE [61] | 70.28 | 68.27 | -2.01 | 1814.00 | 968.68 | 46.60 | 11.69 | - | - |
| 2.3.2 | 2020 | SCOP [80] | 69.76 | 68.62 | -1.14 | 1814.00 | 997.70 | 45.00 | 11.69 | 6.60 | 43.50 |
| 2.6.3 | 2020 | ABCPruner [50] | 69.66 | 67.28 | -2.38 | 1814.00 | 999.91 | 44.88 | 11.69 | 6.60 | 43.54 |
| 2.3.2 | 2021 | ABP [39] | 70.29 | 67.83 | -2.46 | 1814.00 | 1021.28 | 43.70 | 11.69 | 6.31 | 46.00 |
| 2.6.2 | 2022 | DAIS [84] | 65.36 | 67.56 | 2.20 | 1814 | 1028.54 | 43.30 | 11.69 | - | - |
| 2.1.2 | 2019 | COP [51] | 70.29 | 66.98 | -3.31 | 1814.00 | 1028.54 | 43.30 | 11.69 | 6.42 | 45.10 |
| 2.2.3 | 2020 | PFP [37] | 69.74 | 65.65 | -4.09 | 1814.00 | 1031.80 | 43.12 | 11.69 | 4.62 | 60.48 |
| 2.6.2 | 2022 | DNCP [109] | 70.10 | 69.20 | -0.90 | 1814.00 | 1048.09 | 42.22 | 11.69 | - | - |
| 2.6.2 | 2020 | DMCP [121] | 70.10 | 68.40 | -1.90 | 1814.00 | 1048.09 | 42.22 | 11.69 | - | - |
| 2.6.2 | 2022 | MFP [91] | 70.28 | 67.66 | -2.62 | 1814.00 | 1055.75 | 41.80 | 11.69 | - | - |
| 2.5.1 | 2018 | SFP [90] | 70.28 | 67.10 | -3.18 | 1814.00 | 1055.75 | 41.80 | 11.69 | - | - |
| 2.1.2 | 2019 | FPGM [82] | 70.28 | 68.41 | -1.87 | 1814.00 | 1055.75 | 41.80 | 11.69 | - | - |
| 2.6.2 | 2022 | MFP [91] | 70.28 | 68.31 | -1.97 | 1814.00 | 1055.75 | 41.80 | 11.69 | - | - |
| 2.6.2 | 2020 | DSA [79] | 69.72 | 68.61 | -1.11 | 1814.00 | 1088.40 | 40.00 | 11.69 | - | - |
| 2.3.2 | 2020 | SCOP [80] | 69.76 | 69.18 | -0.58 | 1814.00 | 1110.17 | 38.80 | 11.69 | 7.10 | 39.30 |
| 2.6.2 | 2019 | TAS [88] | 67.65 | 69.15 | 1.50 | 1814.00 | 1209.94 | 33.30 | 11.69 | - | - |
| 2.2.3 | 2020 | PFP [37] | 69.74 | 67.38 | -2.36 | 1814.00 | 1282.50 | 29.30 | 11.69 | 6.57 | 43.80 |
| 2.4.1 | 2022 | SOSP [75] | 69.76 | 68.78 | -0.98 | 1814.00 | 1287.94 | 29.00 | 11.69 | 6.43 | 45.00 |
| 2.5.1 | 2021 | DCP-CAC [76] | 68.46 | 68.17 | -0.29 | 1814.00 | 1310.11 | 27.78 | 11.69 | 7.49 | 35.94 |
| 2.7.1 | 2020 | EB [64] | 69.57 | 68.28 | -1.29 | 1814.00 | 1372.19 | 24.36 | 11.69 | 8.18 | 30.00 |
| 2.4.1 | 2022 | SOSP [75] | 69.76 | 69.63 | -0.13 | 1814.00 | 1378.64 | 24.00 | 11.69 | 7.13 | 39.00 |
| 2.7.2 | 2020 | DJPQ [122] | 69.74 | 69.27 | -0.47 | 1814.00 | 1393.07 | 23.20 | 11.69 | - | - |
| 2.2.3 | 2020 | PFP [37] | 69.74 | 68.66 | -1.08 | 1814.00 | 1451.38 | 19.99 | 11.69 | 8.06 | 31.03 |
| 2.5.1 | 2021 | DCP-CAC [76] | 68.46 | 68.62 | 0.16 | 1814.00 | 1612.44 | 11.11 | 11.69 | 9.44 | 19.22 |
| 2.7.1 | 2020 | EB [64] | 69.57 | 69.84 | 0.27 | 1814.00 | 1695.89 | 6.51 | 11.69 | 10.52 | 10.00 |
| 2.5.1 | 2022 | CHEX [123] | - | 69.60 | - | 1814.00 | - | - | 11.69 | - | - |
| 2.6.3 | 2022 | EDropout [111] | 68.48 | 65.43 | -3.05 | 1814.00 | - | - | 11.69 | 5.39 | 53.91 |

TABLE 18: ResNet-18 on ImageNet-1K.

| Section | Year | Method | Baseline Top-1 Acc.(%) | Pruned Top-1 Acc.(%) | Top-1 Acc. ↓(%) | Baseline FLOPs (M) | Pruned FLOPs (M) | FLOPs ↓(%) | Baseline Params (M) | Pruned Params (M) | Params ↓(%) |
|---|---|---|---|---|---|---|---|---|---|---|---|
| ResNet-34 | | | | | | | | | | | |
| 2.5.2 | 2021 | DDG [83] | 73.31 | 71.95 | -1.36 | 3663.00 | 1201.46 | 67.20 | 21.80 | - | - |
| 2.5.2 | 2021 | DDG [83] | 73.31 | 73.01 | -0.30 | 3663.00 | 1490.84 | 59.30 | 21.80 | - | - |
| 2.6.3 | 2020 | ABCPruner [50] | 73.28 | 70.45 | -2.83 | 3663.00 | 1503.10 | 58.97 | 21.80 | 10.47 | 51.96 |
| 2.7.3 | 2022 | SOKS [54] | 74.01 | 73.52 | -0.49 | 3663.00 | 1612.52 | 55.98 | 21.80 | 11.27 | 48.30 |
| 2.5.2 | 2021 | ManiDP [81] | 73.31 | 72.74 | -0.57 | 3663.00 | 1637.36 | 55.30 | 21.80 | - | - |
| 2.5.2 | 2022 | FTWT [53] | 73.30 | 71.71 | -1.59 | 3663.00 | 1749.45 | 52.24 | 21.80 | - | - |
| 2.3.2 | 2021 | ABP [39] | 73.86 | 72.15 | -1.71 | 3663.00 | 1923.08 | 47.50 | 21.80 | 10.75 | 50.70 |
| 2.5.2 | 2022 | FTWT [53] | 73.30 | 72.17 | -1.13 | 3663.00 | 1926.01 | 47.42 | 21.80 | - | - |
| 2.5.2 | 2021 | ManiDP [81] | 73.31 | 73.30 | -0.01 | 3663.00 | 1948.72 | 46.80 | 21.80 | - | - |
| 2.5.1 | 2022 | CHEX [123] | - | 73.50 | - | 3663.00 | 1990.76 | 45.65 | 21.80 | - | - |
| 2.4.3 | 2022 | EKG [95] | 73.85 | 73.51 | -0.34 | 3663.00 | 2010.99 | 45.10 | 21.80 | - | - |
| 2.3.2 | 2020 | SCOP [80] | 73.31 | 72.62 | -0.69 | 3663.00 | 2021.98 | 44.80 | 21.80 | 11.86 | 45.60 |
| 2.6.2 | 2022 | DDNP [100] | 73.31 | 73.03 | -0.28 | 3663.00 | 2043.95 | 44.20 | 21.80 | - | - |
| 2.2.3 | 2018 | NISP [106] | - | - | 0.92 | 3663.00 | 2060.07 | 43.76 | 21.80 | 12.28 | 43.68 |
| 2.2.2 | 2020 | GFS [124] | 73.40 | 71.90 | -1.50 | 3663.00 | 2060.44 | 43.75 | 21.80 | 14.70 | 32.57 |
| 2.3.2 | 2020 | DMC [102] | 73.30 | 72.57 | -0.73 | 3663.00 | 2073.26 | 43.40 | 21.80 | - | - |
| 2.6.3 | 2022 | CCEP [59] | 73.30 | 72.67 | -0.63 | 3663.00 | 2120.88 | 42.10 | 21.80 | - | - |
| 2.6.2 | 2022 | DAIS [84] | 72.23 | 72.77 | 0.54 | 3663.00 | 2128.20 | 41.90 | 21.80 | - | - |
| 2.5.1 | 2018 | SFP [90] | 73.92 | 71.83 | -2.09 | 3663.00 | 2157.51 | 41.10 | 21.80 | - | - |
| 2.1.2 | 2019 | FPGM [82] | 73.92 | 72.63 | -1.29 | 3663.00 | 2157.51 | 41.10 | 21.80 | - | - |
| 2.6.3 | 2020 | ABCPruner [50] | 73.28 | 70.98 | -2.30 | 3663.00 | 2161.19 | 41.00 | 21.80 | 10.07 | 53.79 |
| 2.3.2 | 2020 | SCOP [80] | 73.31 | 72.93 | -0.38 | 3663.00 | 2230.77 | 39.10 | 21.80 | 13.15 | 39.70 |
| 2.5.2 | 2022 | FTWT [53] | 73.30 | 72.79 | -0.51 | 3663.00 | 2279.48 | 37.77 | 21.80 | - | - |
| 2.2.2 | 2020 | GFS [124] | 73.40 | 73.50 | 0.10 | 3663.00 | 2627.80 | 28.26 | 21.80 | 17.20 | 21.10 |
| 2.2.3 | 2018 | NISP [106] | - | - | 0.28 | 3663.00 | 2662.27 | 27.32 | 21.80 | 15.88 | 27.14 |
| 2.5.2 | 2022 | FTWT [53] | 73.30 | 73.25 | -0.05 | 3663.00 | 2715.75 | 25.86 | 21.80 | - | - |
| 2.6.3 | 2022 | CCEP [59] | 73.30 | 73.64 | 0.34 | 3663.00 | 2731.13 | 25.44 | 21.80 | - | - |
| 2.3.3 | 2021 | GREG [97] | 73.31 | 73.61 | 0.30 | 3663.00 | 2775.00 | 24.24 | 21.80 | - | - |
| 2.1.1 | 2017 | PFEC [69] | 73.23 | 72.17 | -1.06 | 3663.00 | 2776.55 | 24.20 | 21.80 | 19.45 | 10.80 |
| 2.4.1 | 2019 | Mol-19 [114] | 73.31 | 72.83 | -0.48 | 3663.00 | 2847.88 | 22.25 | 21.80 | 17.20 | 21.10 |
| 2.7.1 | 2022 | PaT [125] | - | 73.50 | - | 3663.00 | 2897.55 | 20.90 | 21.80 | - | - |
| 2.1.1 | 2017 | PFEC [69] | 73.23 | 72.56 | -0.67 | 3663.00 | 3099.46 | 15.38 | 21.80 | 20.08 | 7.87 |
| 2.6.3 | 2022 | EDropout [111] | 73.42 | 71.82 | -1.60 | 3663.00 | - | - | 21.80 | 9.59 | 56.02 |
| 2.7.2 | 2020 | NM [118] | 73.31 | 66.77 | -6.54 | 3663.00 | - | - | 21.80 | 19.72 | 9.52 |
| 2.7.2 | 2020 | NM [118] | 73.31 | 55.70 | -17.61 | 3663.00 | - | - | 21.80 | 17.65 | 19.05 |
| 2.7.2 | 2020 | NM [118] | 73.31 | 37.43 | -35.88 | 3663.00 | - | - | 21.80 | 15.57 | 28.57 |

TABLE 19: ResNet-34 on ImageNet-1K.

19| Section | Year | Method | Baseline Top-1 Acc.(%) | Pruned Top-1 Acc.(%) | Top-1 Acc. ↓(%) | Baseline FLOPs (M) | Pruned FLOPs (M) | FLOPs ↓(%) | Baseline Params (M) | Pruned Params (M) | Params ↓(%) |
|---|---|---|---|---|---|---|---|---|---|---|---|
| ResNet-50 | | | | | | | | | | | |
| 2.6.2 | 2020 | DMCP [121] | 76.60 | 66.40 | -10.00 | 4089.00 | 277.25 | 93.22 | 25.56 | - | - |
| 2.7.3 | 2021 | JMDP [126] | 76.60 | 73.50 | -3.10 | 4089.00 | 897.59 | 78.05 | 25.56 | - | - |
| 2.5.1 | 2022 | SMCP [127] | 77.20 | 74.40 | -2.80 | 4089.00 | 897.59 | 78.05 | 25.56 | - | - |
| 2.1.2 | 2022 | CLR-RNF [49] | 76.01 | 71.11 | -4.90 | 4089.00 | 925.25 | 77.37 | 25.56 | 6.90 | 73.00 |
| 2.2.1 | 2021 | CHIP [42] | 76.15 | 73.30 | -2.85 | 4089.00 | 952.74 | 76.70 | 25.56 | 8.03 | 68.60 |
| 2.2.1 | 2020 | HRank [45] | 76.15 | 69.10 | -7.05 | 4089.00 | 979.76 | 76.04 | 25.56 | 8.29 | 67.57 |
| 2.5.1 | 2022 | SMCP [127] | 76.20 | 74.60 | -1.60 | 4089.00 | 997.32 | 75.61 | 25.56 | - | - |
| 2.5.1 | 2022 | CHEX [123] | - | 76.00 | - | 4089.00 | 997.32 | 75.61 | 25.56 | - | - |
| 2.3.1 | 2020 | EagleEye [101] | - | 74.20 | - | 4089.00 | 997.32 | 75.61 | 25.56 | - | - |
| 2.6.3 | 2019 | MetaPruning [128] | 76.60 | 73.40 | -3.20 | 4089.00 | 997.32 | 75.61 | 25.56 | - | - |
| 2.6.2 | 2022 | PaS [129] | 74.68 | 74.80 | 0.12 | 4089.00 | 997.32 | 75.61 | 25.56 | - | - |
| 2.4.1 | 2021 | GFP [130] | 76.79 | 73.94 | -2.85 | 4089.00 | 1020.00 | 75.06 | 25.56 | - | - |
| 2.6.2 | 2022 | DNCP [109] | 76.60 | 74.30 | -2.30 | 4089.00 | 1097.05 | 73.17 | 25.56 | - | - |
| 2.6.2 | 2020 | DMCP [121] | 76.60 | 74.40 | -2.20 | 4089.00 | 1097.05 | 73.17 | 25.56 | - | - |
| 2.3.2 | 2019 | GAL [63] | 76.15 | 69.31 | -6.84 | 4089.00 | 1109.73 | 72.86 | 25.56 | 10.22 | 60.00 |
| 2.2.2 | 2017 | ThiNet [115] | 72.88 | 68.42 | -4.46 | 4089.00 | 1165.26 | 71.50 | 25.56 | 8.68 | 66.04 |
| 2.6.1 | 2022 | DECORE [35] | 76.15 | 69.71 | -6.44 | 4089.00 | 1189.71 | 70.90 | 25.56 | 6.13 | 76.00 |
| 2.1.2 | 2022 | CLR-RNF [49] | 76.01 | 73.34 | -2.67 | 4089.00 | 1223.72 | 70.07 | 25.56 | 9.00 | 64.79 |
| 2.1.2 | 2022 | CLR-RNF [49] | 76.01 | 72.67 | -3.34 | 4089.00 | 1223.72 | 70.07 | 25.56 | 9.00 | 64.79 |
| 2.7.1 | 2020 | EB [64] | 75.99 | 70.16 | -5.83 | 4089.00 | 1282.37 | 68.64 | 25.56 | 7.67 | 70.00 |
| 2.5.2 | 2021 | DDG [83] | 76.13 | 75.12 | -1.01 | 4089.00 | 1312.57 | 67.90 | 25.56 | - | - |
| 2.3.3 | 2021 | GREG [97] | 76.13 | 73.90 | -2.23 | 4089.00 | 1336.27 | 67.32 | 25.56 | - | - |
| 2.4.1 | 2019 | Mol-19 [114] | 76.18 | 71.69 | -4.49 | 4089.00 | 1339.67 | 67.24 | 25.56 | 7.89 | 69.14 |
| 2.4.1 | 2022 | HAP [107] | 75.62 | 71.18 | -4.44 | 4089.00 | 1343.24 | 67.15 | 25.56 | 5.23 | 79.53 |
| 2.3.3 | 2021 | OTO [38] | 76.10 | 75.10 | -1.00 | 4089.00 | 1410.71 | 65.50 | 25.56 | 9.07 | 64.50 |
| 2.3.3 | 2021 | OTO [38] | 76.10 | 74.70 | -1.40 | 4089.00 | 1410.71 | 65.50 | 25.56 | 9.07 | 64.50 |
| 2.6.3 | 2022 | CCEP [59] | 76.13 | 74.87 | -1.26 | 4089.00 | 1468.36 | 64.09 | 25.56 | - | - |
| 2.3.2 | 2022 | WhiteBox [46] | 76.15 | 74.21 | -1.94 | 4089.00 | 1492.49 | 63.50 | 25.56 | - | - |
| 2.2.1 | 2021 | CHIP [42] | 76.15 | 75.26 | -0.89 | 4089.00 | 1521.11 | 62.80 | 25.56 | 11.07 | 56.70 |
| 2.7.2 | 2021 | CC [60] | 76.15 | 74.54 | -1.61 | 4089.00 | 1525.90 | 62.68 | 25.56 | 10.58 | 58.61 |
| 2.2.1 | 2020 | HRank [45] | 76.15 | 71.98 | -4.17 | 4089.00 | 1549.62 | 62.10 | 25.56 | 13.80 | 46.00 |
| 2.3.2 | 2021 | ResRep [93] | 76.15 | 75.30 | -0.85 | 4089.00 | 1549.73 | 62.10 | 25.56 | - | - |
| 2.3.2 | 2019 | GAL [63] | 76.15 | 69.88 | -6.27 | 4089.00 | 1579.64 | 61.37 | 25.56 | 14.70 | 42.47 |
| 2.3.3 | 2021 | GREG [97] | 76.13 | 74.93 | -1.20 | 4089.00 | 1597.27 | 60.94 | 25.56 | - | - |
| 2.6.1 | 2022 | DECORE [35] | 76.15 | 72.06 | -4.09 | 4089.00 | 1599.61 | 60.88 | 25.56 | 8.89 | 65.22 |
| 2.6.2 | 2020 | LFPC [92] | 76.15 | 74.46 | -1.69 | 4089.00 | 1602.89 | 60.80 | 25.56 | - | - |
| 2.6.2 | 2020 | LFPC [92] | 76.15 | 74.18 | -1.97 | 4089.00 | 1602.89 | 60.80 | 25.56 | - | - |
| 2.5.2 | 2021 | DDG [83] | 76.13 | 76.41 | 0.28 | 4089.00 | 1647.87 | 59.70 | 25.56 | - | - |
| 2.4.1 | 2022 | HAP [107] | 75.62 | 74.00 | -1.62 | 4089.00 | 1653.59 | 59.56 | 25.56 | 8.88 | 65.26 |
| 2.5.1 | 2018 | GDP-Lin [113] | 75.13 | 70.93 | -4.20 | 4089.00 | 1663.14 | 59.33 | 25.56 | - | - |
| 2.4.1 | 2022 | SOSP [75] | 76.15 | 73.38 | -2.77 | 4089.00 | 1676.49 | 59.00 | 25.56 | 9.97 | 61.00 |
| 2.7.1 | 2022 | PaT [125] | - | 74.85 | - | 4089.00 | 1690.45 | 58.66 | 25.56 | - | - |
| 2.2.2 | 2019 | AOFP [48] | 75.34 | 75.11 | -0.23 | 4089.00 | 1763.05 | 56.88 | 25.56 | - | - |
| 2.3.3 | 2021 | GREG [97] | 75.40 | 75.22 | -0.18 | 4089.00 | 1770.13 | 56.71 | 25.56 | - | - |
| 2.6.3 | 2020 | ABCPruner [50] | 76.01 | 73.52 | -2.49 | 4089.00 | 1774.19 | 56.61 | 25.56 | 11.24 | 56.03 |
| 2.6.3 | 2022 | CCEP [59] | 76.13 | 75.55 | -0.58 | 4089.00 | 1784.85 | 56.35 | 25.56 | - | - |
| 2.3.2 | 2021 | ResRep [93] | 76.15 | 75.97 | -0.18 | 4089.00 | 1794.66 | 56.11 | 25.56 | - | - |
| 2.6.2 | 2021 | EE [61] | 76.15 | 75.66 | -0.49 | 4089.00 | 1799.16 | 56.00 | 25.56 | - | - |
| 2.2.2 | 2017 | ThiNet [115] | 72.88 | 71.01 | -1.87 | 4089.00 | 1806.15 | 55.83 | 25.56 | 12.41 | 51.45 |
| 2.4.3 | 2019 | C-SGD [96] | 75.33 | 74.54 | -0.79 | 4089.00 | 1808.97 | 55.76 | 25.56 | - | - |
| 2.5.1 | 2021 | DCP-CAC [76] | - | - | -0.79 | 4089.00 | 1817.33 | 55.56 | 25.56 | 12.35 | 51.69 |
| 2.2.3 | 2018 | DCP [74] | - | - | 1.06 | 4089.00 | 1817.33 | 55.56 | 25.56 | 12.41 | 51.46 |
| 2.6.2 | 2022 | DAIS [84] | 72.75 | 74.45 | 1.70 | 4089.00 | 1827.78 | 55.30 | 25.56 | - | - |
| 2.1.2 | 2021 | SRR [87] | 76.13 | 75.11 | -1.02 | 4089.00 | 1835.96 | 55.10 | 25.56 | - | - |
| 2.3.1 | 2019 | GBN [94] | 74.51 | 75.18 | 0.67 | 4089.00 | 1837.60 | 55.06 | 25.56 | 11.91 | 53.40 |
| 2.3.2 | 2019 | GAL [63] | 76.15 | 71.80 | -4.35 | 4089.00 | 1839.55 | 55.01 | 25.56 | 19.36 | 24.27 |
| 2.3.2 | 2020 | DMC [102] | 76.15 | 75.35 | -0.80 | 4089.00 | 1840.05 | 55.00 | 25.56 | - | - |
| 2.4.3 | 2022 | EKG [95] | 77.00 | 76.60 | -0.40 | 4089.00 | 1840.05 | 55.00 | 25.56 | - | - |
| 2.6.1 | 2022 | RL-MCTS [41] | 77.34 | 76.46 | -0.88 | 4089.00 | 1840.05 | 55.00 | 25.56 | - | - |
| 2.6.2 | 2022 | DDNP [100] | 76.13 | 75.89 | -0.24 | 4089.00 | 1840.05 | 55.00 | 25.56 | - | - |
| 2.3.2 | 2020 | SCOP [80] | 76.15 | 75.26 | -0.89 | 4089.00 | 1856.41 | 54.60 | 25.56 | 12.32 | 51.80 |
| 2.4.2 | 2019 | RBP [55] | - | 73.00 | - | 4089.00 | 1858.64 | 54.55 | 25.56 | - | - |
| 2.3.2 | 2021 | ResRep [93] | 76.15 | 76.15 | 0.00 | 4089.00 | 1858.86 | 54.54 | 25.56 | - | - |
| 2.3.1 | 2020 | SCP [58] | 75.89 | 74.20 | 1.69 | 4089.00 | 1868.67 | 54.30 | 25.56 | - | - |
| 2.3.1 | 2020 | SCP [58] | 75.89 | 75.27 | 0.62 | 4089.00 | 1868.67 | 54.30 | 25.56 | - | - |
| 2.6.3 | 2020 | ABCPruner [50] | 76.01 | 73.86 | -2.15 | 4089.00 | 1869.25 | 54.29 | 25.56 | 11.75 | 54.03 |
| 2.4.1 | 2019 | CCP [99] | 76.15 | 76.98 | 0.83 | 4089.00 | 1876.85 | 54.10 | 25.56 | - | - |
| 2.3.1 | 2020 | PR [62] | 76.15 | 75.63 | 0.52 | 4089.00 | 1880.94 | 54.00 | 25.56 | - | - |
| 2.2.3 | 2022 | DLRFC [44] | 76.13 | 75.84 | 0.29 | 4089.00 | 1880.94 | 54.00 | 25.56 | 15.34 | 40.00 |
| 2.7.3 | 2021 | JMDP [126] | 76.60 | 76.00 | -0.60 | 4089.00 | 1894.90 | 53.66 | 25.56 | - | - |
| 2.5.1 | 2022 | SMCP [127] | 77.20 | 76.60 | -0.60 | 4089.00 | 1894.90 | 53.66 | 25.56 | - | - |

Continue on the next page...

TABLE 20: ResNet-50 on ImageNet-1K. (Part 1)

| Section | Year | Method | Baseline Top-1 Acc.(%) | Pruned Top-1 Acc.(%) | Top-1 Acc. ↓(%) | Baseline FLOPs (M) | Pruned FLOPs (M) | FLOPs ↓(%) | Baseline Params (M) | Pruned Params (M) | Params ↓(%) |
|---|---|---|---|---|---|---|---|---|---|---|---|
| ResNet-50 (cont.) | | | | | | | | | | | |
| 2.6.2 | 2022 | MFP [91] | 76.15 | 74.13 | -2.02 | 4089.00 | 1901.38 | 53.50 | 25.56 | - | - |
| 2.6.2 | 2022 | MFP [91] | 76.15 | 74.86 | -1.29 | 4089.00 | 1901.38 | 53.50 | 25.56 | - | - |
| 2.1.2 | 2019 | FPGM [82] | 76.15 | 75.59 | -0.56 | 4089.00 | 1901.38 | 53.50 | 25.56 | - | - |
| 2.7.2 | 2020 | Hinge [66] | - | 74.70 | - | 4089.00 | 1903.43 | 53.45 | 25.56 | - | - |
| 2.6.1 | 2022 | GNN-RL [85] | 76.10 | 74.28 | -1.82 | 4089.00 | 1921.83 | 53.00 | 25.56 | - | - |
| 2.7.2 | 2021 | CC [60] | 76.15 | 75.59 | -0.56 | 4089.00 | 1924.82 | 52.93 | 25.56 | 13.20 | 48.36 |
| 2.2.1 | 2021 | LRMF [68] | 76.15 | 74.70 | -1.45 | 4089.00 | 1934.10 | 52.70 | 25.56 | - | - |
| 2.5.1 | 2018 | GDP-Lin [113] | 75.13 | 71.89 | -3.24 | 4089.00 | 1991.53 | 51.30 | 25.56 | - | - |
| 2.5.1 | 2022 | SMCP [127] | 76.20 | 76.80 | 0.60 | 4089.00 | 1994.63 | 51.22 | 25.56 | - | - |
| 2.5.1 | 2022 | CHEX [123] | - | 77.40 | - | 4089.00 | 1994.63 | 51.22 | 25.56 | - | - |
| 2.3.1 | 2020 | EagleEye [101] | - | 76.40 | - | 4089.00 | 1994.63 | 51.22 | 25.56 | - | - |
| 2.6.3 | 2019 | MetaPruning [128] | 76.60 | 75.40 | -1.20 | 4089.00 | 1994.63 | 51.22 | 25.56 | - | - |
| 2.6.2 | 2022 | PaS [129] | 76.65 | 76.70 | 0.05 | 4089.00 | 1994.63 | 51.22 | 25.56 | - | - |
| 2.7.1 | 2022 | RRCP [105] | - | 75.13 | - | 4089.00 | 2003.20 | 51.01 | 25.56 | 13.83 | 45.88 |
| 2.4.1 | 2022 | SOSP [75] | 76.15 | 74.39 | -1.76 | 4089.00 | 2003.61 | 51.00 | 25.56 | 11.76 | 54.00 |
| 2.4.1 | 2021 | GFP [130] | 76.79 | 76.42 | -0.37 | 4089.00 | 2040.00 | 50.11 | 25.56 | - | - |
| 2.6.2 | 2020 | DSA [79] | 76.02 | 74.69 | -1.33 | 4089.00 | 2044.50 | 50.00 | 25.56 | - | - |
| 2.3.3 | 2019 | OICSR [108] | 76.31 | 75.95 | -0.36 | 4089.00 | 2044.50 | 50.00 | 25.56 | - | - |
| 2.4.1 | 2019 | CCP [99] | 76.15 | 76.80 | 0.65 | 4089.00 | 2093.57 | 48.80 | 25.56 | - | - |
| 2.2.1 | 2021 | CHIP [42] | 76.15 | 76.15 | 0.00 | 4089.00 | 2097.66 | 48.70 | 25.56 | 14.26 | 44.20 |
| 2.6.2 | 2022 | DNCP [109] | 76.60 | 76.30 | -0.30 | 4089.00 | 2194.10 | 46.34 | 25.56 | - | - |
| 2.6.2 | 2020 | DMCP [121] | 76.60 | 76.20 | -0.40 | 4089.00 | 2194.10 | 46.34 | 25.56 | - | - |
| 2.4.3 | 2019 | C-SGD [96] | 75.33 | 74.93 | -0.40 | 4089.00 | 2198.25 | 46.24 | 25.56 | - | - |
| 2.6.2 | 2021 | EE [61] | 76.15 | 76.05 | -0.10 | 4089.00 | 2203.97 | 46.10 | 25.56 | - | - |
| 2.6.1 | 2022 | RL-MCTS [41] | 77.34 | 76.80 | -0.54 | 4089.00 | 2203.97 | 46.10 | 25.56 | - | - |
| 2.3.2 | 2022 | WhiteBox [46] | 76.15 | 75.32 | -0.83 | 4089.00 | 2224.42 | 45.60 | 25.56 | - | - |
| 2.3.2 | 2020 | SCOP [80] | 76.15 | 75.95 | -0.20 | 4089.00 | 2236.68 | 45.30 | 25.56 | 14.62 | 42.80 |
| 2.4.1 | 2022 | SOSP [75] | 76.15 | 75.21 | -0.94 | 4089.00 | 2248.95 | 45.00 | 25.56 | 13.04 | 49.00 |
| 2.4.1 | 2019 | Mol-19 [114] | 76.18 | 74.50 | -1.68 | 4089.00 | 2249.45 | 44.99 | 25.56 | 14.18 | 44.53 |
| 2.2.1 | 2021 | CHIP [42] | 76.15 | 76.30 | 0.15 | 4089.00 | 2257.13 | 44.80 | 25.56 | 15.13 | 40.80 |
| 2.6.3 | 2022 | CCEP [59] | 76.13 | 76.06 | -0.07 | 4089.00 | 2266.94 | 44.56 | 25.56 | - | - |
| 2.4.1 | 2022 | HAP [107] | 75.62 | 75.36 | -0.26 | 4089.00 | 2268.99 | 44.51 | 25.56 | 13.74 | 46.26 |
| 2.3.3 | 2019 | OICSR [108] | 76.31 | 76.30 | -0.01 | 4089.00 | 2273.48 | 44.40 | 25.56 | - | - |
| 2.1.2 | 2021 | SRR [87] | 76.13 | 75.76 | -0.37 | 4089.00 | 2285.75 | 44.10 | 25.56 | - | - |
| 2.2.3 | 2018 | NISP [106] | - | - | 0.89 | 4089.00 | 2289.43 | 44.01 | 25.56 | 14.36 | 43.82 |
| 2.2.1 | 2020 | HRank [45] | 76.15 | 74.98 | -1.17 | 4089.00 | 2299.44 | 43.77 | 25.56 | 16.19 | 36.67 |
| 2.6.2 | 2019 | TAS [88] | 74.94 | 76.20 | 1.26 | 4089.00 | 2310.28 | 43.50 | 25.56 | - | - |
| 2.3.2 | 2019 | GAL [63] | 76.15 | 71.95 | -4.20 | 4089.00 | 2329.43 | 43.03 | 25.56 | 21.25 | 16.86 |
| 2.5.1 | 2021 | SEP [73] | 76.12 | 75.22 | -0.90 | 4089.00 | 2330.73 | 43.00 | 25.56 | - | - |
| 2.3.2 | 2021 | ABP [39] | 75.88 | 74.80 | -1.08 | 4089.00 | 2338.91 | 42.80 | 25.56 | 13.04 | 49.00 |
| 2.6.1 | 2022 | DECORE [35] | 76.15 | 74.58 | -1.57 | 4089.00 | 2359.42 | 42.30 | 25.56 | 14.13 | 44.71 |
| 2.6.2 | 2022 | MFP [91] | 76.15 | 75.67 | -0.48 | 4089.00 | 2363.44 | 42.20 | 25.56 | - | - |
| 2.5.1 | 2018 | GDP-Lin [113] | 75.13 | 72.61 | -2.52 | 4089.00 | 2372.89 | 41.97 | 25.56 | - | - |
| 2.5.1 | 2018 | SFP [90] | 76.15 | 62.14 | -14.01 | 4089.00 | 2379.80 | 41.80 | 25.56 | - | - |
| 2.5.1 | 2018 | SFP [90] | 76.15 | 74.61 | -1.54 | 4089.00 | 2379.80 | 41.80 | 25.56 | - | - |
| 2.3.1 | 2019 | GBN [94] | 76.50 | 76.19 | -0.31 | 4089.00 | 2431.32 | 40.54 | 25.56 | 17.42 | 31.83 |
| 2.1.2 | 2022 | CLR-RNF [49] | 76.01 | 74.85 | -1.16 | 4089.00 | 2437.48 | 40.39 | 25.56 | 16.92 | 33.80 |
| 2.6.2 | 2020 | DSA [79] | 76.02 | 75.10 | -0.92 | 4089.00 | 2453.40 | 40.00 | 25.56 | - | - |
| 2.7.1 | 2020 | EB [64] | 75.99 | 73.35 | -2.64 | 4089.00 | 2474.71 | 39.48 | 25.56 | 12.78 | 50.00 |
| 2.3.3 | 2019 | OICSR [108] | 76.31 | 76.53 | 0.22 | 4089.00 | 2563.80 | 37.30 | 25.56 | - | - |
| 2.2.2 | 2017 | ThiNet [115] | 72.88 | 72.04 | -0.84 | 4089.00 | 2584.76 | 36.79 | 25.56 | 16.98 | 33.57 |
| 2.4.3 | 2019 | C-SGD [96] | 75.33 | 75.27 | -0.06 | 4089.00 | 2586.29 | 36.75 | 25.56 | - | - |
| 2.4.1 | 2019 | Mol-19 [114] | 76.18 | 75.48 | -0.70 | 4089.00 | 2659.35 | 34.96 | 25.56 | 17.87 | 30.08 |
| 2.7.3 | 2022 | GKP-TMI [89] | 76.15 | 75.53 | -0.62 | 4089.00 | 2709.37 | 33.74 | 25.56 | 17.07 | 33.21 |
| 2.2.2 | 2019 | AOFP [48] | 75.34 | 75.63 | 0.29 | 4089.00 | 2740.16 | 32.99 | 25.56 | - | - |
| 2.3.3 | 2021 | GREG [97] | 76.13 | 76.13 | 0.00 | 4089.00 | 2744.30 | 32.89 | 25.56 | - | - |
| 2.7.3 | 2021 | JMDP [126] | 76.60 | 76.50 | -0.10 | 4089.00 | 2792.49 | 31.71 | 25.56 | - | - |
| 2.6.2 | 2020 | DMCP [121] | 76.60 | 77.00 | 0.40 | 4089.00 | 2792.49 | 31.71 | 25.56 | - | - |
| 2.2.3 | 2020 | PFP [37] | 76.13 | 75.21 | -0.92 | 4089.00 | 2860.26 | 30.05 | 25.56 | 14.30 | 44.04 |
| 2.4.1 | 2022 | SOSP [75] | 76.15 | 76.60 | 0.45 | 4089.00 | 2944.08 | 28.00 | 25.56 | 17.89 | 30.00 |
| 2.2.3 | 2018 | NISP [106] | - | - | 0.21 | 4089.00 | 2972.29 | 27.31 | 25.56 | 18.63 | 27.12 |
| 2.4.1 | 2022 | SOSP [75] | 76.15 | 75.85 | -0.30 | 4089.00 | 2984.97 | 27.00 | 25.56 | 15.34 | 40.00 |
| 2.6.3 | 2019 | MetaPruning [128] | 76.60 | 76.20 | -0.40 | 4089.00 | 2991.95 | 26.83 | 25.56 | - | - |
| 2.6.2 | 2022 | PaS [129] | 77.53 | 77.60 | 0.07 | 4089.00 | 2991.95 | 26.83 | 25.56 | - | - |
| 2.5.1 | 2022 | SMCP [127] | 76.20 | 77.10 | 0.90 | 4089.00 | 2991.95 | 26.83 | 25.56 | - | - |
| 2.3.1 | 2020 | EagleEye [101] | - | 77.10 | - | 4089.00 | 2991.95 | 26.83 | 25.56 | - | - |
| 2.4.1 | 2021 | GFP [130] | 76.79 | 76.95 | 0.16 | 4089.00 | 3060.00 | 25.17 | 25.56 | - | - |
| 2.5.1 | 2022 | SMCP [127] | 77.20 | 77.60 | 0.40 | 4089.00 | 3091.68 | 24.39 | 25.56 | - | - |
| 2.7.3 | 2022 | GKP-TMI [89] | 76.15 | 75.96 | -0.19 | 4089.00 | 3168.98 | 22.50 | 25.56 | 19.91 | 22.10 |
| 2.7.1 | 2020 | EB [64] | 75.99 | 73.86 | -2.13 | 4089.00 | 3228.48 | 21.04 | 25.56 | 17.89 | 30.00 |
| 2.4.1 | 2022 | SOSP [75] | 76.15 | 76.56 | 0.41 | 4089.00 | 3230.31 | 21.00 | 25.56 | 19.94 | 22.00 |
| 2.4.1 | 2019 | Mol-19 [114] | 76.18 | 76.43 | 0.25 | 4089.00 | 3269.20 | 20.05 | 25.56 | 22.56 | 11.72 |
| 2.6.1 | 2022 | DECORE [35] | 76.15 | 76.31 | 0.16 | 4089.00 | 3539.13 | 13.45 | 25.56 | 22.74 | 11.02 |
| 2.2.3 | 2020 | PFP [37] | 76.13 | 75.91 | -0.22 | 4089.00 | 3646.57 | 10.82 | 25.56 | 20.96 | 18.01 |
| 2.6.3 | 2022 | EDropout [111] | 75.27 | 73.72 | -1.55 | 4089.00 | - | - | 25.56 | 10.86 | 57.50 |
| 2.7.3 | 2022 | 1xN [131] | 77.01 | 76.65 | -0.35 | 4089.00 | - | - | 25.56 | 12.78 | 50.00 |

TABLE 20: ResNet-50 on ImageNet-1K. (Part 2)

| Section | Year | Method | Baseline Top-1 Acc.(%) | Pruned Top-1 Acc.(%) | Top-1 Acc. ↓(%) | Baseline FLOPs (M) | Pruned FLOPs (M) | FLOPs ↓(%) | Baseline Params (M) | Pruned Params (M) | Params ↓(%) |
|---|---|---|---|---|---|---|---|---|---|---|---|
| ResNet-101 | | | | | | | | | | | |
| 2.4.1 | 2019 | Mol-19 [114] | 77.37 | 74.16 | -3.21 | 7801.00 | 1760.23 | 77.44 | 44.55 | 13.55 | 69.57 |
| 2.5.1 | 2022 | CHEX [123] | - | 77.60 | - | 7801.00 | 1957.98 | 74.90 | 44.55 | - | - |
| 2.6.3 | 2020 | ABCPruner [50] | 77.38 | 74.76 | -2.62 | 7801.00 | 1958.69 | 74.89 | 44.55 | 12.94 | 70.95 |
| 2.5.1 | 2022 | SMCP [127] | 77.40 | 76.80 | -0.60 | 7801.00 | 2000.26 | 74.36 | 44.55 | - | - |
| 2.4.1 | 2019 | Mol-19 [114] | 77.37 | 75.38 | -1.99 | 7801.00 | 2470.32 | 68.33 | 44.55 | 17.74 | 60.18 |
| 2.5.1 | 2022 | SMCP [127] | 77.40 | 77.30 | -0.10 | 7801.00 | 2600.33 | 66.67 | 44.55 | - | - |
| 2.4.1 | 2019 | Mol-19 [114] | 77.37 | 75.27 | -2.10 | 7801.00 | 2850.37 | 63.46 | 44.55 | 20.63 | 53.69 |
| 2.3.2 | 2020 | SCOP [80] | 77.37 | 77.36 | -0.01 | 7801.00 | 3104.80 | 60.20 | 44.55 | 18.80 | 57.80 |
| 2.6.3 | 2020 | ABCPruner [50] | 77.38 | 75.82 | -1.56 | 7801.00 | 3137.80 | 59.78 | 44.55 | 17.72 | 60.22 |
| 2.3.2 | 2020 | DMC [102] | 77.37 | 77.40 | 0.03 | 7801.00 | 3432.44 | 56.00 | 44.55 | - | - |
| 2.5.1 | 2022 | CHEX [123] | - | 78.80 | - | 7801.00 | 3503.75 | 55.09 | 44.55 | - | - |
| 2.5.1 | 2022 | SMCP [127] | 77.40 | 77.80 | 0.40 | 7801.00 | 3600.46 | 53.85 | 44.55 | - | - |
| 2.3.1 | 2018 | RSNLI [132] | 76.40 | 74.56 | -1.84 | 7801.00 | 3690.47 | 52.69 | 44.55 | 17.37 | 61.00 |
| 2.2.2 | 2019 | AOFP [48] | 76.63 | 76.40 | -0.23 | 7801.00 | 3885.04 | 50.20 | 44.55 | - | - |
| 2.4.1 | 2021 | GFP [130] | 78.29 | 78.33 | 0.04 | 7801.00 | 3900.00 | 50.01 | 44.55 | - | - |
| 2.3.1 | 2018 | RSNLI [132] | 76.40 | 75.27 | -1.13 | 7801.00 | 3962.55 | 49.20 | 44.55 | 23.61 | 47.00 |
| 2.5.1 | 2022 | SMCP [127] | 77.40 | 78.10 | 0.70 | 7801.00 | 4000.51 | 48.72 | 44.55 | - | - |
| 2.3.2 | 2020 | SCOP [80] | 77.37 | 77.75 | 0.38 | 7801.00 | 4009.71 | 48.60 | 44.55 | 23.70 | 46.80 |
| 2.1.2 | 2019 | FPGM [82] | 77.37 | 77.32 | -0.05 | 7801.00 | 4134.53 | 47.00 | 44.55 | - | - |
| 2.2.3 | 2020 | PFP [37] | 77.37 | 76.43 | -0.94 | 7801.00 | 4284.31 | 45.08 | 44.55 | 22.07 | 50.45 |
| 2.5.1 | 2018 | SFP [90] | 77.37 | 77.51 | 0.14 | 7801.00 | 4508.98 | 42.20 | 44.55 | - | - |
| 2.5.1 | 2018 | SFP [90] | 77.37 | 77.03 | -0.34 | 7801.00 | 4508.98 | 42.20 | 44.55 | - | - |
| 2.4.1 | 2019 | Mol-19 [114] | 77.37 | 77.35 | -0.02 | 7801.00 | 4700.60 | 39.74 | 44.55 | 31.10 | 30.20 |
| 2.2.2 | 2019 | AOFP [48] | 76.63 | 76.88 | 0.25 | 7801.00 | 5451.43 | 30.12 | 44.55 | - | - |
| 2.2.3 | 2020 | PFP [37] | 77.37 | 76.78 | -0.59 | 7801.00 | 5509.07 | 29.38 | 44.55 | 29.83 | 33.04 |
| 2.6.3 | 2022 | EDropout [111] | 75.94 | 73.94 | -2.00 | 7801.00 | - | - | 44.55 | 19.13 | 57.07 |
| ResNet-152 | | | | | | | | | | | |
| 2.6.3 | 2020 | ABCPruner [50] | 78.31 | 76.00 | -2.31 | 11514.00 | 2697.93 | 76.57 | 60.19 | 15.62 | 74.05 |
| 2.2.2 | 2019 | AOFP [48] | 77.37 | 76.40 | -0.97 | 11514.00 | 2858.09 | 75.18 | 60.19 | - | - |
| 2.2.2 | 2019 | AOFP [48] | 77.37 | 77.00 | -0.37 | 11514.00 | 4215.68 | 63.39 | 60.19 | - | - |
| 2.6.3 | 2020 | ABCPruner [50] | 78.31 | 77.12 | -1.19 | 11514.00 | 4275.39 | 62.87 | 60.19 | 24.07 | 60.01 |
| 2.2.2 | 2019 | AOFP [48] | 77.37 | 77.47 | 0.10 | 11514.00 | 6246.96 | 45.74 | 60.19 | - | - |

TABLE 21: ResNet-101/152 on ImageNet-1K.

| Section | Year | Method | Baseline Top-1 Acc.(%) | Pruned Top-1 Acc.(%) | Top-1 Acc. ↓(%) | Baseline FLOPs (M) | Pruned FLOPs (M) | FLOPs ↓(%) | Baseline Params (M) | Pruned Params (M) | Params ↓(%) |
|---|---|---|---|---|---|---|---|---|---|---|---|
| MobileNet-V1 | | | | | | | | | | | |
| 2.6.3 | 2019 | MetaPruning [128] | 70.60 | 57.20 | -13.40 | 569.00 | 41.00 | 92.79 | 4.20 | - | - |
| 2.6.3 | 2019 | MetaPruning [128] | 70.60 | 66.10 | -4.50 | 569.00 | 149.00 | 73.81 | 4.20 | - | - |
| 2.5.1 | 2022 | SMCP [127] | 72.60 | 68.30 | -4.30 | 569.00 | 208.00 | 63.44 | 4.20 | - | - |
| 2.6.1 | 2018 | AMC [103] | 70.60 | 68.90 | -1.70 | 569.00 | 227.60 | 60.00 | 4.20 | - | - |
| 2.6.1 | 2021 | AGMC [86] | 70.60 | 69.40 | -1.20 | 569.00 | 227.60 | 60.00 | 4.20 | - | - |
| 2.6.1 | 2022 | GNN-RL [85] | 70.90 | 69.50 | -1.40 | 569.00 | 227.60 | 60.00 | 4.20 | - | - |
| 2.3.1 | 2020 | EagleEye [101] | - | 70.90 | - | 569.00 | 284.00 | 50.09 | 4.20 | - | - |
| 2.4.3 | 2022 | RollBack [104] | - | 49.34 | - | 569.00 | 284.50 | 50.00 | 4.20 | - | - |
| 2.5.2 | 2020 | DRLP [120] | 70.90 | 70.60 | -0.30 | 569.00 | 284.50 | 50.00 | 4.20 | - | - |
| 2.6.1 | 2018 | AMC [103] | 70.60 | 70.20 | -0.40 | 569.00 | 284.50 | 50.00 | 4.20 | - | - |
| 2.6.3 | 2019 | MetaPruning [128] | 70.60 | 70.90 | 0.30 | 569.00 | 324.00 | 43.06 | 4.20 | - | - |
| 2.5.2 | 2022 | FTWT [53] | 69.57 | 69.66 | 0.09 | 569.00 | 335.31 | 41.07 | 4.20 | - | - |
| 2.5.1 | 2022 | SMCP [127] | 72.60 | 71.00 | -1.60 | 569.00 | 356.00 | 37.43 | 4.20 | - | - |
| 2.7.3 | 2022 | 1xN [131] | 71.15 | 70.28 | -0.87 | 569.00 | - | - | 4.20 | 2.10 | 50 |
| MobileNet-V2 | | | | | | | | | | | |
| 2.6.3 | 2019 | MetaPruning [128] | 74.70 | 58.30 | -16.40 | 300.00 | 22.05 | 92.65 | 3.50 | - | - |
| 2.6.2 | 2020 | DMCP [121] | 72.30 | 59.10 | -13.30 | 300.00 | 43.00 | 85.67 | 3.50 | - | - |
| 2.6.3 | 2019 | MetaPruning [128] | 74.70 | 63.80 | -10.90 | 300.00 | 43.08 | 85.64 | 3.50 | - | - |
| 2.6.3 | 2019 | MetaPruning [128] | 74.70 | 65.00 | -9.70 | 300.00 | 53.85 | 82.05 | 3.50 | - | - |
| 2.6.2 | 2020 | DMCP [121] | 72.30 | 62.70 | -9.70 | 300.00 | 59.00 | 80.33 | 3.50 | - | - |
| 2.6.3 | 2019 | MetaPruning [128] | 74.70 | 67.30 | -7.40 | 300.00 | 63.59 | 78.80 | 3.50 | - | - |
| 2.6.3 | 2019 | MetaPruning [128] | 74.70 | 68.20 | -6.50 | 300.00 | 71.79 | 76.07 | 3.50 | - | - |
| 2.6.2 | 2020 | DMCP [121] | 72.30 | 66.10 | -6.30 | 300.00 | 87.00 | 71.00 | 3.50 | - | - |
| 2.4.1 | 2021 | GFP [130] | 75.94 | 65.94 | -9.80 | 300.00 | 90.00 | 70.00 | 3.50 | - | - |
| 2.6.2 | 2022 | DNCP [109] | 72.30 | 66.50 | -5.80 | 300.00 | 97.00 | 67.67 | 3.50 | - | - |
| 2.6.2 | 2020 | DMCP [121] | 72.30 | 67.00 | -5.30 | 300.00 | 97.00 | 67.67 | 3.50 | - | - |
| 2.2.2 | 2020 | GFS [124] | 72.00 | 66.90 | -5.10 | 300.00 | 102.23 | 65.92 | 3.50 | 1.90 | 45.71 |
| 2.2.2 | 2020 | GFS [124] | 72.00 | 68.80 | -3.20 | 300.00 | 131.85 | 56.05 | 3.50 | 2.00 | 42.86 |
| 2.2.2 | 2020 | GFS [124] | 72.00 | 69.70 | -2.30 | 300.00 | 145.22 | 51.59 | 3.50 | 2.20 | 37.14 |
| 2.5.2 | 2021 | ManiDP [81] | 71.80 | 69.62 | -2.18 | 300.00 | 146.40 | 51.20 | 3.50 | - | - |
| 2.6.3 | 2019 | MetaPruning [128] | 74.70 | 72.70 | -2.00 | 300.00 | 149.23 | 50.26 | 3.50 | - | - |
| 2.4.1 | 2021 | GFP [130] | 75.74 | 69.16 | -6.58 | 300.00 | 150.00 | 50.00 | 3.50 | - | - |
| 2.3.2 | 2020 | DMC [102] | 71.88 | 68.37 | -3.51 | 300.00 | 162.00 | 46.00 | 3.50 | - | - |
| 2.2.2 | 2020 | GFS [124] | 72.00 | 70.40 | -1.60 | 300.00 | 162.42 | 45.86 | 3.50 | 2.30 | 34.29 |
| 2.5.2 | 2021 | DDG [83] | 71.88 | 71.62 | -0.26 | 300.00 | 168.00 | 44.00 | 3.50 | - | - |
| 2.6.1 | 2022 | GNN-RL [85] | 71.87 | 70.04 | -1.83 | 300.00 | 174.00 | 42.00 | 3.50 | - | - |
| 2.4.2 | 2019 | RBP [55] | - | 69.70 | - | 300.00 | 176.47 | 41.18 | 3.50 | - | - |
| 2.4.3 | 2022 | RollBack [104] | - | 52.86 | - | 300.00 | 180.00 | 40.00 | 3.50 | - | - |
| 2.5.2 | 2021 | ManiDP [81] | 71.80 | 71.42 | -0.38 | 300.00 | 188.40 | 37.20 | 3.50 | - | - |
| 2.2.2 | 2020 | GFS [124] | 72.00 | 71.20 | -0.80 | 300.00 | 192.04 | 35.99 | 3.50 | 2.70 | 22.86 |
| 2.6.1 | 2021 | AGMC [86] | 71.80 | 70.87 | -0.93 | 300.00 | 210.00 | 30.00 | 3.50 | - | - |
| 2.2.3 | 2022 | DLRFC [44] | 71.80 | 71.88 | -0.08 | 300.00 | 210.0 | 30.00 | - | - | - |
| 2.6.1 | 2018 | AMC [103] | 71.80 | 70.80 | -1.00 | 300.00 | 210.00 | 30.00 | 3.50 | - | - |
| 2.2.2 | 2020 | GFS [124] | 72.00 | 71.60 | -0.40 | 300.00 | 210.19 | 29.94 | 3.50 | 2.90 | 17.14 |
| 2.6.2 | 2022 | DNCP [109] | 72.30 | 72.70 | 0.40 | 300.00 | 211.00 | 29.67 | 3.50 | - | - |
| 2.6.2 | 2020 | DMCP [121] | 72.30 | 73.50 | 1.20 | 300.00 | 211.00 | 29.67 | 3.50 | - | - |
| 2.6.2 | 2020 | DMCP [121] | 72.30 | 72.40 | 0.10 | 300.00 | 211.00 | 29.67 | 3.50 | - | - |
| 2.6.2 | 2022 | DDNP [100] | 72.05 | 72.20 | 0.15 | 300.00 | 211.50 | 29.50 | 3.50 | - | - |
| 2.7.1 | 2022 | RRCP [105] | - | 70.90 | - | 300.00 | 212.61 | 29.13 | 3.50 | - | - |
| 2.7.2 | 2021 | CC [60] | 71.88 | 70.91 | -0.97 | 300.00 | 215.00 | 28.33 | 3.50 | - | - |
| 2.3.1 | 2020 | PR [62] | 72.00 | 71.80 | 0.20 | 300.00 | 216.00 | 28.00 | 3.50 | - | - |
| 2.2.2 | 2020 | GFS [124] | 72.00 | 71.90 | -0.10 | 300.00 | 246.50 | 17.83 | 3.50 | 3.20 | 8.57 |
| 2.6.2 | 2020 | DMCP [121] | 72.30 | 73.90 | 1.60 | 300.00 | 300.00 | 0.00 | 3.50 | - | - |
| 2.6.2 | 2020 | DMCP [121] | 72.30 | 74.60 | 2.30 | 300.00 | 300.00 | 0.00 | 3.50 | - | - |
| 2.7.3 | 2022 | 1xN [131] | 71.74 | 70.23 | -1.50 | 300.00 | - | - | 3.50 | 1.75 | 50 |
| MobileNet-V3-Small | | | | | | | | | | | |
| 2.2.2 | 2020 | GFS [124] | 67.50 | 65.80 | -1.70 | - | - | 23.44 | - | - | 20.00 |

TABLE 22: MobileNet-V1/V2/V3-Small on ImageNet-1K.

| Section | Year | Method | Baseline Top-1 Acc.(%) | Pruned Top-1 Acc.(%) | Top-1 Acc. ↓(%) | Baseline FLOPs (M) | Pruned FLOPs (M) | FLOPs ↓(%) | Baseline Params (M) | Pruned Params (M) | Params ↓(%) |
|---|---|---|---|---|---|---|---|---|---|---|---|
| ProxylessNet-Mobile | | | | | | | | | | | |
| 2.2.2 | 2020 | GFS [124] | 74.60 | 74.00 | -0.60 | 324.00 | 232.00 | 28.40 | 4.10 | 3.40 | 17.07 |
| GoogLeNet | | | | | | | | | | | |
| 2.2.3 | 2018 | NISP [106] | - | - | 0.21 | 1560.00 | 649.90 | 58.34 | 6.80 | 4.50 | 33.76 |
| ResNeXt-50 | | | | | | | | | | | |
| 2.4.1 | 2021 | GFP [130] | 77.97 | 77.53 | -0.44 | 4230.00 | 2110.00 | 50.12 | 25.02 | - | - |
| NAS | | | | | | | | | | | |
| 2.7.2 | 2021 | NPAS [133] | - | 68.30 | - | - | 98.00 | - | - | 2.80 | - |
| 2.7.1 | 2022 | SuperTickets [134] | - | 74.20 | - | - | 125 | - | - | 2.70 | - |
| 2.7.2 | 2021 | NPAS [133] | - | 70.90 | - | - | 147.00 | - | - | 3.00 | - |
| 2.7.2 | 2021 | NPAS [133] | - | 75.00 | - | - | 201.00 | - | - | 3.50 | - |
| 2.7.2 | 2021 | NPAS [133] | - | 78.20 | - | - | 385.00 | - | - | 5.30 | - |
| 2.6.2 | 2022 | ReCNAS [112] | - | 76.90 | - | - | - | - | - | 6.30 | - |
| 2.6.2 | 2022 | ReCNAS [112] | - | 75.20 | - | - | - | - | - | 5.20 | - |
| 2.6.2 | 2022 | ReCNAS [112] | - | 77.10 | - | - | - | - | - | 6.80 | - |

TABLE 23: Other models on ImageNet-1K, including ProxylessNet-Mobile, GoogLeNet, ResNeXt-50, and models searched by Neural Architecture Search (NAS).



\end{document}